\documentclass[lettersize,journal]{IEEEtran}
\usepackage{amsmath,amsfonts, amssymb}
\usepackage{algorithmic}
\usepackage{array}
\usepackage[caption=false,font=normalsize,labelfont=sf,textfont=sf]{subfig}
\usepackage{textcomp}
\usepackage{stfloats}
\usepackage{url}
\usepackage{verbatim}
\usepackage{graphicx}
\usepackage{hyperref}
\usepackage{cite}
\usepackage{bm}
\usepackage{algorithm}
\usepackage{booktabs}
\usepackage{pifont}
\usepackage{multirow}
\usepackage{svg} 
\usepackage{color}
\usepackage{makecell}
\usepackage{tabularx}
\newcolumntype{Y}{>{\centering\arraybackslash}X}
\def\BibTeX{{\rm B\kern-.05em{\sc i\kern-.025em b}\kern-.08em
    T\kern-.1667em\lower.7ex\hbox{E}\kern-.125emX}}
\usepackage{balance}
\usepackage[T1]{fontenc}
\usepackage{xcolor}
\definecolor{coreblock}{RGB}{240,248,255} 
\usepackage{tcolorbox} 
\usepackage{soul}

\begin{document}

\title{Trusted Multi-View Learning under \\  Noisy Supervision}
\author{Yilin Zhang, Cai Xu, Han Jiang, Ziyu Guan, Wei Zhao, Xiaofei He,~\IEEEmembership{Fellow,~IAPR}, and Murat Sensoy
	\thanks{Yilin Zhang, Cai Xu, Han Jiang, Ziyu Guan, and Wei Zhao are with the School of Computer Science and Technology, Xidian University, Xi'an, Shaanxi 710071, China. \textit{E-mail: \{ylzhang\_3@stu., cxu@\}xidian.edu.cn, han522708@gmail.com, zyguan@xidian.edu.cn, ywzhao@mail.xidian.edu.cn}.}
        \thanks{Xiaofei He is with the State Key Laboratory of CAD\&CG, Zhejiang University, Hangzhou 310030, China. \textit{E-mail: xiaofeihe@cad.zju.edu.cn.}}
	\thanks{Murat Sensoy is with the Amazon Alexa AI, London, EC2A 2FA, UK. \textit{E-mail: msensoy@amazon.co.uk}.}
	\thanks{(Corresponding author: Cai Xu.)}
	}

\markboth{}%
{}
\maketitle

\begin{abstract}
	Multi-view learning methods often focus on improving decision accuracy while neglecting the decision uncertainty, which significantly restricts their applications in safety-critical scenarios. To address this, trusted multi-view learning methods estimate prediction uncertainties by learning class distributions from each instance. However, these methods heavily rely on high-quality ground-truth labels. This motivates us to delve into a new problem: how to develop a reliable multi-view learning model under the guidance of noisy labels? We propose the Trusted Multi-view Noise Refining (TMNR) method to address this challenge by modeling label noise arising from low-quality data features and easily-confused classes. 
	TMNR employs evidential deep neural networks to construct view-specific opinions that capture both beliefs and uncertainty. These opinions are then transformed through noise correlation matrices to align with the noisy supervision, where matrix elements are constrained by sample uncertainty to reflect label reliability. 
	Furthermore, considering the challenge of jointly optimizing the evidence network and noise correlation matrices under noisy supervision, we further propose Trusted Multi-view Noise Re-Refining (TMNR$^\textbf{2}$), which disentangles this complex co-training problem by establishing different training objectives for distinct modules.
	TMNR$^\textbf{2}$ identifies potentially mislabeled samples through evidence-label consistency and generates pseudo-labels from neighboring information. By assigning clean samples to optimize evidential networks and noisy samples to guide noise correlation matrices, respectively, TMNR$^\textbf{2}$ reduces mapping interference and achieves stabilizes training.
	We empirically evaluate our methods against state-of-the-art baselines on 7 multi-view datasets. Experimental results demonstrate that TMNR$^\textbf{2}$ significantly outperforms baseline methods, with average accuracy improvements of 7\% on datasets with 50\% label noise. The code and appendix are released at \href{https://github.com/YilinZhang107/TMNRR}{https://github.com/YilinZhang107/TMNRR}.
\end{abstract}


\section{Introduction}
\IEEEPARstart{M}{ulti-view} data is widely present in various real-world scenarios. For instance, in the field of healthcare, a patient's comprehensive condition can be reflected through multiple types of examinations; social media applications often include multi-modal contents such as textual and visual reviews \cite{liu2024dida}. Multi-view learning synthesizes both consistency and complementary information to obtain a more comprehensive understanding of the data \cite{MultiviewSurvey}. It has generated significant and wide-ranging influence across multiple research areas, including classification  \cite{chen2024virtual,wang2022deep, M3Net}, clustering  \cite{xu2023untie,huang2022multi,wen2022survey}, recommendation systems  \cite{lin2023contrastive,nikzad2021berters}, and large language models \cite{min2023recent,LargeMM}.

\begin{figure}[t]
	\includegraphics[width=\columnwidth]{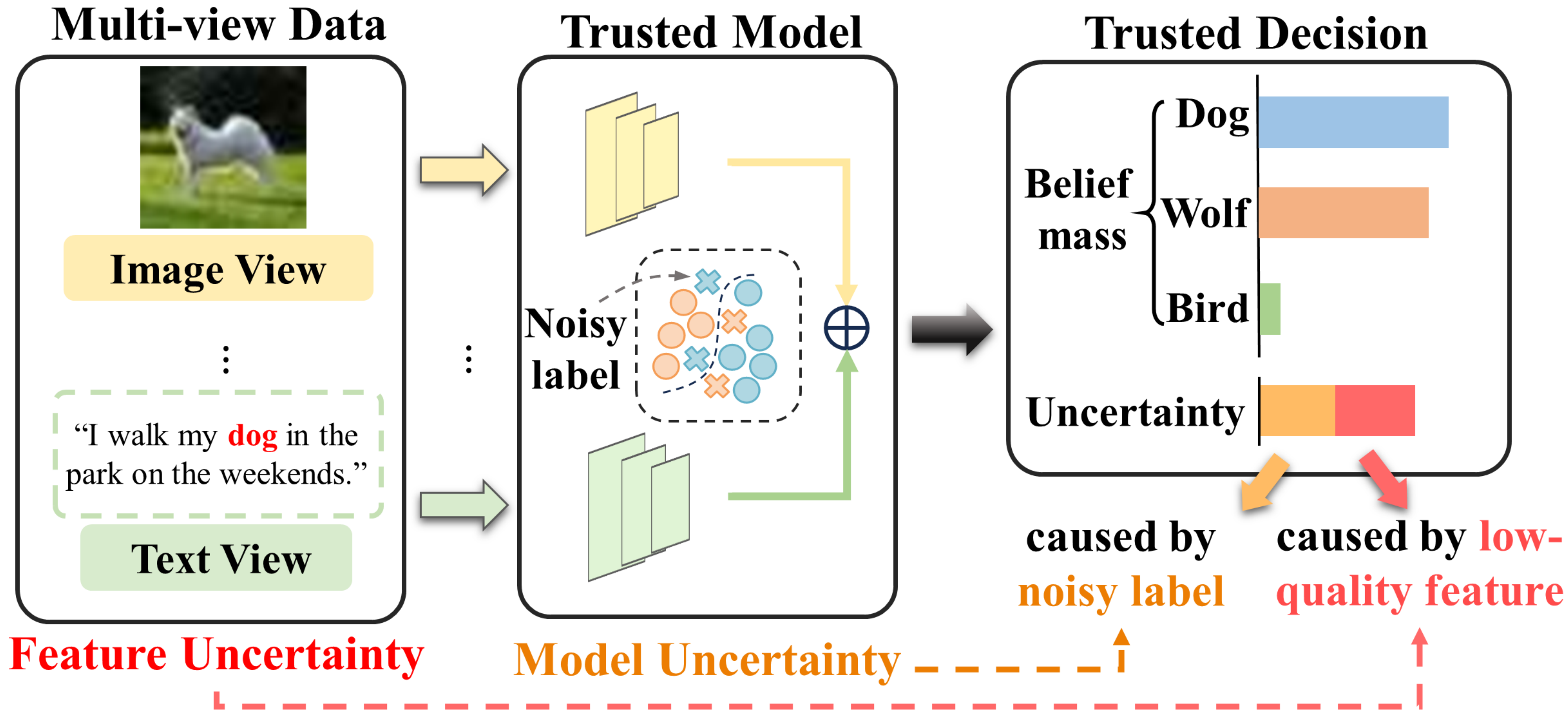}
	\caption{The generalized trusted multi-view learning problem: the model should recognize the feature and model uncertainty caused by low-quality feature and noisy labels, respectively.}
	\label{fig:introduction}
\end{figure}

Most existing multi-view learning methods focus on improving decision accuracy while neglecting the decision uncertainty. This significantly restricts the application of multi-view learning in safety-critical scenes, such as healthcare. Recently, Han \textit{et al.} propose a pioneering work \cite{TMC}, Trusted Multi-view Classification (TMC), to solve this problem. TMC calculates and aggregates the evidences of all views from the original data features. It then utilizes these evidences to parameterize the class distribution, which could be used to estimate the class probabilities and uncertainty. To train the entire model, TMC requires the estimated class probabilities to be consistent with the ground-truth labels. Following this line, researchers propose novel evidence aggregation methods, aiming to enhance the reliability and robustness in the presence of feature noise  \cite{gan2021brain,qin2022deep}, conflictive views \cite{ECML} and incomplete views  \cite{UIMC}.

Unfortunately, these trusted multi-view learning methods consistently rely on high-quality ground-truth labels. The labeling task is time-consuming and expensive especially when dealing with large scale datasets, such as user generated multi-modal contents in social media applications. This motivates us to delve into a new \textit{Generalized Trusted Multi-view Learning (GTML) problem: how to develop a reliable multi-view learning model under the guidance of noisy labels? This problem encompasses two key objectives: 1) detecting and refining the noisy labels during the training stage; 2) recognizing the model's uncertainty caused by noisy labels.}  For example, instances belonging to classes like ``dog'' and ``wolf'' might exhibit similarities and are prone to being mislabelled. Consequently, the model should exhibit higher decision uncertainty in such cases. An intuitive analogy is an intern animal researcher (model) may not make high-confidence decisions for all animals (instances), but is aware of the cases where a definitive decision is challenging.

In this paper, we propose an Trusted Multi-view Noise Refining (TMNR)\footnote{This work is our previous conference paper\cite{TMNR}.} method for the GTML problem. We consider label noises arising from two sources of aleatoric uncertainty: low-quality data features (such as blurred and incomplete features making labeling difficult) and inherent class overlap (such as naturally ambiguous cases between ``dog'' and ``wolf'' classes leading to inconsistent labeling). Our objective is to leverage multi-view consistent information for noise detection. To achieve this, TMNR first constructs the view-specific evidential Deep Neural Networks (DNNs) to learn view-specific evidence, then models the view-specific distributions of class probabilities using the Dirichlet distribution, parameterized with view-specific evidence. These distributions allow us to construct opinions, which consist of belief mass vectors and uncertainty estimates. TMNR designs view-specific noise correlation matrices (square matrices modeling the probability of true labels being corrupted into noisy labels) to transform the original opinions into noisy opinions aligned with the noisy labels. Specifically, the diagonal elements of these matrices represent the probability of retaining the current label as the ground truth. To account for the fact that low-quality data features are prone to mislabeling, TMNR constrains the diagonal elements to be inversely proportional to the uncertainty. Meanwhile, the off-diagonal elements capture class confusion probabilities, with higher values assigned to easily-confused classes. Next, TMNR aggregates the noisy opinions to obtain the common evidence. Finally, TMNR employs a generalized maximum likelihood loss on the common evidence, guided by the noisy labels, for model training.

TMNR establishes the relationship between the predicted evidence distribution and the corrupted labels by constructing view-specific noise correlation matrices. To establish these associations, we need to fit the mapping from samples to latent clean evidence distributions, as well as the mapping from clean evidence distributions to noisy labels \cite{dualT}. However, due to the lack of supervision of the latent clean evidence distribution, it is difficult for TMNR to learn the noise correlation matrix and the evidence neural network in parallel. They are prone to interfere with each other, leading to inadequate optimization of both. Especially in datasets with a high noise ratio, overfitting to erroneous labels that overshadow correct labels leads to large errors in the posterior probability estimation for noisy samples, thereby substantially reducing classification performance.

Therefore, we propose a Trusted Multi-view Noise Re-Refining (TMNR$^\textbf{2}$) method. By distinguishing between clean and noise samples to cope with the training objectives of different modules in TMNR: clean samples are used to optimize the evidential neural network, while noise samples focus on identifying more accurate noise association patterns. Specifically, TMNR$^\textbf{2}$ considers samples with low similarity between the label vector and the predicted evidence distribution (including itself and its neighboring samples) as noise samples and relabels them. Subsequently, the noisy samples are fused with the clean samples with higher similarity by mixup operation to extend the training distribution of the samples and improve the noise immunity of the model. Finally, a bounded $l_2$ loss is used to mildly constrain the original evidence distribution of the mixed samples, which effectively reduces the difficulty of co-training the modules in TMNR. In summary, the contributions of this work are:
\begin{enumerate}
	\item{We propose the generalized trusted multi-view learning problem, which necessitates the model's ability to make reliable decisions despite the presence of noisy guidance.}
	\item{We developed a trusted multi-view noise learning framework that leverages multi-view consistent information to detect and refine noisy labels, while assigning higher decision uncertainty to instances more susceptible to mislabeling.} 
	\item{We proposed two trusted multi-view noise learning methods, TMNR and TMNR$^\textbf{2}$. TMNR learns the latent patterns of label corruption through the designed noise correlation matrix. On this basis, TMNR$^\textbf{2}$ explicitly utilizes neighboring information to identify and refine noisy samples, supervising the intermediate variables in the two-stage mapping from samples to clean predictions and then to noisy labels, effectively reducing the framework's optimization complexity.}
	\item{We verify the effectiveness of TMNR and TMNR$^\textbf{2}$ on 7 publicly available datasets for the trusted classification task in noisy labeled scenario. Experimental results  show their superiority over the state-of-the-art trusted multi-view learning and noisy labeling learning baselines.}
\end{enumerate}

The remainder of this article is organized as follows. In Section \ref{relatedwork}, we review some related works on deep multi-view fusion and label-noise learning. Section \ref{TMNR} elaborates the TMNR together with its implementation. Section \ref{TMNR2} introduces the TMNR$^\textbf{2}$. Section \ref{experiments} shows the experimental evaluations and discussions. Section \ref{conclusion} concludes this article.


\section{Related Work}
\label{relatedwork}

\subsection{Deep Multi-view Fusion}
Multi-view fusion has demonstrated its superior performance in various tasks by effectively combining information from multiple sources or modalities  \cite{liang2021evolutionary,Calm, dong2023multi}. According to the fusion strategy, existing deep multi-view fusion methods can be roughly classified into feature fusion \cite{hu2024deep,xu2022multi,liu2023information,MMatch} and decision fusion \cite{jillani2020multi,OpinionAggregation}. A major challenge of feature fusion methods is each view might exhibit different types and different levels of noise at different points in time. Trust decision fusion methods solve this making view-specific trust decisions to obtain the view-specific reliabilities, then assigning large weights to these views with high reliability in the multi-view fusion stage. Following this line,  Xie \textit{et al.} \cite{UIMC} tackle the challenge of incomplete multi-view classification through a two-stage approach, involving completion and evidential fusion. Xu \textit{et al.} \cite{ECML} focus on making trust decisions for instances that exhibit conflicting information across multiple views. They propose an effective strategy for aggregating conflicting opinions and theoretically prove this strategy can exactly model the relation of multi-view common and view-specific reliabilities. However, it should be noted that these trusted multi-view learning methods heavily rely on high-quality ground-truth labels, which may not always be available or reliable in real-world scenarios. This limitation motivates for us to delve into the problem of GTML, which aims to learn a reliable multi-view learning model under the guidance of noisy labels.

\begin{figure}[tb] 
	\centering
	\includegraphics[width=\columnwidth]{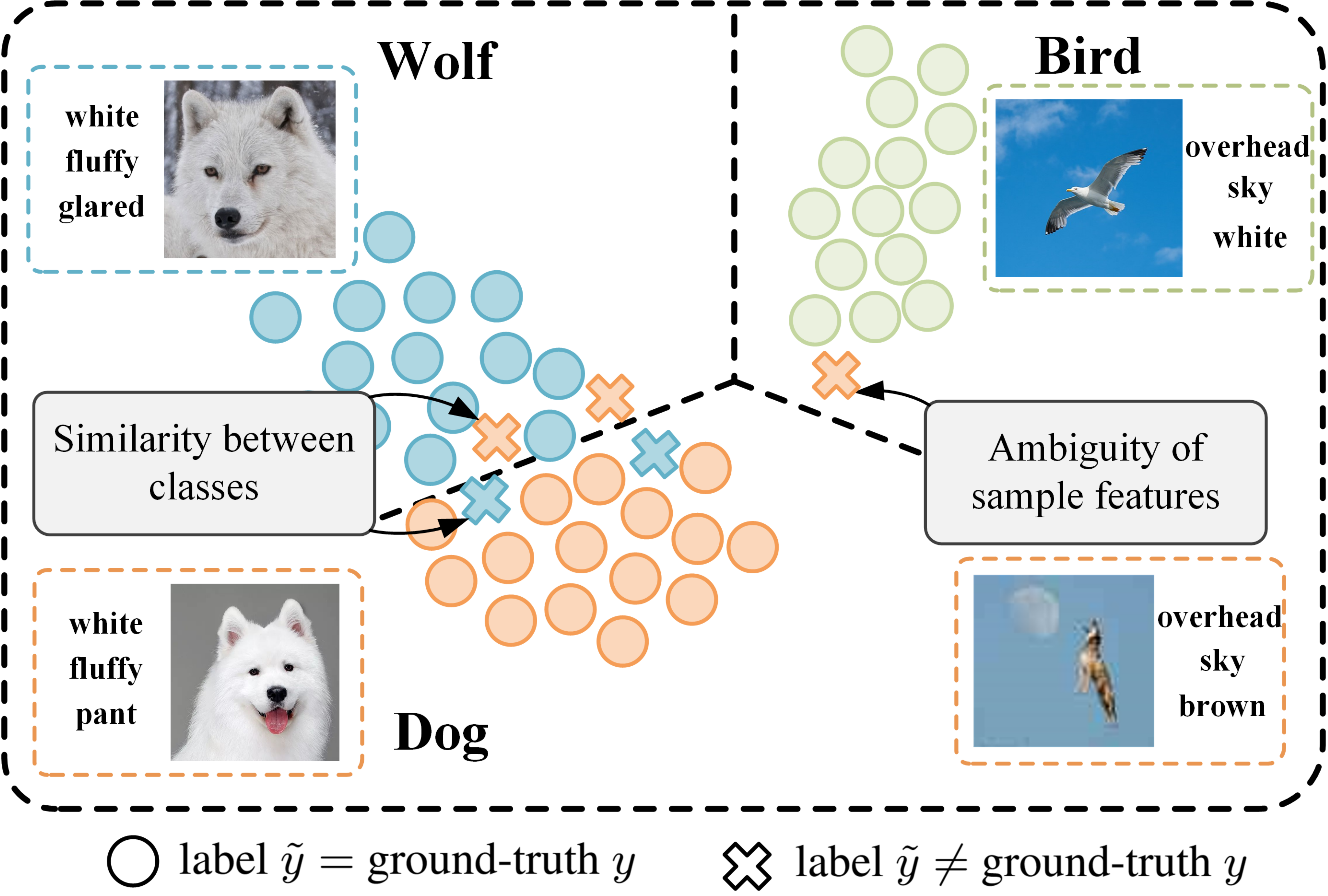} 
	\caption{Illustration of the instance-dependent label noise. Each color represents a ground-truth category $y$.}
	\label{fig:two type}
\end{figure}

\subsection{Label-Noise Learning}
In real-world scenarios, data labeling is often error-prone, subjective, or costly, resulting in noisy labels. Label-noise learning addresses the challenge of training models with datasets contaminated by such label inaccuracies.
In multi-class classification tasks, label noise is commonly classified into two main categories: Class-Conditional Noise (CCN) and Instance-Dependent Noise (IDN)\cite{noiseSurvey2022TNNLS}. CCN arises  when the label corruption process is independent of the data features, and instances in a class are assigned to other classes with a fixed probability. 
To mitigate CCN, existing methods typically correct losses by estimating a global category transition matrix \cite{FC,gold_correct}. IDN refers to instances being mislabeled based on their class and features. In this work, we focus on IDN as it closely resembles real-world noise. The primary challenge lies in approximating the complex and high-dimensional instance-dependent transition matrix. Several approaches have been proposed to address this challenge. For instance,  Cheng \textit{et al.} \cite{NoiseMethod} proposes an instance-dependent sample sieve method that enables model to process clean and corrupted samples individually.  Cheng \textit{et al.} \cite{manifold-regularized}  effectively reduce the complexity of the instance-dependent matrix by streaming embedding.  Berthon \textit{et al.} \cite{CSIDN} approximate the transition distributions of each instance using confidence scores. However, this approach relies on pre-trained models, which may not always be reliable.

As discussed, while much effort has been dedicated to multi-view learning and label-noise handling methods, existing algorithms suffer from the following limitations and challenges:
\begin{itemize}
    \item{Most trusted multi-view learning methods rely heavily on clean labels for reliable decision-making, lacking mechanisms to handle noisy supervision effectively.}
    \item{Existing label-noise learning methods typically focus on single-view data and fail to leverage the complementary information from multiple views for noise mitigation.}  
\end{itemize}

The proposed TMNR framework constructs the noise correlation matrix by leveraging multi-view opinions, enabling effective learning in noisy scenarios. Building upon TMNR, TMNR$^\textbf{2}$ further refines noisy samples through neighboring information and Mixup. Their quantification of uncertainty induced by noisy labels further ensures the reliability of decisions.


\section{Trusted Multi-view Noise Refining Classification}
\label{TMNR}
In this section, we first define the generalized trusted multi-view learning problem, then present Trusted Multi-view Noise Refining (TMNR) in detail, together with its implementation.

\subsection{Notations and Problem Statement}
We use $\{ \bm{x}^v_n \in \mathbb{R}^{d_v} \}_{v=1}^{V}$ to denote the feature of the $n$-th instance, which contains $V$ views, $d_v$ denotes the dimension of the $v$-th view. $y_n\in\{1,... ,C\}$ denotes the ground-truth category, where $C$ is the number of all categories. In the generalized trusted multi-view classification problem, the labels of some data instances contain noise as shown in Figure \ref{fig:two type}. Therefore, we utilize $\{ \tilde{y}_n\in\{1,... ,C\} \}_{n=1}^{N}$ as the set of noisy labels that may have been corrupted

The objective is to learn a trusted classification model according to noisy training instances ${\{\{\bm{x}_n^v\}_{v=1}^V},\tilde{y}_n\}_{n=1}^{N_{train}}$. For the instances of the test sets, the model should predict the category $\{ {y}_n \}$ and uncertainty $\{ {u}_n \}$, which can quantify the uncertainty caused by low-quality feature and noisy labels.

\begin{figure*}[t]
	\centering
	\includegraphics[width=\textwidth]{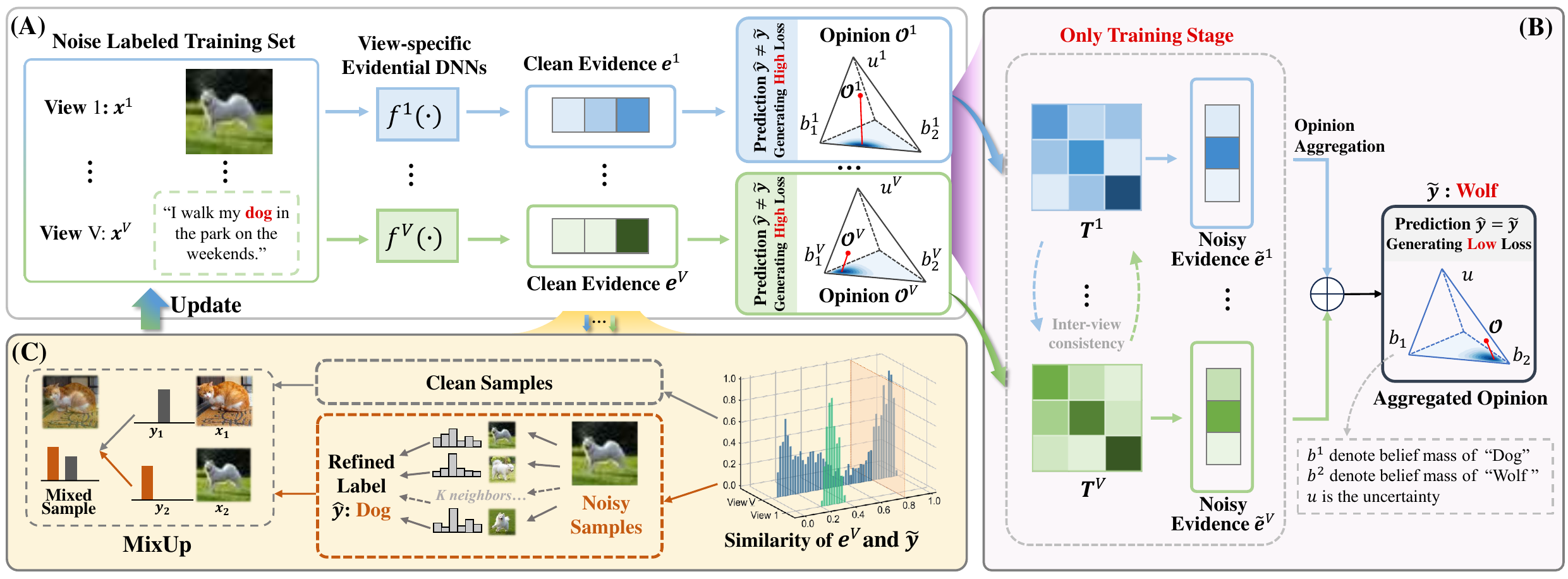}
	\caption{
		Illustration of our algorithm. In TMNR, we first use (A) Evidence Extraction Module to extract the evidence for each view and form its opinion through view-specific evidential DNNs $\{ f^v( \cdot ) \}_{v=1}^V$, then employ (B) Noise Opinion Alignment and Aggregation Module to transform original opinions into noisy opinions aligned with noisy labels through view-specific noise correlation matrices $\{ \bm{T}^v \}_{v=1}^V$, followed by aggregation and training the entire network using only the noisy labels. 
		In TMNR$^\textbf{2}$, we further introduce (C) Noise Identification and Re-refining Module, which mitigates overfitting caused by noisy labels and reduces joint optimization complexity by identifying noise samples and obtaining pseudo-labels $\hat{y}$ for mislabeled samples based on clean evidence distributions $\bm{e}^v$ of the samples themselves and their neighbors. Noise samples are then mixed with clean samples through mixup operations to extend training distributions and increase model noise resistance, while the labels of mixed samples directly impose mild constraints on the model's intermediate variables (clean evidence distributions), effectively reducing the complexity of the parameter space.}
	\label{fig:model}
\end{figure*}

\subsection{Trusted Multi-view Noise Refining Pipeline}
As shown in Figure \ref{fig:model}, We first construct view-specific opinions using evidential DNNs $\{ f^v( \cdot ) \}_{v=1}^V$. To account for the presence of label noise, the view-specific noise correlation matrices $\{ \bm{T}^v \}_{v=1}^V$ transform the original opinions into noisy opinions aligned with the noisy labels. Finally, we aggregates the noisy opinions and trains the whole model by the noisy labels. Details regarding each component will be elaborated as below.

\subsubsection{\textbf{View-specific Evidence Learning}}
\label{Evidence-BasedMulti-viewLearning}
In this subsection, we introduce the evidence theory to quantify uncertainty. Traditional multi-classification neural networks usually use a Softmax activation function to obtain the probability distribution of the categories. However, this provides only a single-point estimate of the predictive distribution, which can lead to overconfident results even if the predictions are incorrect. This limitation affects the reliability of the results. To address this problem, EDL \cite{EDL} introduces the evidential framework of subjective logic \cite{SubjectiveLogic}. It converts traditional DNNs into evidential neural networks by making a small change in the activation function to get a non-negative output (e.g.,ReLU) to extract the amount of support (called evidence) for each class.

In this framework, the parameter $\bm{\alpha}$ of the Dirichlet distribution $Dir(\bm{p}|\bm{\alpha})$ is associated with the belief distribution in the framework of evidence theory, where $\bm{p}$ is a simplex representing the probability of class assignment. We collect evidence, $\{ \boldsymbol{\mathit{e}}_{n}^{v}  \}$ by view-specific evidential DNNs $\{ f^v (\cdot) \}_{v=1}^V$. The corresponding Dirichlet distribution parameter is $\bm{\alpha}^v = \bm{e}^v+1=[\alpha_1^v,\cdots,\alpha_C^v]^T$. After obtaining the distribution parameter, we can calculate the subjective opinion ${\mathcal{O}}^v=({\bm{b}^v,u^v})$ of the view including the quality of beliefs $\bm{b}^v$ and the quality of uncertainty $u$, where $\bm{b}^v=(\bm{\alpha}^v - 1) / S^v=\bm{e}^v / S^v$, $u^v=C / S^v$, and $S^v=\sum_{c=1}^C\alpha_c^v$ is the Dirichlet intensity.

\subsubsection{\textbf{Evidential Noise Forward Correction}}
In the GTML problem, we expect to train the evidence network so that its output is a clean evidence distribution about the input. To minimise the negative impact of IDN in the training dataset, we modify the outputs of the DNNs with an additional structure to adjust the loss of each training sample before updating the parameters of the DNNs. This makes the optimisation process immune to label noise, called evidential noise forward correction \cite{BeyondimagesLabelnoise,Areanchorpoints}. This structure should be removed when predicting test data.

For each view of a specific instance, we construct view-specific noise correlation matrix to model the label corruption  process:
\begin{equation}
	\label{defineT}
	\bm{T}^v =[t_{cj}^v]_{c,j=1}^C  \in [0,1]^{C \times C},
\end{equation}
where ${t}_{cj}^v:=P(\tilde{y}=j|y=c ,\bm{x}^v)$ represents the probability that the instance with a true label $c$ is mislabeled as class $j$, and $\sum_{j=1}^{C}t^v_{cj}=1$.

As a bridge between the true and noisy distribution, the noisy category-posterior probability can be inferred from the noise correlation matrix $\bm{T}^v$ and the clean category-posterior probability:
\begin{align}
	P(\tilde{y}=j | \bm{x}^v) & =\sum_{c=1}^{C}P(\tilde{y}=j,y=c|\bm{x}^v) \nonumber \\
	                          & =\sum_{c=1}^{C}{t}^v_{cj}P(y=c|\bm{x}^v).
\end{align}

Building upon the evidence theory described in the previous subsection \ref{Evidence-BasedMulti-viewLearning} and considering the internal constraints within $\bm{T}^v$, we convert the process of transferring the predicted probabilities of instances into a transfer of the extracted support at the evidence level, as described by the following equation:
\begin{align}
	\label{PtoE}
	\tilde{e}_j^v & =\sum_{c=1}^{C}t_{cj}^v \ e_c^v ,
\end{align}
where $\tilde{e}_j^v$/${e}_c^v$ denotes the $j$-/$c$-th element in the noise/clean class-posterior evidence vectors $\tilde{\bm{e}}^v$/${\bm{e}}^v$.

Therefore, the clean posterior $\bm{e}^v$ obtained from the prediction of each view is transferred to the noisy posterior $\tilde{\bm{e}}^v$. The parameter $\tilde{\bm{\alpha}}$ in the noisy Dirichlet distribution is computed in order to align the supervised labels $\tilde{y}$ that contain the noise. The entire evidence vector could be calculabed by:
\begin{equation}
	\label{transferE}
	\tilde{\bm{e}}^v = {\bm{T}^v}^\top\bm{e}^v, ~ ~ \tilde{\bm{\alpha}}^v=\tilde{\bm{e}}^v+1 .
\end{equation}

\subsubsection{\textbf{Trust Evidential Multi-view Fusion}}
After obtaining opinions from multiple views, we consider dynamically integrating them based on uncertainty to produce a combined opinion. We achieve this via the Dempster's combination rule \cite{SubjectiveLogic}. We take the clean opinion aggregation as example. Given clean opinions of two views of the same instance, i.e., ${\mathcal{O}}^1=(\bm{b}^1, u^1)$ and ${\mathcal{O}}^2=(\bm{b}^2, u^2)$, the computation to obtain the aggregated opinion ${\mathcal{O}}=\mathcal{O}^1 \diamond \mathcal{O}^2=(\bm{b}, u)$ is defined as follows:
\begin{equation}
	\label{aggregation}
	b_c=\frac{1}{1-Con}\left(b_c^1b_c^2+b_c^1u^2+b_c^2u^1\right),u=\frac{1}{1-Con}u^{1}u^{2},
\end{equation}
where $Con=\sum_{i\neq j}b_{i}^{1}b_{j}^{2}$ measures the degree of conflict between two belief mass sets and $\frac{1}{1-Con}$ is used as a scaling factor to normalize the fusion uncertainty.

For a set of opinions from multiple views ${\{\mathcal{O}^v\}}_{v=1}^V$, the joint multi-view subjective opinion is obtained by $\mathcal{O}=\mathcal{O}^1\diamond\mathcal{O}^2\diamond\cdots \mathcal{O}^V$. The corresponding parameters of the $Dir(\bm{p}|\bm{\alpha})$ are obtained with $S=C/u$, $\alpha_c=b_c \times S + 1$ and the final probability could be estimated via $p_c=\alpha_c / S$.

\subsection{Loss Function}
\label{similaritymatrix}
In this section, we explore the optimization of the parameter set $\{\bm{\theta}, \bm{\omega}\}$ in the evidence extraction network $f(\cdot; \bm{\theta})$ and the correlation matrices $\{\{\bm{T}_n^v\}_{v=1}^V\}_{n=1}^{N_{train}}$.


\subsubsection{\textbf{Classification Loss}}

We capture view-specific evidence from a single view $\bm{x}^v$ of a sample. The vector $\bm{e}^v = f^v(\bm{x}^v)$ denotes the clean class-posterior evidence obtained from the corresponding view network prediction. This evidence undergoes correction through Eq.\eqref{transferE} to yield the noisy class-posterior evidence, denoted as $\tilde{\bm{e}}^v$, along with the associated Dirichlet parameter $\tilde{\bm{\alpha}}^v$.
Then in the constructed inference framework, the evidence multi-classification loss $\mathcal{L}$ containing the classification loss $\mathcal{L}_{ace}$ and the Kullback-Leibler (KL) divergence term $\mathcal{L}_{KL}$ is defined, where the classification loss $\mathcal{L}_{ace}$ is obtained by adjusting for the conventional cross-entropy loss, i.e., the generalized maximum likelihood loss, as follows:
\begin{equation}
	\label{lossace}
	\mathcal{L}_{ace}(\tilde{\bm{\alpha}}^v)=\sum_{c=1}^{C} \tilde{y}_{c}\left(\psi\left(\tilde{S}^v\right)-\psi\left(\tilde{\alpha}_{c}^v\right)\right),
\end{equation}
where $\psi(\cdot)$ is the digamma function,  $\tilde{\alpha}_{j}^v$ denote the $j$-th element in $\tilde{\bm{\alpha}}^v$.

Eq.\eqref{lossace} does not ensure that the wrong classes in each sample produce lower evidence, and we would like it to be reduced to zero. Thus the KL divergence term is expressed as:
\begin{align}
	\mathcal{L}_{KL}(\tilde{\bm{\alpha}}^v) & = \mathrm{KL}\left[D(\bm{p}^v|\bar{\bm{\alpha}}^v)\parallel D(\bm{p}^v|\mathbf{1})\right] \nonumber                                                                         \\[3pt]
	                                        & =  \log \left( \begin{small} \frac{\smash{\Gamma\bigl(\sum\nolimits_{c=1}^C \bar{\alpha}_c\bigr)}}{\smash{\Gamma(C) \prod\nolimits_{c=1}^C \Gamma\bigl(\bar{\alpha}_c\bigr)}} \end{small} \right)                                                                      \\[2pt]
	                                        & + \textstyle\sum\nolimits_{c=1}^C \left(\bar{\alpha}_c-1\right)\left[\psi\bigl(\bar{\alpha}_c\bigr)-\psi\bigl(\sum\nolimits_{j=1}^C \bar{\alpha}_j\bigr)\right],  \nonumber
\end{align}
\noindent where $\bar{\bm{\alpha}}^v=\tilde{\mathbf{y}}+(1-\tilde{\mathbf{y}})\odot \tilde{\bm{\alpha}}^v$ is the Dirichlet distribution parameters after removal of the non-misleading evidence from predicted parameters $\tilde{\bm{\alpha}}^v$, thereby avoiding penalization for class $\tilde{y}$. $\tilde{\mathbf{y}}$ denotes the noisy label $\tilde{y}$ in the form of a one-hot vector, and $\Gamma(\cdot)$ is the gamma function. Thus for a given view's Dirichlet parameter $\tilde{\bm{\alpha}}^v$, the view-specific loss is
\begin{equation}
	\mathcal{L}(\tilde{\bm{\alpha}}^v) = \mathcal{L}_{ace}(\tilde{\bm{\alpha}}^v) + \delta \mathcal{L}_{KL}(\tilde{\bm{\alpha}}^v),
\end{equation}
where $\delta \in [0,1]$ is a changeable parameter, and we gradually increase its value during training to avoid premature convergence of misclassified instances to a uniform distribution.

\subsubsection{\textbf{Uncertainty-Guided Correlation Loss}}

Optimizing the view-specific correlation matrix, being a function of the high-dimensional input space, presents challenges without any underlying assumptions. In the context of the IDN problem, it is important to recognize that the probability of a sample being mislabeled depends not only on its category but also on its features. When the features contain noise or are difficult to discern, the likelihood of mislabeling increases significantly. As highlighted earlier, the uncertainty provided by the evidence theory has proven effective in assessing the quality of sample features. Therefore, it is natural for us to combine the uncertainty estimation with the IDN problem, leveraging its potential to enhance the overall performance.

In our work, we do not directly reduce the complexity of correlation matrix by simplifying it. Instead, we propose an assumption that ``the higher the uncertainty of the model on the decision, the higher the probability that the sample label is noisy''. Based on this assumption, a mild constraint is imposed on the correlation matrix to effectively reduce the degrees of freedom of its linear system. Specifically, based on the obtained Dirichlet parameters $\tilde{\bm{\alpha}}^v$ with its corresponding opinion uncertainty $u^v$.
We impose different constraints on various parts of the correlation matrix $\bm{T}^v$, aiming to encourage it to transfer evidence for instances with higher uncertainty and uncover potential labeling-related patterns.

\textbf{Diagonal elements.} Since the diagonal element $\{t_{cc}^v\}_{c=1}^C$ in the $\bm{T}^v$ corresponds to the probability that the labelled category is equal to its true category. Meanwhile the confidence we obtain from subjective opinions is only relevant for the diagonal elements corresponding to their labelled category $\tilde{y}$, for $\{t_{cc}^v\}_{c=1,c\neq \tilde{y}}^C$, $u^v$ no longer provides any direct information. Therefore, we simply make the other diagonal elements close to the confidence mean of the corresponding class of samples from the current batch. It can be expressed as:
\begin{equation}
	\mathcal{M_D}(\tilde{\bm{\alpha}}^v) = \sum_{c=1}^C \mathcal{M_D}_{c}(\tilde{\bm{\alpha}}^v),
\end{equation}
\begin{equation}
	\mathcal{M_D}_{c}(\tilde{\bm{\alpha}}^v) =
	\begin{cases}
		[(1-u^v) - t_{cc}^v]^2 ,      & \text{if} ~~ c=\tilde{y} ,   \\
		[(1-\bar{u}_c^v)-t_{cc}^v]^2, & \text{if}~~c\neq \tilde{y} ,
	\end{cases}
\end{equation}
where $u^v= C/\sum_{c=1}^C\tilde{\alpha}_{c}^v$ ,  $\bar{u}_c^v$ is the average of the $u^v$ of all samples with label $\tilde{y} = c$ in the current batch.


\textbf{Non-diagonal elements.} The constraints on the diagonal elements could be regarded as guiding the probability of the sample being mislabeled. The probability of being mislabeled as another category is influenced by the inherent relationship between the different categories. For example, ``dog'' is more likely to be labeled as ``wolf'' than as ``plane''. Considering that samples in the same class can be easily labeled as the same error class, the transfer probabilities of their non-diagonal elements should be close. In addition, since the labeled information may contain noise, we aim to eliminate the misleading of the error samples in the same class. To solve this problem, we construct the view-specific similarity graph (affinity matrix)  $\{\bm{S}^v\}_{v=1}^V$ for each view and identify the $K$ samples most similar to each instance, assuming they may share similar noise transition relationships:

\begin{equation}
	s_{nm}^v= \operatorname{exp} \left({-\frac{||\bm{x}_n^v-\bm{x}_m^v||^2}{\sigma^2}}\right),  n,m\in \{1,\dots,N_{train}\},
\end{equation}
\begin{equation}
	\label{neighbor}
	\mathcal{N}(\bm{x}_n^v, K) = \{k \mid k \in \operatorname{argsort}_{k \neq n} \left( -s_{nk} \right)[:K]\},
\end{equation}
where $s_{nm}^v$ denotes the $(n, m)$-th element in the affinity matrix ${\bm{S}}^v$ for the $v$-th view, which measures the similarity between $\bm{x}_n^v$ and $\bm{x}_m^v$.
\ $\mathcal{N}(\bm{x}_n^v,K)$ represent the index set of the $K$ nearest neighbors to the instance $\bm{x}_n^v$, and $\sigma$ is the average Euclidean distance between  $\bm{x}_n^v$  and its $K$ nearest neighbors.
Then, constraints are imposed on the off-diagonal elements of the noise correlation matrix based on their adjacencies as follows,
\begin{equation}
	\mathcal{M_O} (\tilde{\bm{\alpha}}_n^v)= \sum_{k=1}^{K}s_{nk}^v\| \bar{ \bm{T}}^v_n-\bar{\bm{T}}^v_k\|^{2},
	\  k \in\mathcal{N}(\bm{x}_n^v,K),
\end{equation}
\noindent where $\bar{\bm{T}}^v_i$ is the matrix after zeroing the diagonal elements of $\bm{T}^v_i$.
Thus, the overall regularization term for the inter-sample uncertainty bootstrap is expressed as:
\begin{equation}
	\mathcal{M} (\tilde{\bm{\alpha}}^v)= \mathcal{M_D} (\tilde{\bm{\alpha}}^v)+\mathcal{M_O} (\tilde{\bm{\alpha}}^v).
\end{equation}

\textbf{Inter-view consistency.}
In multi-view learning, each view represents different dimensional features of the same instance. In our approach, we leverage the consistency principle of these views to ensure the overall coherence of the correlation matrix across all views. The consistency loss is denoted as:
\begin{equation}
	\mathcal{L}_{con} =  \frac{1}{V}\sum_{v = 1}^{V}\left (
	{\textstyle \sum_{c=1}^{C}}   {\textstyle \sum_{j=1}^{C}  | t_{cj}^v- \bar{t}_{cj}   |}
	\right )  ,
\end{equation}
where $\bar{t}_{cj}=  (   \sum_{v=1}^{V}   t_{cj}^v   )/\ V$.

\subsubsection{\textbf{Overall Loss}}
To sum up, for a multi-view instance $\{\bm{x}^v\}^V_{v=1}$, to ensure that view-speicfic and aggregated opinions receive supervised guidance. We use a multitasking strategy and bootstrap the correlation matrix according to the designed regularization term:
\begin{equation}
	\label{overallLoss}
	\mathcal{L}_{all}=\mathcal{L}(\tilde{\bm{\alpha}})+
	\sum_{v=1}^V \left[ \mathcal{L}\left(\tilde{\bm{\alpha}}^v\right)
		+ \beta \left(  \mathcal{M}(\tilde{\bm{\alpha}}^v) \right)
		\right] + \gamma \mathcal{L}_{con},
\end{equation}
where $\beta$ and $\gamma$ are hyperparameters that balances the adjusted cross-entropy loss with the uncertainty bootstrap regularisation and the inter-view consistency loss. ${\tilde{\bm{\alpha}}}$ is obtained by aggregating multiple noise class-posterior parameters $\{ \tilde{\bm{\alpha}}^v \}_{v=1}^V$.

\section{Trusted Multi-view Noise Re-refining Classification}
\label{TMNR2}
TMNR identifies potential corrupting patterns of instance-dependent labelling noise by means of noise correlation matrix $\bm{T}$ and sample uncertainty, however, the following limitations still exist during practical implementation.
\begin{enumerate}
	\item{TMNR needs to simultaneously fit a two-stage mapping from sample features to potentially clean evidence distributions, and then to noise labels. However, due to the lack of explicit supervision on the potential clean distribution, the joint learning process of the evidence neural network and the noise association matrix is prone to interfere with each other and difficult to converge to the global optimum.}
	\item{When the proportion of noise in the dataset is high, TMNR tends to overfit a large number of noisy samples, leading to a significant underestimation of the prediction uncertainty on the noisy samples. This not only destroys the assumed relationship between uncertainty and labelling error probability, but also makes the noise correlation matrix unable to accurately identify and correct the noisy samples, hence seriously affecting the generalization performance of the model.}
\end{enumerate}

To address the aforementioned challenges, we propose Trusted Multi-view Noise Re-Refining (TMNR$^\textbf{2}$) inspired by the sample division-based noise label learning strategies \cite{mixmatch,DivideMix}. By separating the clean and noisy training samples, reliable supervised information is provided for the two-stage mapping fitting of TMNR. Additionally, the use of mixed samples smooths the training data distribution, enhancing the model's accuracy in predicting uncertainty for noisy instances.

\begin{figure}[tb] 
	\centering
	\includegraphics[width=\columnwidth]{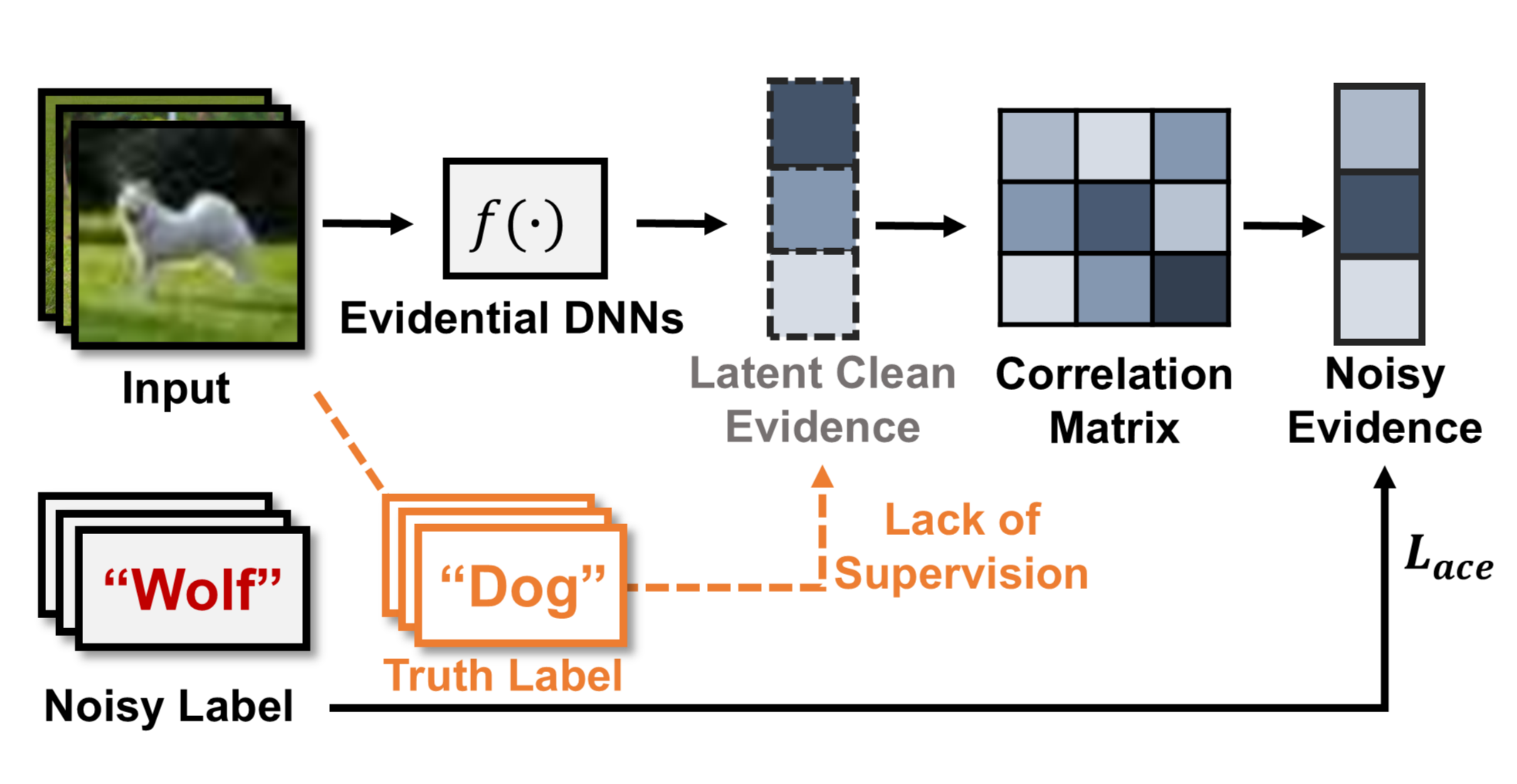}
	\caption{The motivation for TMNR$^\textbf{2}$ arises from the difficulty TMNR faces in jointly learning the evidential DNNs and noise correlation matrix, due to the lack of supervision for the latent clean evidence distribution in the two-stage mapping process.}
	\label{fig:TMNRR-motivation}
\end{figure}

The overall architecture of TMNR$^\textbf{2}$ is illustrated in Figure \ref{fig:model}. Its key improved components include Multi-view Collaborative Noise Identification and Noise Sample Re-refining. The noise identification module calculates the similarity between evidence distribution $\bm{e}$ and the one-hot label vector $\tilde{\mathbf{y}}$ for the sample and their neighbors across different views.
These similarity results are aggregated using the variance-based weighting strategy formulated in Eq.(\ref{viewfusion}), which dynamically prioritizes reliable views to enhance inter-sample discrimination. 
The Noise Sample Re-refining module generates pseudo-labels for the noisy samples and incorporates the mixup operation to extend the training distribution, thereby enhancing the model's robustness against noise. Additionally, TMNR$^\textbf{2}$ employs bounded $l_2$ loss to mildly constrain the original evidence distributions of mixed samples directly, reducing the overall optimization difficulty while preventing the pseudo-labels from leading to over-accumulation of cognitive bias in the model. Details regarding each component and the overall optimization will be elaborated as below.

\subsection{Multi-view Collaborative Noise Identification}
Deep networks tend to learn clean samples faster than noisy ones \cite{arpit2017closer, arazo2019unsupervised}, leading to earlier convergence on clean sample labels \cite{chen2019understanding, NLC}. Inspired by this, we identify mislabeled samples using predicted evidence distributions from multiple views. A sample is considered likely to be mislabeled if both its own predicted evidence distribution and those of its feature-similar neighbors deviate significantly from the sample's label vector. Specifically, as in the backbone network of TMNR, we first infer the predicted evidence distributions $\{\{ \bm{e}_{n}^{v} \}_{v=1}^V\}_{n=1}^{N_{train}}$ for all training samples  $\{\{ \bm{x}_{n}^{v} \}_{v=1}^V\}_{n=1}^{N_{train}}$ through the view-specific evidence networks $\{ f^v(\cdot)\}_{v=1}^V$.

For the $n$-th sample's observation $\bm{x}_n^v$ in the $v$-th view, we can obtain the index set $\mathcal{N}$ of its $K$-nearest neighbors based on Eq.\eqref{neighbor}. The consistency between its label vector $\tilde{\mathbf{y}}_n$ and the evidence distribution of $\bm{x}_n^v$ along with those of its neighbors is defined as follows:
\begin{align}
	\mathcal{H}^v(\bm{x}_n^v) & =\sum_{k=1}^{K} s_{nm_k}^v \operatorname{JS}\left[\tilde{\mathbf{y}}_n \parallel \operatorname{softmax}(\bm{e}_{m_k}^v)\right]                      \\
	                          & + \operatorname{JS}\left[\tilde{\mathbf{y}}_n \parallel \operatorname{softmax}(\bm{e}_n^v)\right], \  m_k \in \mathcal{N}(\bm{x}_n^v, K), \nonumber
\end{align}
where $s_{n{m_k}}^v$ acts as a weighting term, denotes the similarity between the observation and its $k$-th nearest neighbor, while the similarity of $\bm{x}_n^v$ to itself defined as 1.

We employ Jensen-Shannon (JS) divergence \cite{JS} instead of KL divergence to measure the consistency between two distributions as it is upper bounded and symmetric, aligning better with our intuitive expectations for consistency measurement. Given two probability distributions $P$ and $Q$, the computation of their JS divergence can be expressed as:
\begin{equation}
	\operatorname{JS} \left[P\parallel Q  \right] = \frac{1}{2} \operatorname{KL}[P \parallel M] + \frac{1}{2} \operatorname{KL}[Q \parallel M],
\end{equation}
where $M = \frac{1}{2}(P + Q)$, the JS divergence is bounded within $[0,1]$. When $\operatorname{JS}[\cdot \parallel \cdot ]=0$, it means that the two probability distributions are absolutely identical.

\begin{figure}[t]
	\centering
	\includegraphics[width=2.8in]{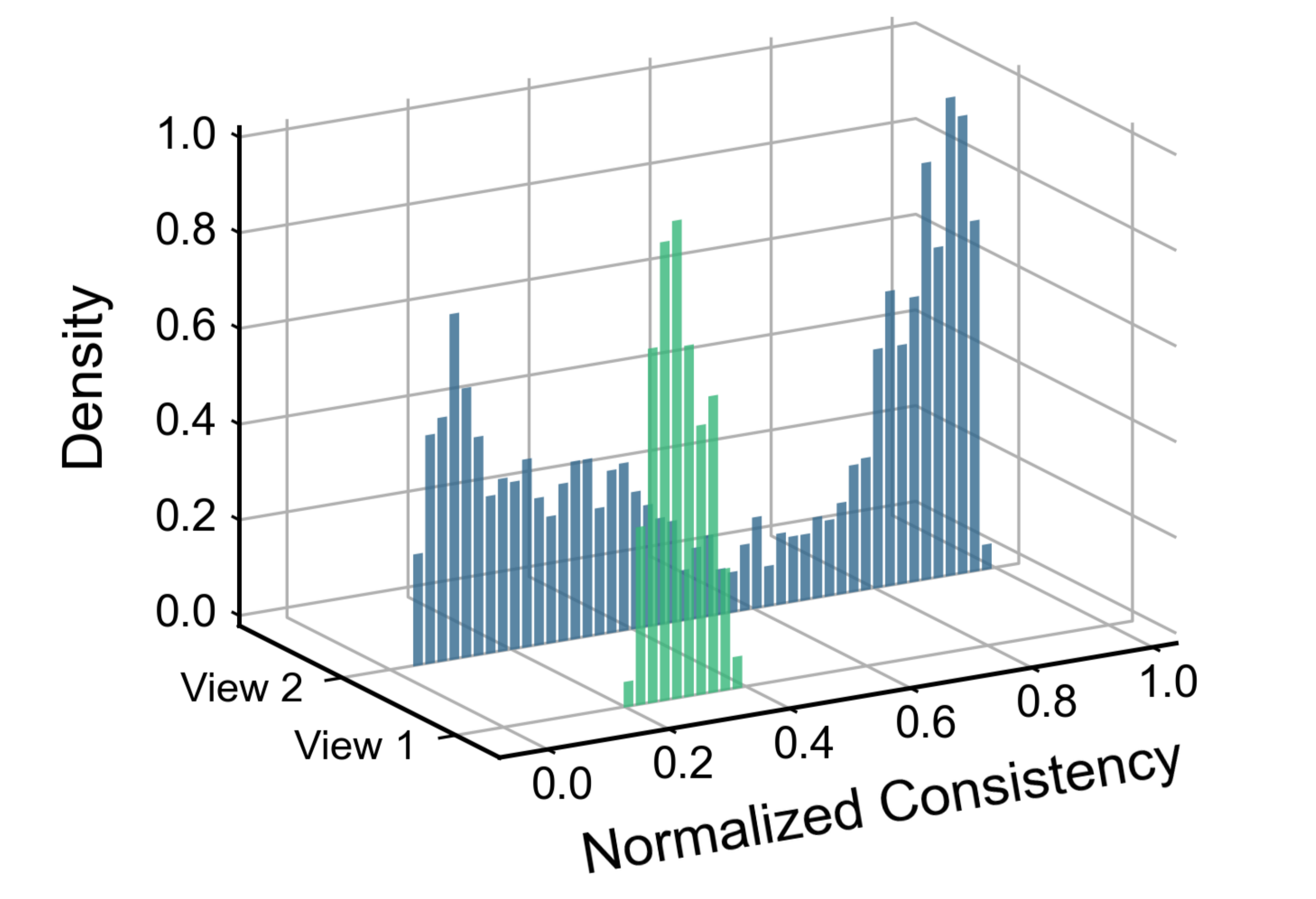}%
	\caption{The evidence-label consistency distribution computed across all samples from the two views of the UCI dataset. View 2 shows a bimodal distribution centered on `0' and `1', clearly dividing samples into high-consistency and low-consistency groups, while View 1 exhibits a unimodal distribution. This suggests significant differences in the ability to distinguish noisy samples between different views.}
	\label{fig:diffview}
\end{figure}

After obtaining the evidence-label consistency of all samples in different views (e.g., Figure \ref{fig:diffview}), we note that the ability to distinguish noisy samples varies due to the different learning difficulties of each view \cite{viewDiff}. A larger variance in consistency within a view indicates stronger capability of that view to separate noisy samples. Therefore, we use the variance of evidence-label consistency across all samples within each view as a weight to combine the consistencies:
\begin{equation}
	\label{viewfusion}
	\mathcal{H}\left(\bm{x}_n\right)=\sum_{v=1}^V \eta^v\left(\mathcal{H}^v\left(\bm{x}_n^v\right)\right), \  \ \eta^v=\frac{\operatorname{Var}\left(\bm{\mathcal{H}}^v\right)}{\sum_{v=1}^V \operatorname{Var}\left(\bm{\mathcal{H}}^v\right)},
\end{equation}
where $\operatorname{Var}\left(\bm{\mathcal{H}}^v\right)=\frac{1}{N_{train}-1} \sum_{n=1}^{N_{train}} (\mathcal{H}^v (\boldsymbol{x}_n^v)-{\bar{\mathcal{H}}}^v)^2$, ${\bar{\mathcal{H}}}^v$ denotes the mean value of the evidence-label consistency of all instances in view $v$.
A higher consistency $\mathcal{H}\left(\bm{x}_n\right)$ value indicates a greater discrepancy between the sample's label and the predicted evidence distribution (i.e., a higher probability of being mislabeled). We ultimately classify samples with consistency values exceeding a predefined threshold $\epsilon$ as noise samples and record their indices:
\begin{equation}
	\label{separation}
	\mathcal{N}_{noise} = \{n\ |\  n\in\{1,\dots, N_{train} \} \  \text{and} \ \mathcal{H}(\bm{x}_n )> \epsilon \}.
\end{equation}

\begin{algorithm}[tb]
	\caption{The Pseudocode of TMNR$^\textbf{2}$.}
	\label{alg:algorithm2}
	\textbf{/*TRAIN*/} \\
	\textbf{Input}: Noisy training dataset, hyperparameter $\beta$, $\gamma$. \\
	\textbf{Output}: Parameters of model.
	\begin{algorithmic}[1] 
		\STATE  Initialize parameters $\bm{\theta}$ of the evidence neural network.
		\STATE  Initialize all correlation matrices $\bm{T}$ as unit matrices.
		\STATE  $\mathit{WarmUp}$ $\bm{\theta}$ for $\mathbf{w}$ epochs.           
		\FOR{training iteration $\mathbf{i} = \mathbf{w}: \text{MaxEpoch}$}
		\FOR{view $v=1:V$}
		\STATE Obtain clean evidence $\bm{e}^v_n$ with $f^v(\bm{x}^v_n; \bm{\theta})$;
		\STATE Obtain $\tilde{\bm{e}}^v_n$ and $\tilde{\bm{\alpha}}^v_n$ through Eq.\eqref{transferE};
		\ENDFOR
		\STATE Aggregation to obtain $\tilde{\bm{\alpha}}_n$ by Eq.\eqref{aggregation};

		\STATE \underline{Calculate loss with Eq.\eqref{overallLoss} for clean samples;}
		\STATE \underline{Calculate loss with Eq.\eqref{overallLossnew} for $\mathcal{N}_{noise}$;}
		\STATE  Update the parameters;
		\STATE  \underline{Correct $\bm{T}$ to satisfy Eq.\eqref{defineT};}
		\IF   {$(\mathbf{i}-\mathbf{w}) \% 10\ \text{is}\ 0$}
		\STATE  \underline{Separate noisy samples according to Eq.\eqref{separation};}
		\STATE \underline{Mixup the noisy and clean samples by Eq.\eqref{mix2};}
		\STATE \underline{Update training dataset with mixed samples;}
		\ENDIF

		\ENDFOR
	\end{algorithmic}
	\textbf{/*TEST*/} \\
	Calculate the clean joint $Dir(\bm{p}|\bm{\alpha})$ and the corresponding uncertainty $u$ by $f(\cdot;\bm{\theta})$.
\end{algorithm}

\subsection{Noise Sample Re-refining}
\label{rerefinement}
At the end of each epoch, after identifying the noisy samples, TMNR$^\textbf{2}$ utilizes the model's predictions to generate guessed pseudo-labels for these samples. It then performs \textit{Mixup} augmentation by combining these noisy instances with an equal number of instances selected from the clean set, replacing the original noisy samples with the resulting mixed samples. The updated dataset containing mixed labels, directly supervises the potential clean evidence distribution of the instances, thereby reducing the difficulty of collaborative training for the model.

When assigning pseudo-labels to noisy samples, similar to the previous subsection, the guessed labels are derived from the predicted evidence distributions of the samples and their neighbors. By aggregating the evidence weighted by the similarity between each instance and its neighbors, and further fusing them using the view fusion weights, the class with the largest aggregated evidence is defined as the pseudo-label $\hat{y}$ for the noisy sample as follows:
\begin{equation}
	\bm{e}_n^{\mathrm{agg}}=\sum_{v=1}^{V}\eta^v \Big( \bm{e}_n^v + \sum_{k=1}^K s_{n m_k}^v \bm{e}_{m_k}^v\Big) , \ m_k \in \mathcal{N}(\bm{x}_n^v, K),
\end{equation}
\begin{equation}
	\hat{y}_n= \mathop{\arg \max}_{c \in \{1, \dots, C\}} \left[\boldsymbol{e}_n^{\mathrm{agg}}\right]_c,
\end{equation}
where $\eta^v$ represents the fusion weight derived from the consistency variance of each view as shown in Eq.\eqref{viewfusion}, and $[\bm{e}_n^{\textrm{agg}}]_c$ denotes the evidence mass of the $c$-th category in $\bm{e}_n^{\textrm{agg}}$. After that, an equal number of clean samples are selected from the training set excluding the noise sample set and \textit{Mixup}\cite{mixup}  them. This linear mixing not only extends the overall distribution of the training data but also reduces the impact of incorrectly guessed pseudo-labels. For a selected clean sample $\{\{\bm{x}_j^v\}_{v=1}^V,\tilde{y}_j \}$ and a noise sample $\{\{\bm{x}_i^v\}_{v=1}^V,\hat{y}_i \}$, the mixed sample $\{\{{\bm{x}^v}^\prime\}_{v=1}^V,{y}^\prime \}$ is computed as follows:
\begin{align}
	\lambda                                            & \sim \operatorname{Beta}(0.3, 0.3),                                             \\
	\label{mixed}
	\lambda^{\prime}                                   & =\max (\lambda, 1-\lambda),                                                     \\
	\label{mix2}
	{\bm{x}^v}^{\prime}  =\lambda^{\prime} \bm{x}^v_i+ & \left(1 -\lambda^{\prime}\right) \bm{x}_j^v  ,\ \ \ \forall v\in\{1,\dots, V\}, \\
	y^{\prime}                                         & =\lambda^{\prime} \hat{y}_i+\left(1-\lambda^{\prime}\right) \tilde{y}_j,
\end{align}
where Eq.\eqref{mixed} ensures that the mixed sample $\bm{x}^\prime$ is closer to the noise samples $\bm{x}_i$ rather than $\bm{x}_j$, which helps avoid possible loss of training data. Finally we use the mixed sample set to replace the noise samples for the overall optimization pipeline.

\subsection{Optimization}
In TMNR, the optimization process relies entirely on the overall loss function driven by noisy labels, as shown in Eq.(\ref{overallLoss}). However, the strong coupling between $f(\theta)$ and $\bm{T}$, combined with the lack of supervision over the potential clean evidence distribution, presents significant challenges for optimizing the entire model.
To overcome these challenges, TMNR$^\textbf{2}$ identifies noisy samples and performs refined mixing, introducing guessed pseudo-labels to directly impose constraints on the intermediate variable $\bm{e}$ of the two mappings. Thereby enhancing the model's generalization ability in scenarios with noisy labels.

Specifically, to achieve the initial convergence of the algorithm, we first ``warm up” the model for a few epochs by training all the data using only the evidence-level generalized maximum likelihood loss in Eq.\eqref{lossace}. Subsequently, we unfreeze the noise correlation matrix to learn the generation patterns of noisy labels and begin splitting the dataset into noisy and clean samples, either simultaneously or before the model overfits. When optimizing the potential clean evidence distribution in TMNR$^\textbf{2}$ according to the pseudo-labels, we introduce the bounded $l_2$ loss (i.e., MSE loss)\cite{robustloss}. In contrast to the unbounded cross-entropy loss, this loss ensures that the clean evidence space is mildly constrained while avoiding excessive penalties caused by potential inaccuracies in the pseudo-labels. For a mixed sample $\{\{{\bm{x}^v}^\prime\}_{v=1}^V, {y}^\prime\} $, the loss is defined as follows:
\begin{equation}
	\mathcal{L}_{mse} = \frac{1}{V}\sum_{v=1}^V \parallel  \mathbf{y}^\prime-{\bm{p}^v}^\prime \parallel^2,
\end{equation}
where $\bm{p}^v= \bm{\alpha}^v / S$ is the clean class probability for each view. Accordingly, the overall loss in Eq.\eqref{overallLoss} is updated as:
\begin{equation}
	\label{overallLossnew}
	\mathcal{L}_{all}=\mathcal{L}(\tilde{\bm{\alpha}})+
	\sum_{v=1}^V \left[ \mathcal{L}\left(\tilde{\bm{\alpha}}^v\right)
		+ \beta \left(  \mathcal{M}(\tilde{\bm{\alpha}}^v) \right)
		\right] + \mathcal{L}_{mse} + \gamma \mathcal{L}_{con} .
\end{equation}

To ensure the stability of training and full exploration of the parameter space, after the initial dataset partition, we re-identify mislabeled samples based on the threshold $\epsilon$ every 10 epochs and add them to the noisy sample set. The overall optimization process of TMNR$^\textbf{2}$ is illustrated in Algorithm \ref{alg:algorithm2}.

\section{Experiments}
\label{experiments}
In this section, we conduct experiments on multiple scenarios to comprehensively evaluate the proposed method.

\begin{table}[tb]
	\caption{Summary of the baseline methods. \ding{52} denotes the corresponding information is used.}
	\label{tab:methods}
	\centering
	\tabcolsep=0.035\linewidth
	\begin{tabular}{c|c|c|c}
		\toprule
		$\textbf{Methods}$           & \textbf{Trusted} & \textbf{Multi-view} & \textbf{Noise Refining} \\
		\midrule
		\textbf{MCDO}\cite{MCDO}     & \ding{52}        & \ding{56}           & \ding{56}               \\
		\textbf{IEDL}\cite{IEDL}     & \ding{52}        & \ding{56}           & \ding{56}               \\
		\textbf{FC}\cite{FC}         & \ding{56}        & \ding{56}           & \ding{52}               \\
		\textbf{ILFC}\cite{CSIDN}    & \ding{56}        & \ding{56}           & \ding{52}               \\
		\textbf{TNLPAD}\cite{TNLPAD} & \ding{56}        & \ding{56}           & \ding{52}               \\
		\textbf{DCCAE}\cite{DCCAE}   & \ding{56}        & \ding{52}           & \ding{56}               \\
		\textbf{DCP}\cite{DCP}       & \ding{56}        & \ding{52}           & \ding{56}               \\
		\textbf{ETMC}\cite{ETMC}     & \ding{52}        & \ding{52}           & \ding{56}               \\
		\textbf{PDF}\cite{PDF}       & \ding{52}        & \ding{52}           & \ding{56}               \\
		\textbf{ECML}\cite{ECML}     & \ding{52}        & \ding{52}           & \ding{56}               \\
		\cmidrule(lr{1pt}){1-4}
		\textbf{TMNR}\cite{TMNR}     & \ding{52}        & \ding{52}           & \ding{52}               \\
		\textbf{TMNR$^\textbf{2}$}   & \ding{52}        & \ding{52}           & \ding{52}               \\
		\bottomrule
	\end{tabular}
\end{table}

\renewcommand{\arraystretch}{0.8} 
\begin{table*}[htpb]
	\caption{Classification accuracy($\%$) of the proposed and baseline methods on the datasets with different proportions of Instance-Dependent Noise. The `Noise' column shows the percentage of noisy labelled instances, where $0\%$ denote clean datasets. The best and the second best results are highlighted by \textbf{boldface} and \underline{underlined} respectively.}
	\centering
	\tabcolsep=0.016\linewidth
	\begin{tabular}{c|c|cccccc}
		\toprule
		\multirow{2}{*}{\textbf{Datasets}} & \multirow{2}{*}{\textbf{Methods}} & \multicolumn{6}{c}{\textbf{Noise}}                                                                            \\
		 &        & 0\%      & 10\%          & 20\%     & 30\%              & 40\%      & 50\%            \\
		\midrule
		\multirow{13}{*}{$\textbf{UCI}$}
		& MCDO                              & \underline{97.50$\pm$0.72}         & 95.50$\pm$0.61             & 95.40$\pm$0.82             & 92.30$\pm$0.80             & 90.15$\pm$1.44             & 83.65$\pm$1.98             \\
		& IEDL                              & \textbf{97.70$\pm$0.46}            & 95.25$\pm$1.25             & 95.50$\pm$0.75             & 92.50$\pm$1.74             & 90.30$\pm$1.99             & 85.85$\pm$1.17             \\
		& FC                                & 97.05$\pm$0.19                     & 96.20$\pm$0.58             & 95.35$\pm$1.06             & 91.90$\pm$1.21             & 89.90$\pm$1.58             & 83.55$\pm$2.93             \\
		& ILFC                              & 95.50$\pm$0.76                     & 95.45$\pm$0.82             & 95.15$\pm$0.46             & 93.85$\pm$1.51             & 90.95$\pm$1.87             & 84.85$\pm$1.23             \\
		& TNLPAD                            & 96.70$\pm$0.46                     & \underline{96.30$\pm$0.19} & 95.70$\pm$0.37             & 93.85$\pm$0.34             & 91.10$\pm$0.20             & \underline{89.60$\pm$0.82} \\
		& DCCAE                             & 85.75$\pm$0.32                     & 85.50$\pm$0.70             & 85.00$\pm$0.37             & 84.50$\pm$0.98             & 84.50$\pm$0.78             & 81.75$\pm$1.08             \\
		& DCP                               & 96.15$\pm$0.49                     & 95.55$\pm$0.54             & 95.25$\pm$0.87             & 92.40$\pm$1.24             & 90.30$\pm$1.08             & 83.75$\pm$1.64             \\
		& ETMC                              & 96.20$\pm$0.73                     & 95.50$\pm$0.33             & 95.35$\pm$0.49             & 93.15$\pm$1.49             & 91.65$\pm$1.54             & 84.10$\pm$1.83             \\
		& PDF                               & 96.40$\pm$0.72                     & 89.60$\pm$1.71             & 80.00$\pm$2.82             & 74.70$\pm$1.45             & 70.70$\pm$2.68             & 63.95$\pm$2.71             \\
		& ECML                              & 97.05$\pm$0.48                     & 95.85$\pm$0.19             & 95.35$\pm$0.66             & 92.85$\pm$1.42             & 92.35$\pm$1.37             & 86.15$\pm$1.62             \\
\cmidrule(l{0.5pt}){2-8}
	& TMNR                              & 96.90$\pm$0.65                     & 95.95$\pm$0.37             & \underline{95.90$\pm$0.84} & \underline{94.00$\pm$1.46} & \underline{94.65$\pm$1.35} & 88.90$\pm$0.49             \\
	& TMNR$^\textbf{2}$                 & 96.20$\pm$0.42                     & \textbf{96.95$\pm$0.70}    & \textbf{96.35$\pm$0.95}    & \textbf{95.25$\pm$0.31}    & \textbf{95.15$\pm$0.92}    & \textbf{94.55$\pm$1.70}    \\
	\midrule
	\multirow{13}{*}{$\textbf{PIE}$}
	& MCDO                              & 89.71$\pm$3.15                     & 77.50$\pm$3.37             & 64.71$\pm$1.40             & 55.44$\pm$3.28             & 46.32$\pm$2.03             & 38.68$\pm$3.50             \\
	& IEDL                              & \underline{90.88$\pm$3.31}         & 85.74$\pm$4.50             & 80.29$\pm$3.58             & 69.41$\pm$2.73             & 65.29$\pm$4.57             & 52.94$\pm$2.88             \\
	& FC                                & 71.03$\pm$5.02                     & 55.00$\pm$3.87             & 44.85$\pm$4.05             & 33.97$\pm$1.94             & 32.35$\pm$5.29             & 25.44$\pm$2.57             \\
	& ILFC                              & 73.24$\pm$3.27                     & 66.47$\pm$4.12             & 61.18$\pm$3.21             & 58.97$\pm$4.10             & 51.91$\pm$5.29             & 43.97$\pm$4.23             \\
	& TNLPAD                            & 83.97$\pm$0.55                     & 81.91$\pm$0.36             & 76.91$\pm$0.75             & 63.68$\pm$1.10             & 53.82$\pm$0.86             & 54.26$\pm$1.08             \\
	& DCCAE                             & 53.38$\pm$1.82                     & 47.06$\pm$0.76             & 47.79$\pm$1.87             & 36.03$\pm$0.67             & 33.82$\pm$1.19             & 30.88$\pm$1.38             \\
	& DCP                               & 87.21$\pm$2.48                     & 82.65$\pm$3.21             & 75.74$\pm$1.84             & 65.44$\pm$2.49             & 58.53$\pm$2.96             & 50.88$\pm$2.82             \\
	& ETMC                              & 87.06$\pm$2.89                     & 83.38$\pm$1.28             & 77.21$\pm$1.04             & 63.97$\pm$1.86             & 61.76$\pm$2.33             & 51.32$\pm$3.40             \\
	& PDF                               & 86.62$\pm$1.50                     & 62.79$\pm$4.64             & 47.50$\pm$4.66             & 36.62$\pm$4.32             & 27.94$\pm$2.83             & 26.62$\pm$1.57             \\
	& ECML                              & 89.85$\pm$2.66                     & 83.09$\pm$3.75             & 76.47$\pm$1.54             & 70.44$\pm$4.04             & 63.97$\pm$4.53             & 55.59$\pm$2.30             \\
	\cmidrule(l{0.5pt}){2-8}
	& TMNR                              & 89.56$\pm$1.89                     & \underline{86.47$\pm$1.97} & \underline{83.24$\pm$1.70} & \underline{73.24$\pm$2.08} & \underline{71.91$\pm$2.33} & \underline{59.85$\pm$2.89} \\
	& TMNR$^\textbf{2}$                 & \textbf{92.35$\pm$1.08}            & \textbf{91.18$\pm$1.17}    & \textbf{89.41$\pm$4.11}    & \textbf{84.26$\pm$2.08}    & \textbf{77.35$\pm$3.11}    & \textbf{67.35$\pm$5.43}    \\
	\midrule
	\multirow{13}{*}{$\textbf{BBC}$}
	& MCDO                              & 93.28$\pm$1.79                     & 89.34$\pm$1.88             & 85.99$\pm$2.86             & 74.45$\pm$4.05             & 69.64$\pm$2.72             & 56.64$\pm$3.59             \\
	& IEDL                              & 92.55$\pm$2.04                     & \underline{90.37$\pm$1.81} & 86.86$\pm$1.87             & 81.02$\pm$3.56             & 71.09$\pm$1.07             & 58.98$\pm$4.52             \\
	& FC                                & 92.12$\pm$3.01                     & 89.34$\pm$1.57             & 85.11$\pm$2.47             & 73.87$\pm$5.66             & 70.51$\pm$1.76             & 56.93$\pm$2.44             \\
	& ILFC                              & 92.55$\pm$1.87                     & 88.32$\pm$1.34             & 85.99$\pm$2.02             & 77.23$\pm$2.98             & 72.70$\pm$3.88             & 57.23$\pm$3.90             \\
	& TNLPAD                            & 90.07$\pm$0.36                     & 87.30$\pm$0.36             & 85.69$\pm$0.36             & 71.68$\pm$0.55             & 69.49$\pm$0.55             & 63.36$\pm$0.97             \\
	& DCCAE                             & 91.97$\pm$2.48                     & 88.18$\pm$1.54             & 85.40$\pm$2.97             & 77.96$\pm$2.36             & 71.09$\pm$2.51             & 56.20$\pm$3.87             \\
	& DCP                               & 93.14$\pm$1.92                     & 88.32$\pm$1.41             & 86.13$\pm$2.38             & 77.81$\pm$2.49             & 70.80$\pm$3.01             & 56.20$\pm$3.76             \\
	& ETMC                              & \underline{93.58$\pm$1.42}         & 89.93$\pm$1.56             & 86.86$\pm$2.73             & 80.73$\pm$2.47             & 72.85$\pm$3.36             & 57.23$\pm$4.30             \\
	& PDF       & 92.99$\pm$2.64   & 85.99$\pm$0.97     & 79.27$\pm$2.87             & 68.47$\pm$3.65    & 63.50$\pm$3.32     & 54.16$\pm$2.87             \\
	& ECML                              & 91.82$\pm$1.93                     & 88.18$\pm$1.17             & 85.11$\pm$2.87             & 74.16$\pm$1.27             & 72.85$\pm$3.96             & 58.83$\pm$3.05             \\
	\cmidrule(l{0.5pt}){2-8}
	& TMNR                              & 93.58$\pm$1.56                     & 90.07$\pm$1.53             & \underline{87.45$\pm$2.86} & \underline{82.04$\pm$2.98} & \underline{75.91$\pm$3.44} & \underline{63.94$\pm$4.43} \\
	& TMNR$^\textbf{2}$                 & \textbf{94.16$\pm$1.28}            & \textbf{92.26$\pm$3.18}    & \textbf{91.09$\pm$3.28}    & \textbf{82.92$\pm$4.24}    & \textbf{78.69$\pm$4.13}    & \textbf{64.96$\pm$1.98}    \\
	\midrule
	\multirow{13}{*}{$\textbf{Caltech}$}
	& MCDO                              & 71.38$\pm$5.06                     & 68.66$\pm$5.02             & 58.41$\pm$2.43             & 54.60$\pm$3.58             & 48.12$\pm$5.99             & 43.89$\pm$6.41             \\
	& IEDL                              & 92.35$\pm$1.46                     & 91.05$\pm$1.26             & 86.82$\pm$2.22             & 83.01$\pm$0.90             & 71.92$\pm$1.95             & 59.04$\pm$1.84             \\
	& FC                                & 64.64$\pm$5.57                     & 57.28$\pm$3.46             & 42.34$\pm$5.38             & 52.97$\pm$2.54             & 46.57$\pm$4.77             & 35.98$\pm$4.78             \\
	& ILFC                              & 85.24$\pm$1.90                     & 81.92$\pm$2.39             & 77.74$\pm$2.80             & 73.26$\pm$2.11             & 70.17$\pm$3.98             & 63.28$\pm$3.41             \\
	& TNLPAD                            & \textbf{93.41$\pm$0.38}            & \textbf{92.55$\pm$0.25}    & \underline{88.37$\pm$0.94} & 85.27$\pm$0.36             & 78.41$\pm$0.86             & 69.29$\pm$0.48             \\
	& DCCAE                             & 88.03$\pm$0.75                     & 86.19$\pm$0.98             & 83.47$\pm$1.28             & 82.85$\pm$0.93             & 75.94$\pm$1.89             & 61.09$\pm$1.46             \\
	& DCP                               & 91.93$\pm$1.39                     & 90.74$\pm$1.22             & 87.02$\pm$2.31             & 84.98$\pm$1.01             & 76.90$\pm$1.55             & 65.29$\pm$2.52             \\
	& ETMC                              & 92.59$\pm$0.86                     & 90.34$\pm$1.32             & 87.74$\pm$1.24             & 86.07$\pm$1.10             & 78.91$\pm$0.90             & 68.79$\pm$2.02             \\
	& PDF                               & 67.82$\pm$2.28                     & 60.46$\pm$2.78             & 51.00$\pm$2.22             & 50.08$\pm$1.49             & 48.16$\pm$3.31             & 47.99$\pm$1.38             \\
	& ECML                              & 91.13$\pm$1.61                     & 91.38$\pm$1.59             & 87.12$\pm$0.96             & 86.11$\pm$0.49             & 77.82$\pm$0.68             & 68.91$\pm$1.96             \\
	\cmidrule(l{0.5pt}){2-8}
	& TMNR                              & 91.84$\pm$1.08                     & 91.09$\pm$0.59             & 87.78$\pm$1.26             & \underline{86.82$\pm$0.83} & \underline{81.59$\pm$0.96} & \underline{72.89$\pm$1.97} \\
	& TMNR$^\textbf{2}$                 & \underline{92.85$\pm$1.86}         & \underline{91.55$\pm$3.04} & \textbf{91.00$\pm$0.65}    & \textbf{90.88$\pm$2.99}    & \textbf{88.62$\pm$0.78}    & \textbf{82.34$\pm$2.16}    \\
	\midrule
	\multirow{13}{*}{$\textbf{Leaves}$}
	& MCDO                              & 79.94$\pm$3.92                     & 74.12$\pm$5.71             & 68.19$\pm$1.74             & 59.75$\pm$6.20             & 54.44$\pm$3.05             & 44.06$\pm$2.58             \\
	& IEDL                              & \underline{95.50$\pm$0.32}         & 92.25$\pm$0.92             & 90.06$\pm$0.64             & 84.19$\pm$0.64             & \underline{82.06$\pm$1.71} & 75.81$\pm$1.67             \\
	& FC                                & 81.94$\pm$2.76                     & 73.06$\pm$2.83             & 70.62$\pm$4.31             & 59.50$\pm$4.93             & 56.00$\pm$3.08             & 39.88$\pm$2.18             \\
	& ILFC                              & 83.81$\pm$1.82                     & 76.88$\pm$2.37             & 73.12$\pm$3.53             & 66.25$\pm$2.65             & 61.94$\pm$3.73             & 42.50$\pm$2.46             \\
	& TNLPAD                            & \textbf{96.06$\pm$0.23}            & \textbf{92.75$\pm$0.28}    & 90.44$\pm$0.23             & \underline{85.50$\pm$0.40} & 81.50$\pm$0.85             & \underline{76.56$\pm$0.54} \\
	& DCCAE                             & 63.50$\pm$0.75                     & 62.19$\pm$0.97             & 61.25$\pm$1.24             & 58.13$\pm$0.76             & 55.13$\pm$1.87             & 50.31$\pm$1.89             \\
	& DCP                               & 94.06$\pm$1.37                     & 90.88$\pm$1.41             & 86.00$\pm$2.61             & 78.62$\pm$1.77             & 69.25$\pm$2.29             & 60.81$\pm$3.24             \\
	& ETMC                              & 91.44$\pm$2.27                     & 90.25$\pm$1.16             & 88.75$\pm$2.02             & 83.81$\pm$1.83             & 79.19$\pm$1.74             & 72.06$\pm$2.96             \\
	& PDF                               & 75.88$\pm$3.06                     & 65.25$\pm$4.33             & 53.81$\pm$2.94             & 40.62$\pm$3.29             & 36.62$\pm$3.96             & 29.88$\pm$0.76             \\
	& ECML                              & 92.62$\pm$1.80                     & 89.75$\pm$0.86             & 90.44$\pm$1.18             & 85.00$\pm$1.34             & 80.56$\pm$2.81             & 74.19$\pm$2.92             \\
	\cmidrule(l{0.5pt}){2-8}
	& TMNR                              & 95.00$\pm$1.03                     & \underline{92.50$\pm$0.88} & \underline{90.62$\pm$2.16} & 85.00$\pm$1.75             & 81.25$\pm$2.16             & 75.31$\pm$2.43             \\
	& TMNR$^\textbf{2}$                 & 94.44$\pm$0.60                     & 91.94$\pm$1.23             & \textbf{91.25$\pm$0.56}    & \textbf{86.19$\pm$2.59}    & \textbf{83.75$\pm$4.49}    & \textbf{77.50$\pm$4.10}    \\
	\midrule
	\multirow{13}{*}{$\textbf{CUB}$}
	& MCDO                              & 90.50$\pm$0.97                     & 90.17$\pm$0.85             & 88.33$\pm$1.55             & 87.83$\pm$2.61             & 77.67$\pm$3.70             & 72.17$\pm$7.31             \\
	& IEDL                              & \textbf{92.50$\pm$2.26}            & 91.17$\pm$0.62             & \underline{89.83$\pm$2.21} & 86.17$\pm$1.72             & 83.67$\pm$1.72             & 82.00$\pm$3.22             \\
	& FC                                & 91.17$\pm$2.00                     & 91.00$\pm$1.86             & 89.17$\pm$2.26             & 86.17$\pm$3.44             & 76.83$\pm$3.14             & 75.67$\pm$6.00             \\
	& ILFC                              & 91.67$\pm$1.64                     & 90.50$\pm$0.74             & 89.33$\pm$1.75             & 87.50$\pm$2.06             & 78.17$\pm$2.59             & 76.67$\pm$3.21             \\
	& TNLPAD                            & \underline{92.00$\pm$0.67}         & \underline{91.17$\pm$0.33} & 89.67$\pm$1.43             & \textbf{89.67$\pm$1.13}    & 83.67$\pm$0.97             & 82.33$\pm$3.23             \\
	& DCCAE                             & 90.67$\pm$1.83                     & 89.17$\pm$0.97             & 87.67$\pm$1.34             & 83.33$\pm$0.59             & 81.67$\pm$2.21             & 78.33$\pm$1.64             \\
	& DCP                               & 91.33$\pm$1.74                     & 90.67$\pm$2.41             & 88.50$\pm$1.85             & 86.67$\pm$2.51             & 79.00$\pm$2.71             & 74.83$\pm$3.28             \\
	& ETMC                              & 90.67$\pm$2.00                     & 88.83$\pm$1.55             & 89.33$\pm$2.60             & 86.17$\pm$2.30             & 82.83$\pm$3.01             & 82.33$\pm$2.26             \\
	& PDF                               & 92.00$\pm$2.86                     & 83.00$\pm$1.55             & 75.67$\pm$2.38             & 66.17$\pm$3.89             & 59.17$\pm$4.12             & 47.83$\pm$2.82             \\
	& ECML                              & 92.00$\pm$2.39                     & 91.17$\pm$1.63             & 88.33$\pm$1.25             & 87.00$\pm$0.91             & 82.00$\pm$1.83             & 81.67$\pm$1.55             \\
	\cmidrule(l{0.5pt}){2-8}
	& TMNR                              & 91.83$\pm$2.20                     & 91.17$\pm$2.21             & 89.67$\pm$2.92             & 88.50$\pm$4.23             & \underline{84.50$\pm$1.55} & \underline{82.67$\pm$3.51} \\
	& TMNR$^\textbf{2}$                 & 91.67$\pm$2.64                     & \textbf{91.83$\pm$2.08}    & \textbf{89.83$\pm$2.15}    & \underline{89.67$\pm$3.14} & \textbf{84.83$\pm$2.39}    & \textbf{83.50$\pm$2.76}    \\

		\bottomrule
	\end{tabular}
	\label{tab:accTest}
\end{table*}
\renewcommand{\arraystretch}{1.0} 

\subsection{Experimental Setup}
\subsubsection{\textbf{Datasets}}  
Our experiments utilize 6 multi-view datasets with pre-extracted features and an end-to-end image-text dataset.
(1) \textbf{Pre-extracted feature datasets}:
\textbf{UCI} contains features for handwritten numerals (`$0$'-`$9$').
The average of pixels in 240 windows, 47 Zernike moments, and 6 morphological features are used as 3 views.
\textbf{PIE} consists of 680 face images from 68 experimenters. We extracted 3 views from it: intensity, LBP and Gabor.
\textbf{BBC} includes 685 documents from BBC News that can be categorised into 5 categories and are depicted by 4 views.
\textbf{Caltech} contains 8677 images from 101 categories, extracting features as different views with 6 different methods: Gabor, Wavelet Moments, CENTRIST, HOG, GIST, and LBP. we chose the first 20 categories.
\textbf{Leaves} consists of 1600 leaf samples from 100 plant species. We extracted shape descriptors, fine-scale edges, and texture histograms as 3 views.
\textbf{CUB} consists of 11788 instances associated with text descriptions of 200 different categories of birds. For this study, we use the first 10 categories, extracting image features with GoogleNet and corresponding text features using doc2vec.
(2) \textbf{End-to-end dataset}:  
\textbf{UPMC Food101}\cite{food101} contains 86,796 food images and their corresponding textual descriptions collected from websites, spanning 101 food categories.

\begin{figure*}[!t]
	\centering
	\subfloat[UCI]{\includegraphics[width=2.2in]{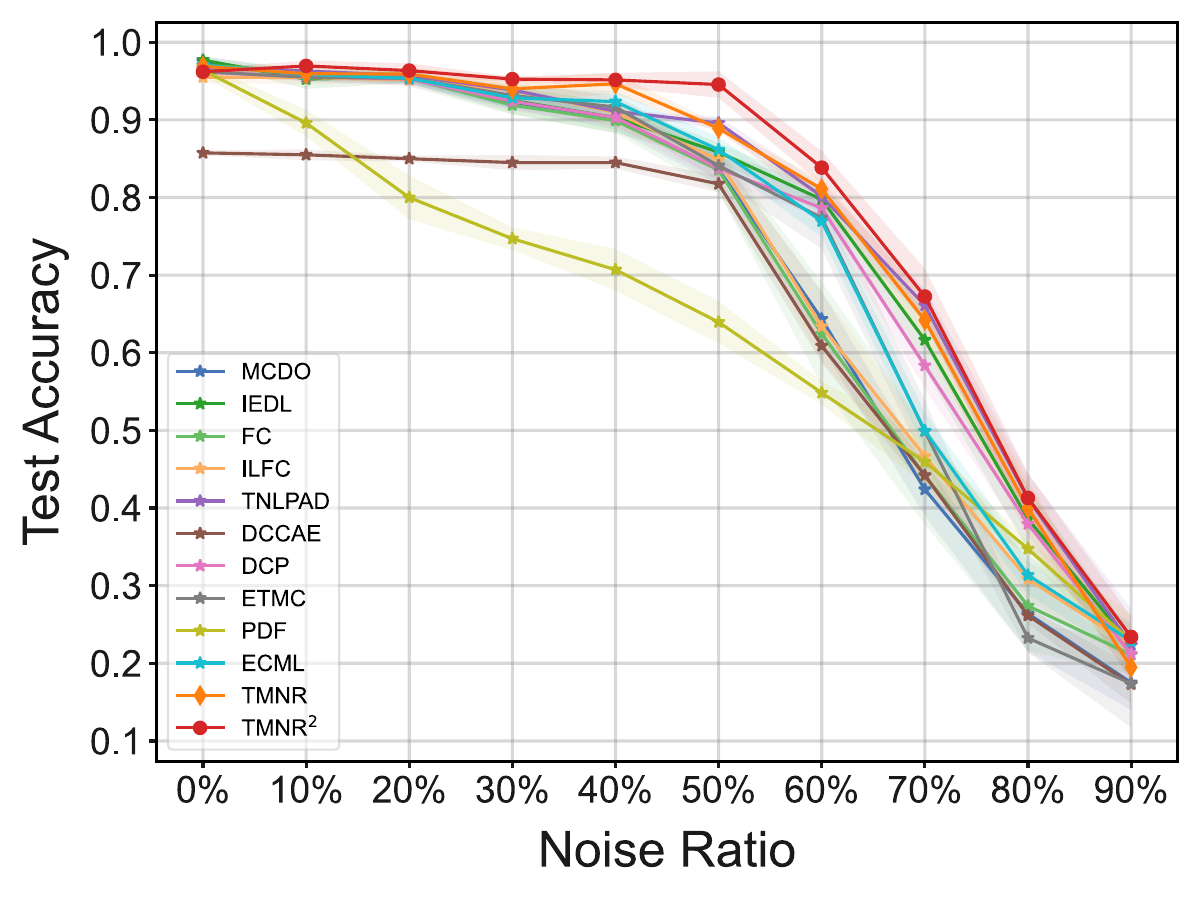}%
		\label{fig_first_case}}
	\hfil
	\subfloat[PIE]{\includegraphics[width=2.2in]{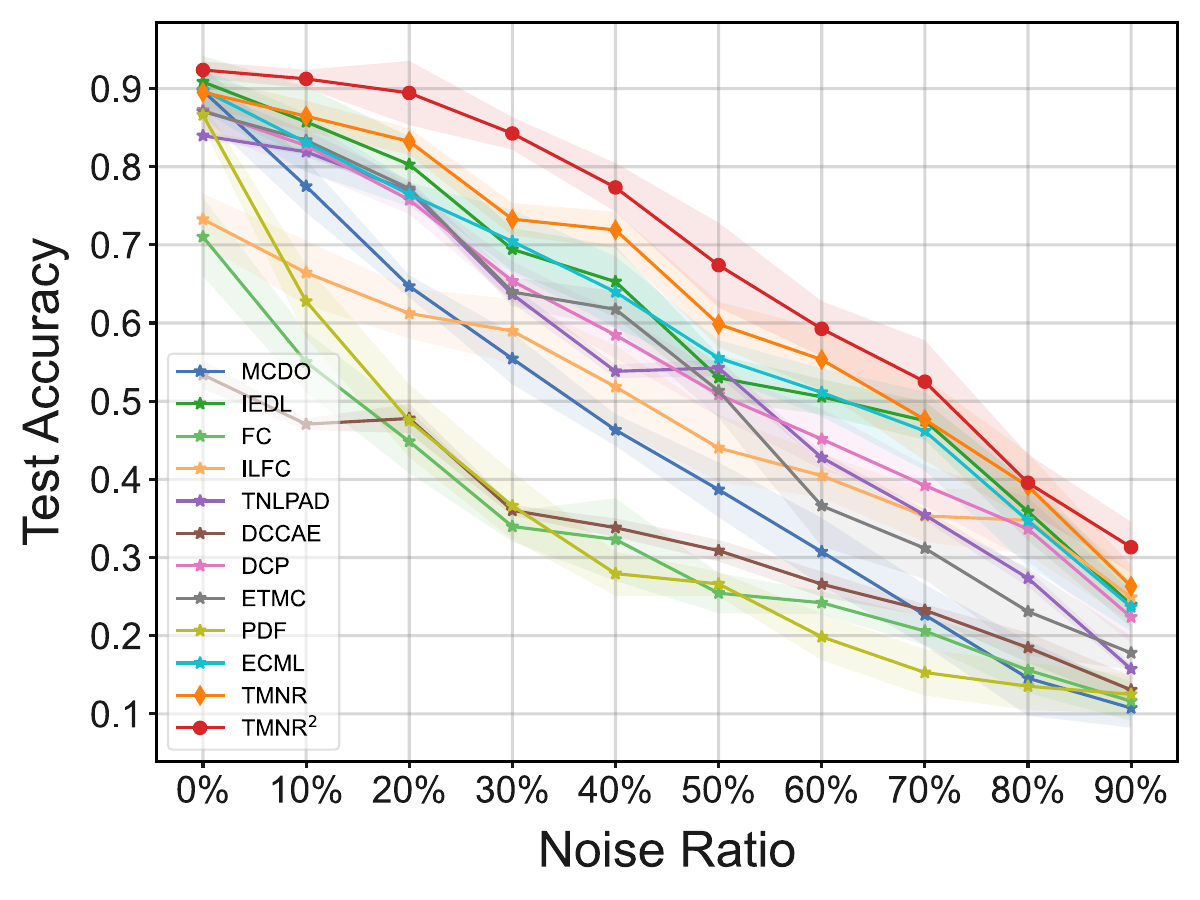}%
		\label{fig_second_case}}
	\hfil
	\subfloat[BBC]{\includegraphics[width=2.2in]{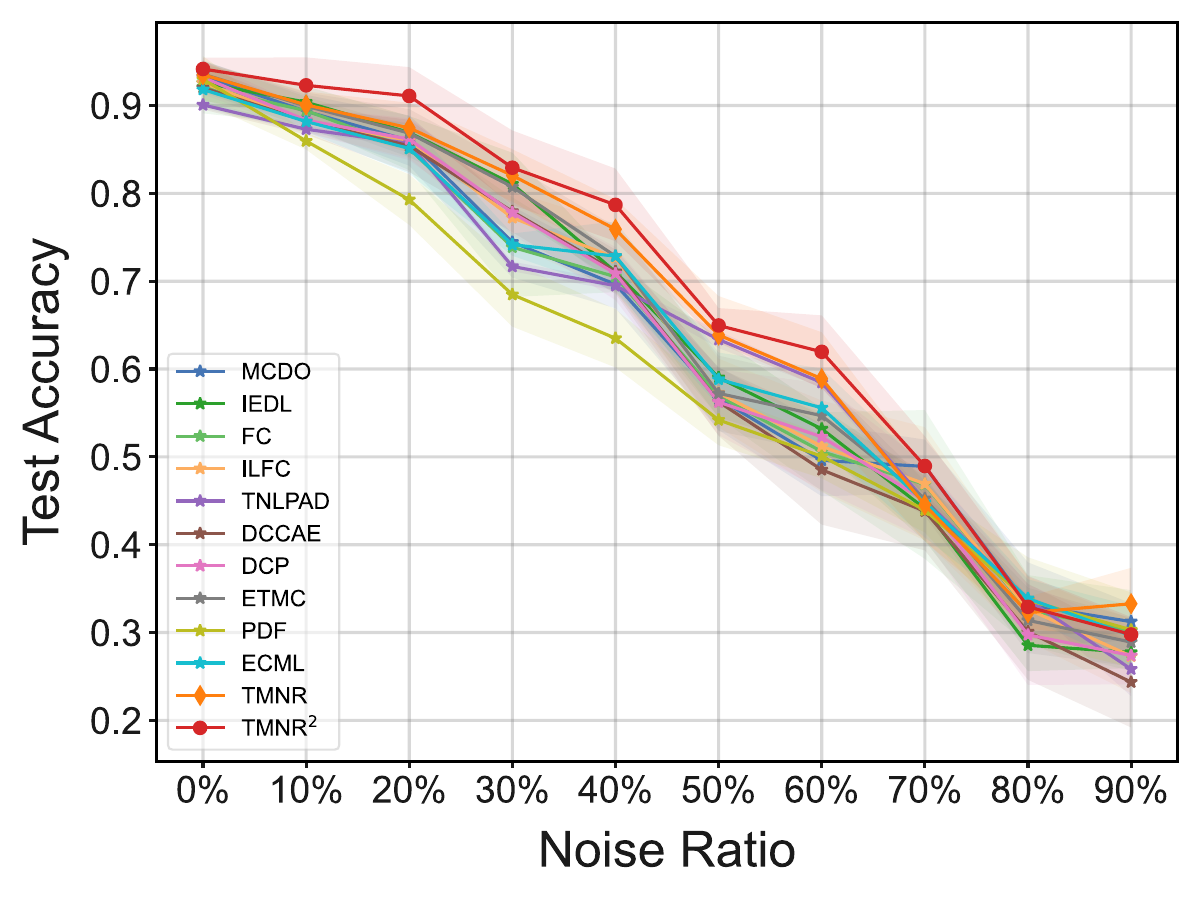}%
		\label{fig_third_case}}
	\\ 
	\subfloat[Caltech]{\includegraphics[width=2.2in]{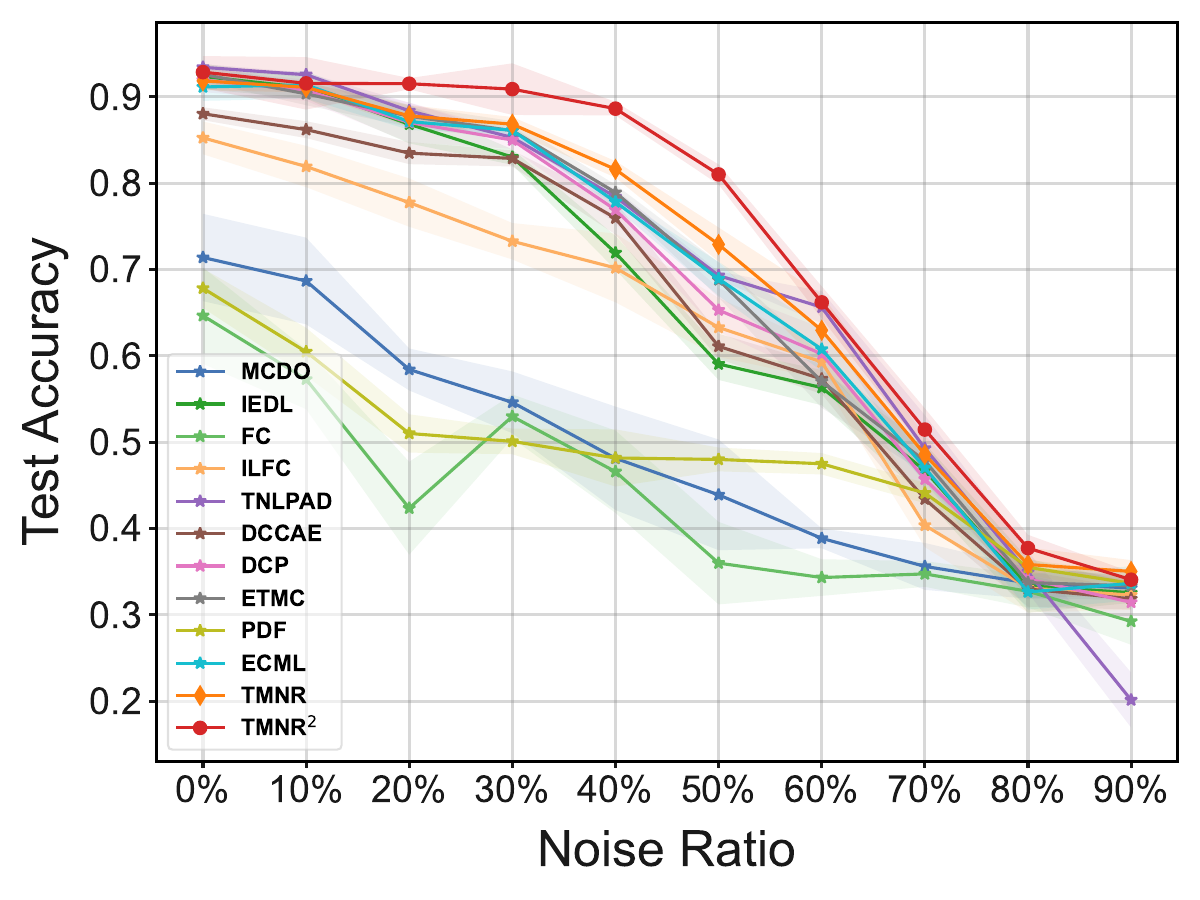}%
		\label{fig_fourth_case}}
	\hfil
	\subfloat[Leaves]{\includegraphics[width=2.2in]{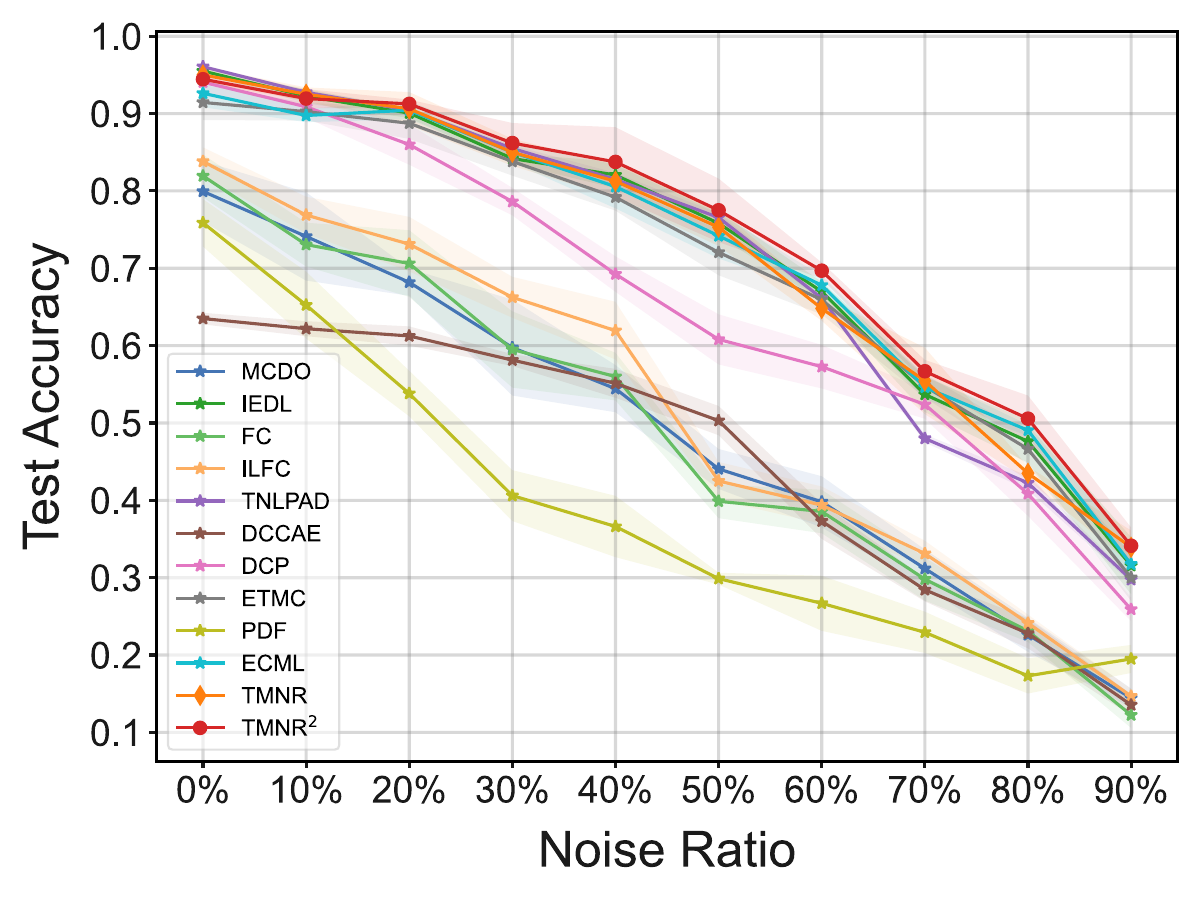}
		\label{fig_fifth_case}}
	\hfil
	\subfloat[CUB]{\includegraphics[width=2.2in]{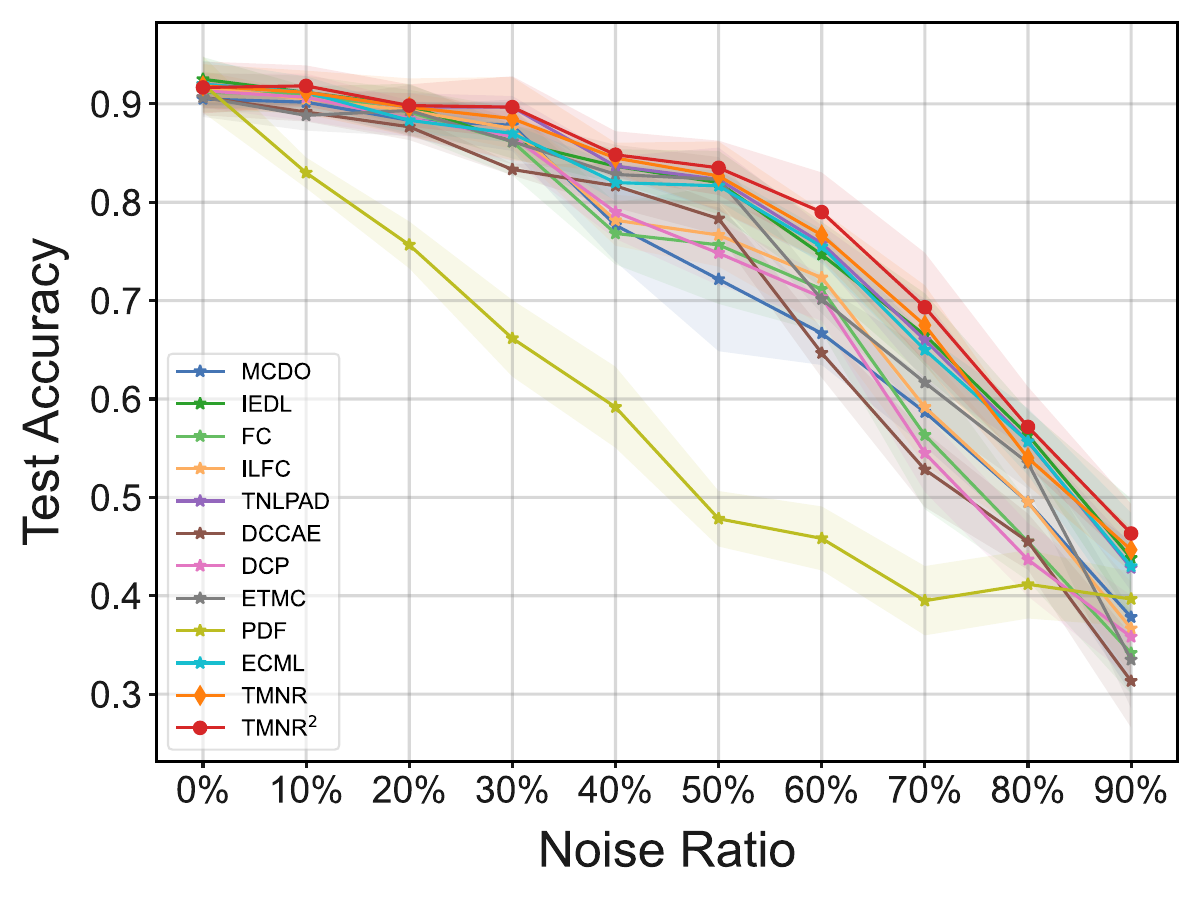}%
		\label{fig_sixth_case}}
	\caption{Comparison of classification performance on 6 datasets containing different ratios of label noise.}
	\label{fig:highNoise}
\end{figure*}

\subsubsection{\textbf{Compared Methods}}
(1) \textbf{Sing-view uncertainty aware methods} contain: MCDO\cite{MCDO} measuring uncertainty by using dropout sampling in both training and inference phases. IEDL \cite{IEDL} is the SOTA method that involving evidential deep learning and Fisher's information matrix.
(2) \textbf{Label noise refining methods} contain: FC \cite{FC} corrects the loss function by a CCN transition matrix. ILFC \cite{CSIDN} explored IDN transition matrix by training a naive model on a subset. TNLPAD \cite{TNLPAD} decouples network parameters into clean and noisy components, applying dynamic constraints to improve robustness against noisy labels.
(3) \textbf{Multi-view feature fusion methods} contain:  DCCAE \cite{DCCAE} train the autoencoder to obtain a common representation between the two views. DCP \cite{DCP} is the SOTA method that obtain a consistent representation through dual contrastive loss and dual prediction loss.
(4) \textbf{Multi-view decision fusion methods} contain:  ETMC \cite{ETMC} estimates uncertainty based on EDL and dynamically fuses the views accordingly to obtain reliable results. PDF \cite{PDF} leverages Co-Belief with theoretical guarantees to reduce generalization error and dynamically adjusts modality weights through relative calibration. ECML \cite{ECML} proposes a novel opinion aggregation strategy to address reliability in conflictive multi-view data.
We summarize baseline methods in Table \ref{tab:methods}. For single-view baselines, we concatenate the features from all views as their input.


\subsubsection{\textbf{Generating Noisy Datasets}}
\label{createNoise}
To simulate the realistic occurrence patterns of noisy labels, we emulate the previous method \cite{NoiseMethod} of destroying labels to generate datasets containing instance-dependent label noise. First we train a trusted classifier $g(\cdot)$ on a subset of the clean dataset using the same model as the Evidence Extraction Network part and make predictions for each sample that will be used for training. Then the noisy labels are obtained by taking into account the amount of evidence $\bm{e}$ obtained for each class and the uncertainty ${u}$ of the opinions.
Specifically, we measure the magnitude of the probability of whether a sample is corrupted in terms of uncertainty, i.e., elements with greater uncertainty are more likely to be corrupted. When a sample is selected for destruction, we pick $\tilde{y}=\mathop{\arg \max}(e_c)$ and $c\neq y \ $, $\ c\in \{1,... ,C\}$.  This means that we set the category that the classifier believes to be most similar to the category to which the instance belongs as the noise label to satisfy our assumption of instance-related label noise.
We added noise with a scale of 10\% to 90\% to all the datasets to evaluate the proposed method.

\subsubsection{\textbf{Implementation Details}}
We performed all experiments using PyTorch 2.2.2  on  NVIDIA RTX 4090 GPUs. For vector-type datasets, the view-specific evidence was extracted by fully connected networks with a ReLU layer. 
We use the Adam optimizer with the learning rate set to 0.003 and a warmup of 40 epochs for the Leaves dataset, while the learning rate for the other datasets is set to 0.001 and a warmup of 15 epochs. The number of neighbors $K$ used for identifying noisy labels is empirically set to 5. The consistency threshold $\epsilon$ is set to 0.95 for datasets with lower noise ratios, 0.8 for UCI in this case, and 0.8 for datasets with higher noise ratios (above 30\%), with UCI adjusted to 0.6. The noise correlation matrix is initialized as unit matrix, and the hyperparameters $\beta$ and $\gamma$ for optimizing the matrix are determined based on the sensitivity analysis described later. Finally, 20\% of the instances in all experiments are split as the test set. We run 5 times for each method to report the mean values and standard deviations. The specific implementation details for the UPMC Food101 dataset are described in Section \ref{sec:food101}.

\begin{figure*}[!t]
	\centering
	\subfloat[UCI IDN-30\%]{\includegraphics[width=2.2in]{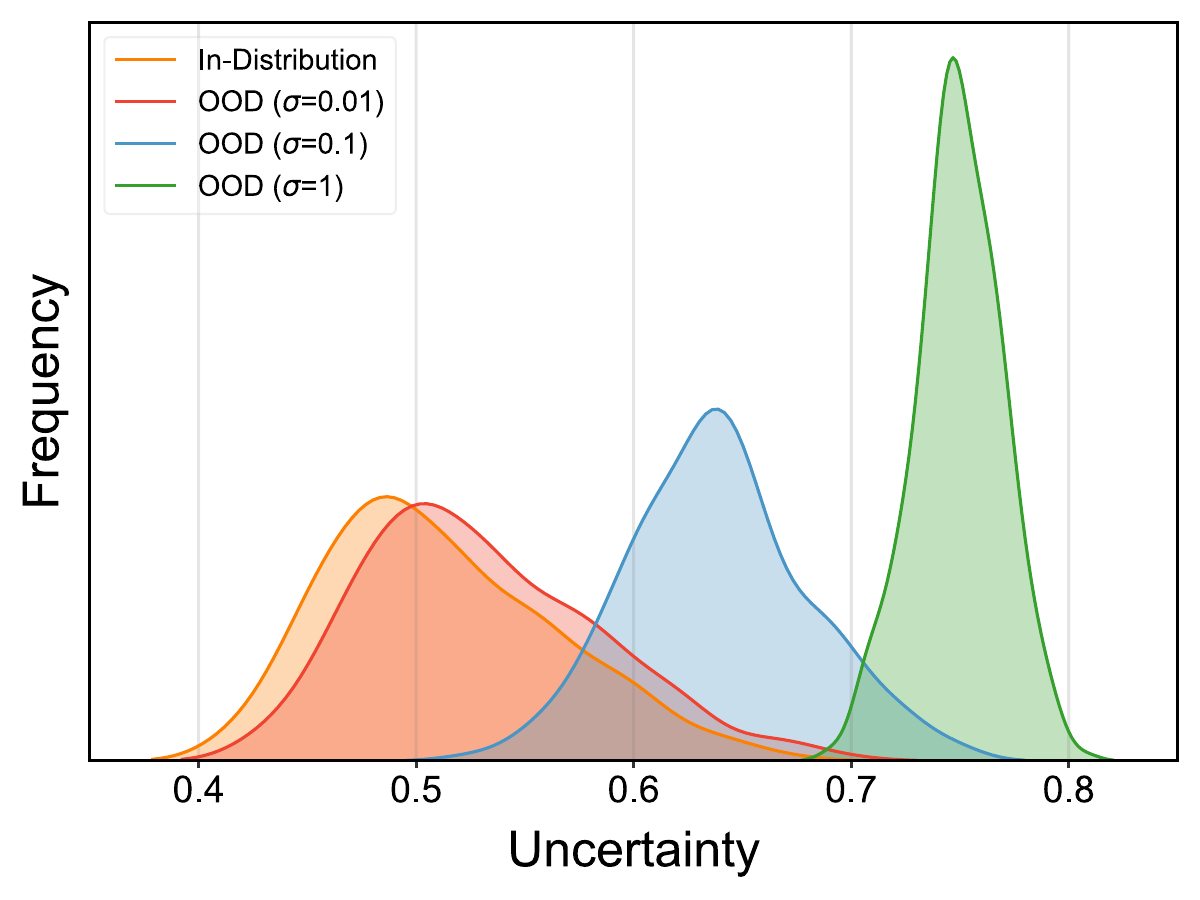}%
		\label{fig_first_case}}
	\hfil
	\subfloat[Caltech IDN-30\%]{\includegraphics[width=2.2in]{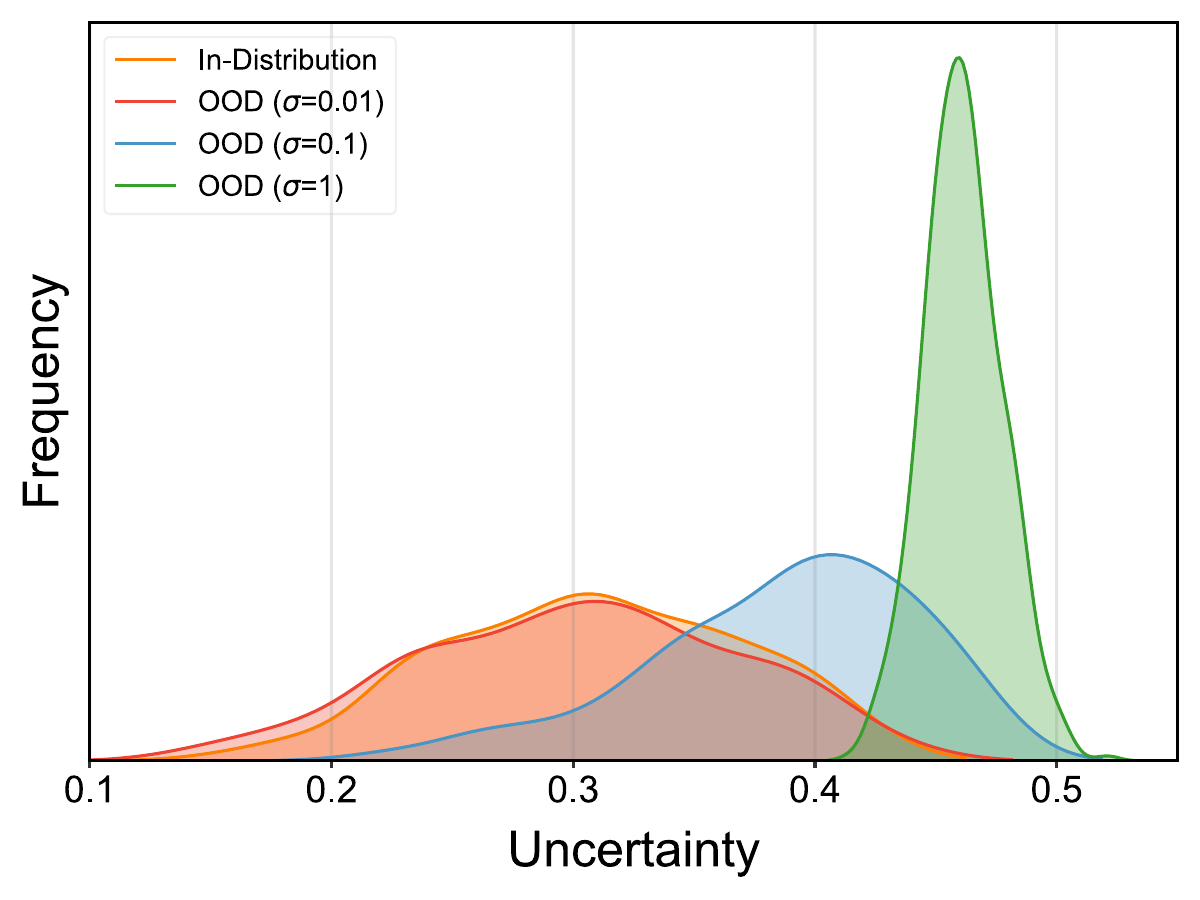}%
		\label{fig_second_case}}
	\hfil
	\subfloat[CUB IDN-30\%]{\includegraphics[width=2.2in]{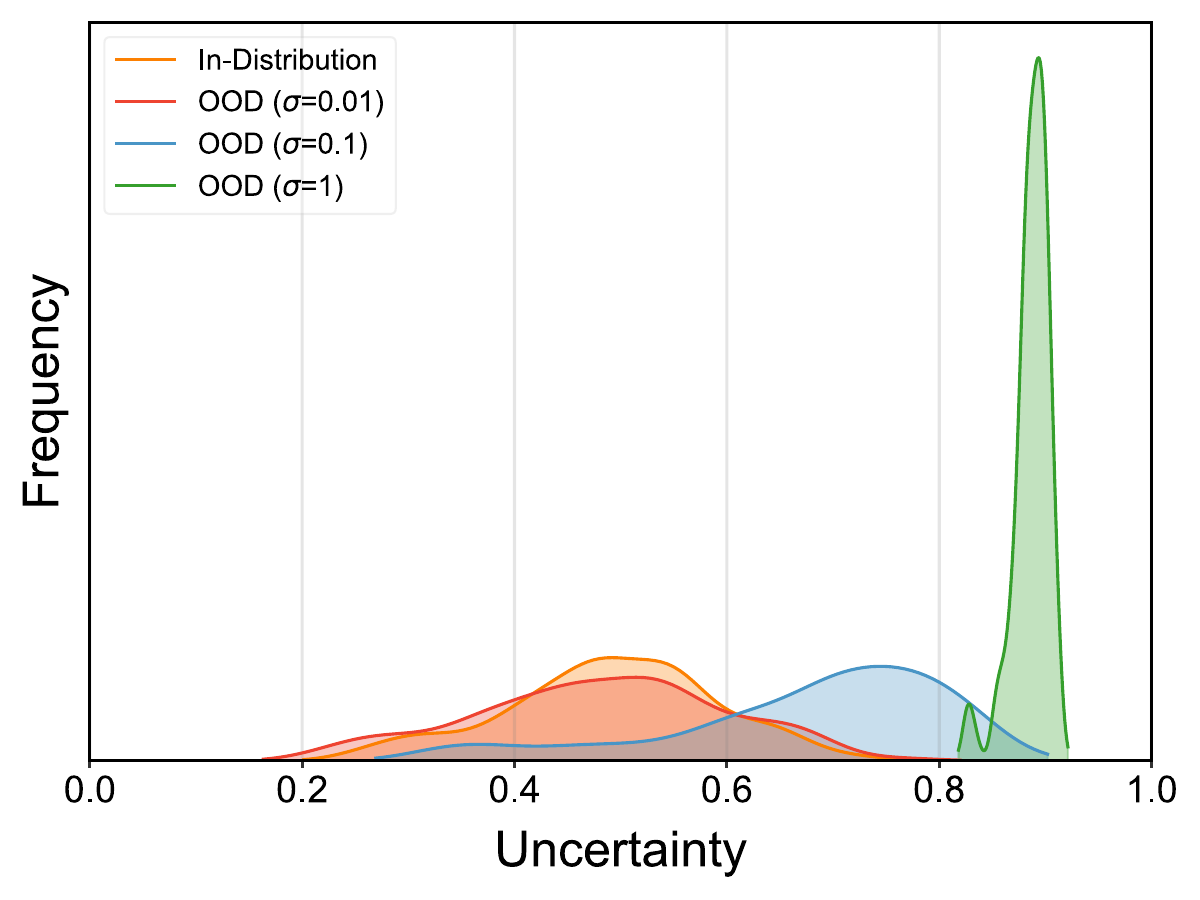}%
		\label{fig_third_case}}
	\\ 
	\subfloat[UCI IDN-50\%]{\includegraphics[width=2.2in]{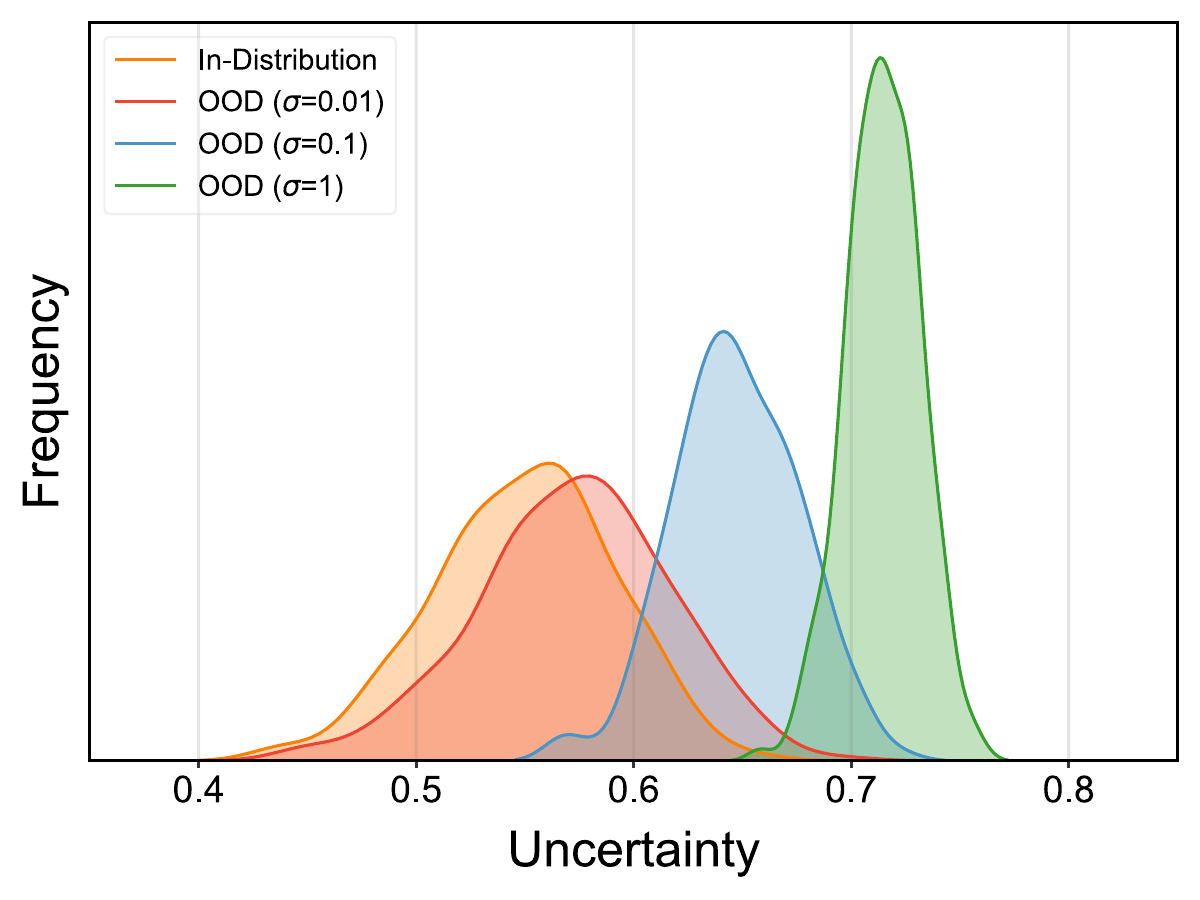}%
		\label{fig_fourth_case}}
	\hfil
	\subfloat[Caltech IDN-50\%]{\includegraphics[width=2.2in]{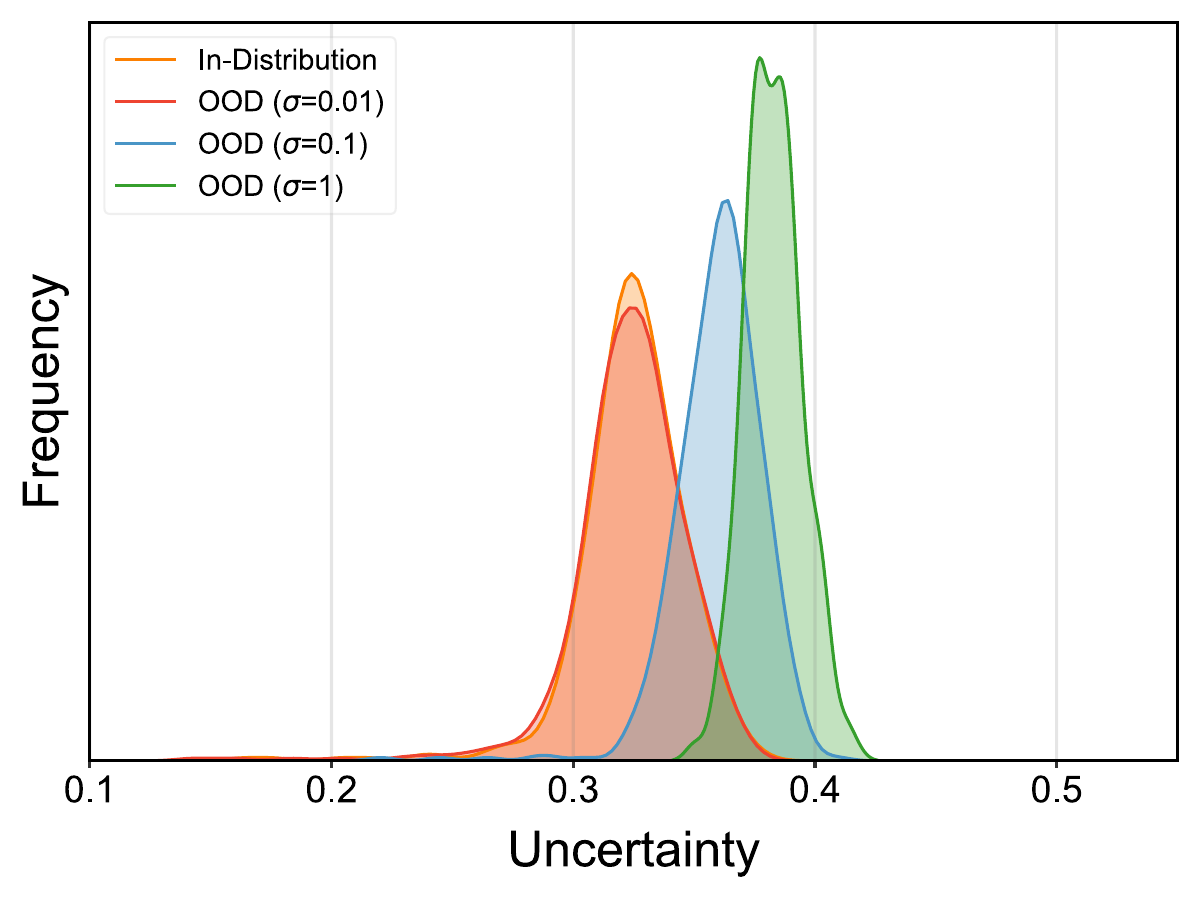}
		\label{fig_fifth_case}}
	\hfil
	\subfloat[CUB IDN-50\%]{\includegraphics[width=2.2in]{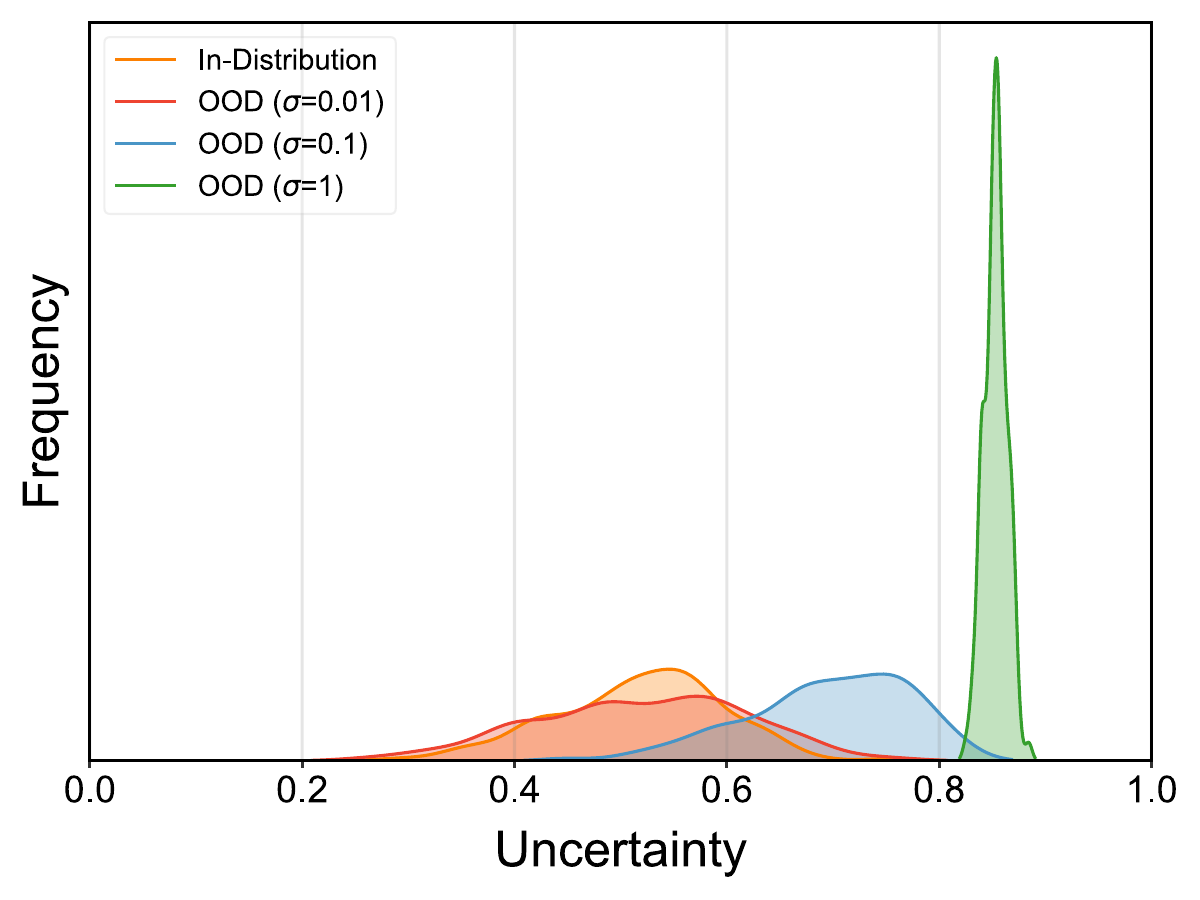}%
		\label{fig_sixth_case}}
	\caption{Uncertainty distribution of in-distribution and out-of-distribution test data obtained from models trained on datasets with 30\% and 50\% label noise.}
	\label{fig:OOD}
\end{figure*}

\subsection{Experimental Evaluation}
To comprehensively evaluate the effectiveness and reliability of our proposed methods, we design experiments to answer the following key questions:

\begin{itemize}
    \item \textbf{Q1: Effectiveness.} Do TMNR and TMNR$^\textbf{2}$ outperform state-of-the-art methods in classification performance under noisy label conditions?
	\item \textbf{Q2: Uncertainty Awareness.} Can our methods effectively recognize and quantify the uncertainty caused by noisy labels or noisy data?
    \item \textbf{Q3: Noise Correction.} How effectively can our methods identify and refine noisy labels across different datasets?
    \item \textbf{Q4: Ablation Study and Parameter Sensitivity.} What are the key components contributing to performance improvements in TMNR$^\textbf{2}$, and how do hyperparameter choices affect the model's effectiveness?
    \item  \textbf{Q5: End-to-End Capability.} Can the proposed methods effectively handle end-to-end image-text classification tasks under noisy supervision?
\end{itemize}

\subsubsection{\textbf{Q1: Effectiveness Comparison}}
The comparison between the proposed methods and baselines on clean and noisy datasets are shown in Table~\ref{tab:accTest} and Figure \ref{fig:highNoise}.
We can observe the following points:
(1) On the clean training dataset, TMNR$^\textbf{2}$ and TMNR achieves performance comparable to state-of-the-art methods.
This finding indicates that the noise forward correction module does not introduce more negative effects even in the absence of noise.
(2) The performance of multi-view feature fusion methods degrade clearly with the noise ratio increase. The reason is the feature fusion would badly affected by noisy labels. 
(3) On noisy training datasets, especially with high noise ratios, TMNR$^\textbf{2}$ and TMNR significantly outperform all baselines. This performance strongly demonstrates the effectiveness of our proposed methods in mitigating the impact of noisy labels through forward correction. Additionally, TMNR$^\textbf{2}$ further enhances classification accuracy on each datasets (especially on the UCI and Caltech datasets, which have fewer classes and are easier to correct) due to its optimized training process and explicit refinement of noisy samples. 

\begin{figure*}[ht]
	\centering
	\subfloat[UCI]{\includegraphics[width=2in]{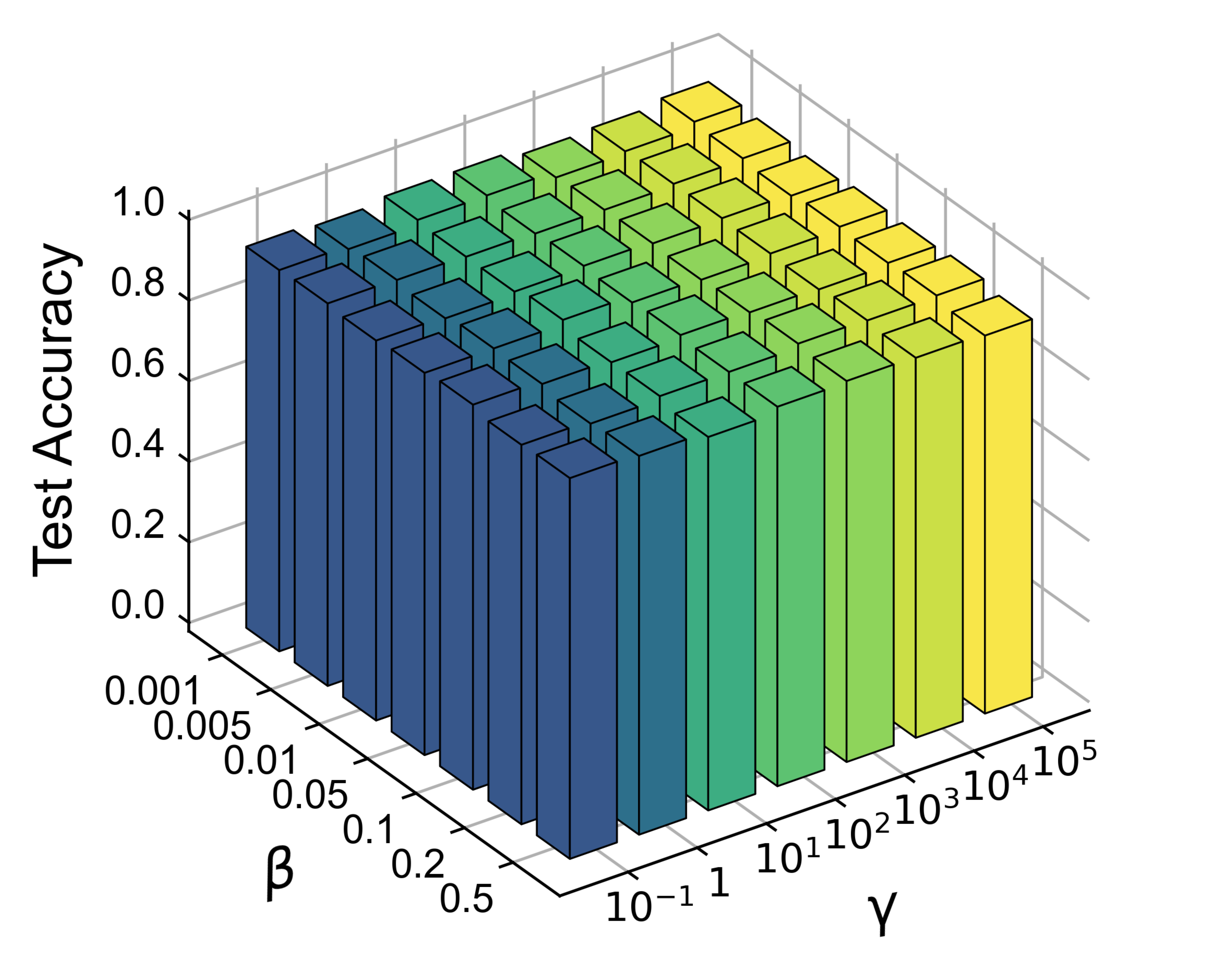}%
		\label{fig_first_case}}
	\hfil
	\subfloat[PIE]{\includegraphics[width=2in]{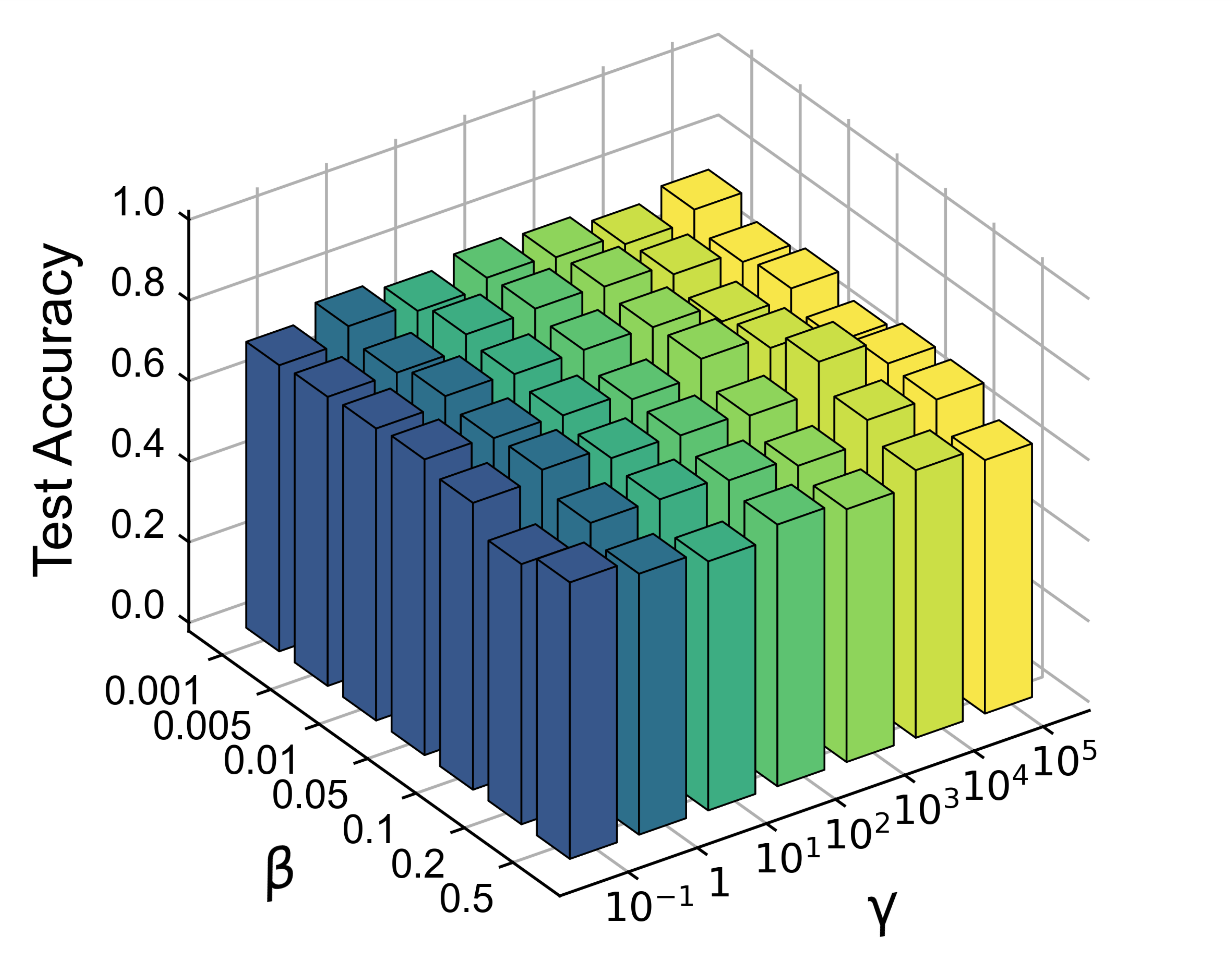}%
		\label{fig_second_case}}
	\hfil
	\subfloat[BBC]{\includegraphics[width=2in]{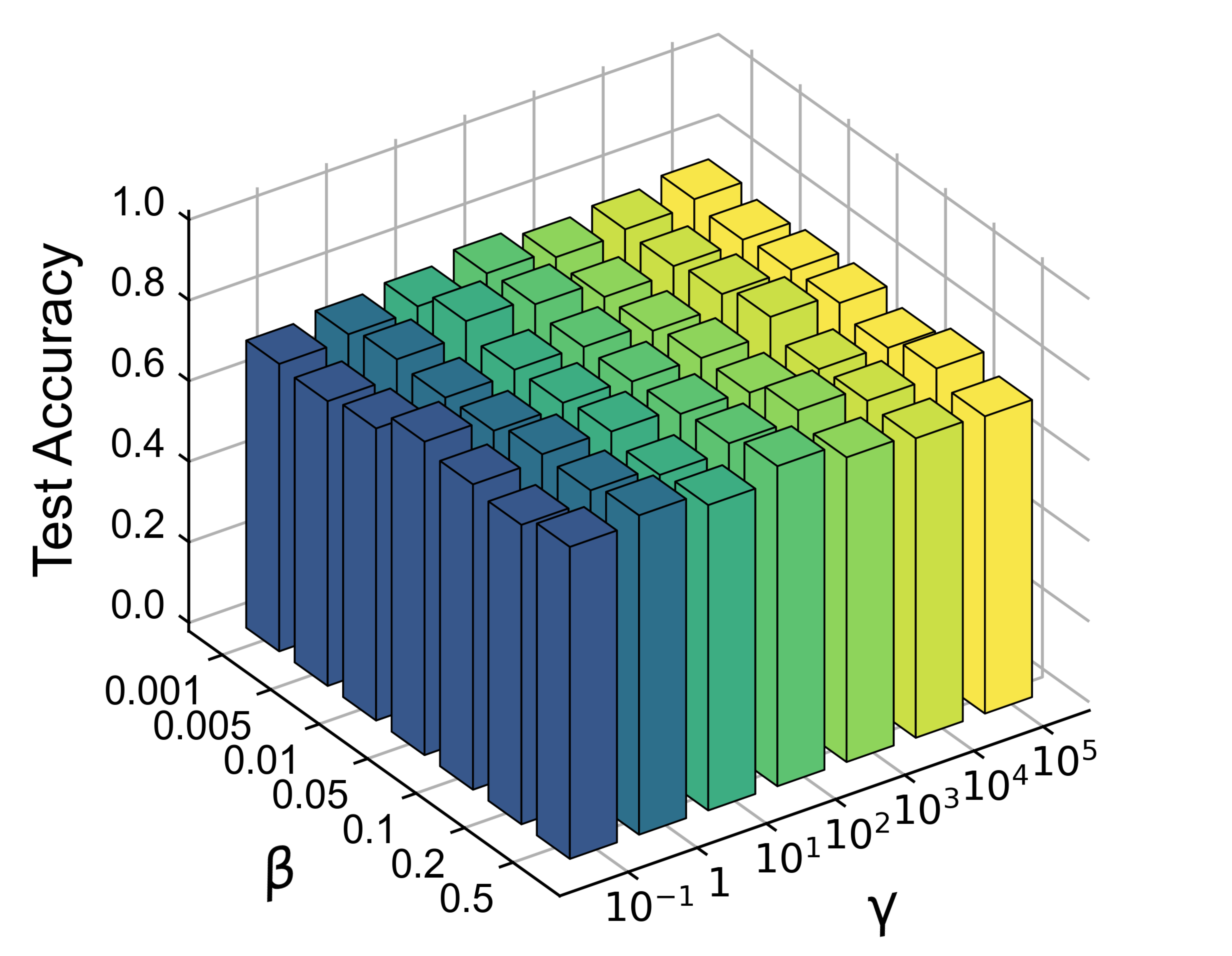}%
		\label{fig_third_case}}
	\\ 
	\subfloat[Caltech]{\includegraphics[width=2in]{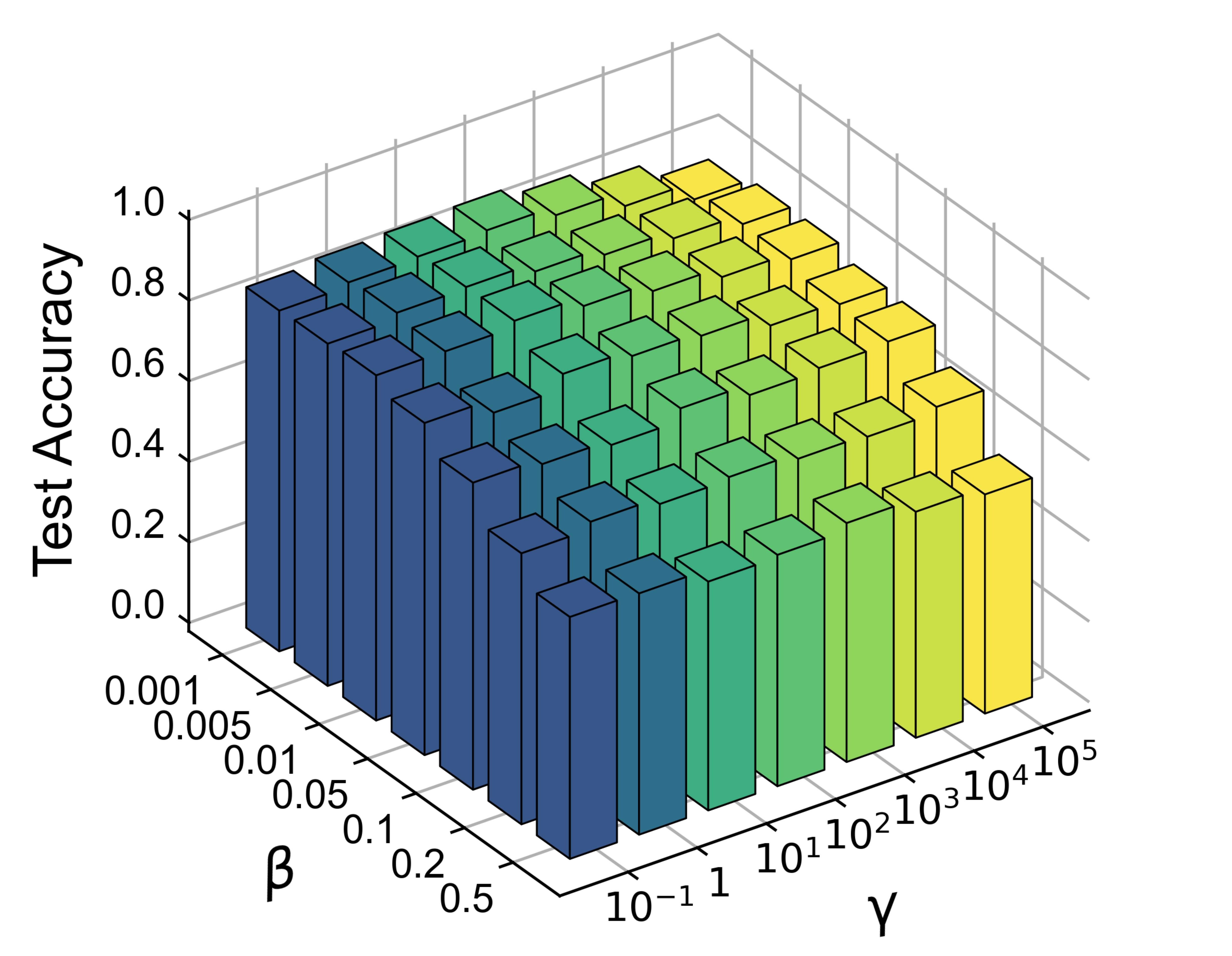}%
		\label{fig_fourth_case}}
	\hfil
	\subfloat[Leaves]{\includegraphics[width=2in]{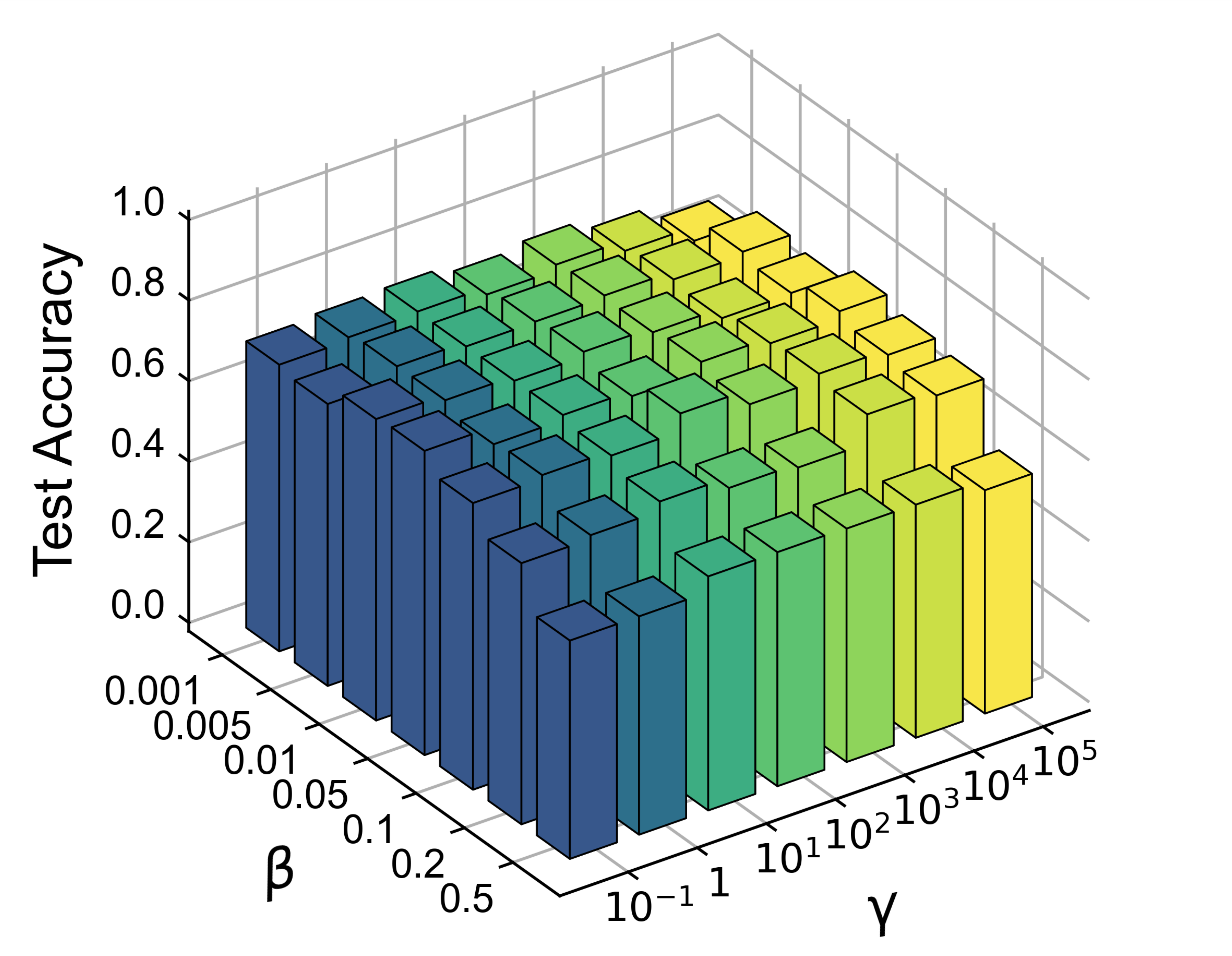}
		\label{fig_fifth_case}}
	\hfil
	\subfloat[CUB]{\includegraphics[width=2in]{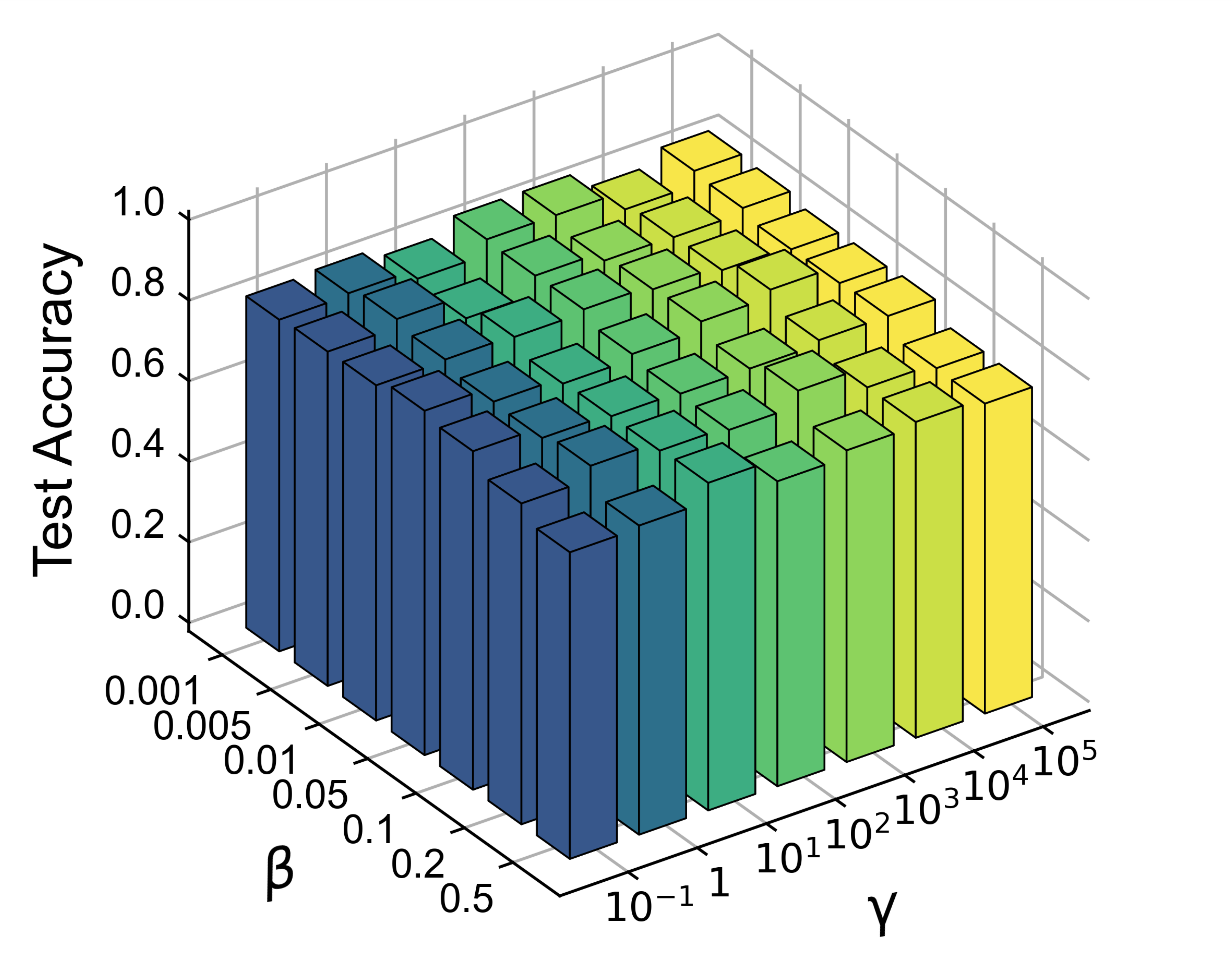}%
		\label{fig_sixth_case}}
	\caption{Sensitivity analysis of the hyperparameters $\beta$ and $\gamma$ in optimizing the noise correlation matrix. Classification accuracy is recorded on datasets with 40\% label noise.}
	\label{fig:hyper}
\end{figure*}

\begin{figure}[t] 
	\includegraphics[width=\columnwidth]{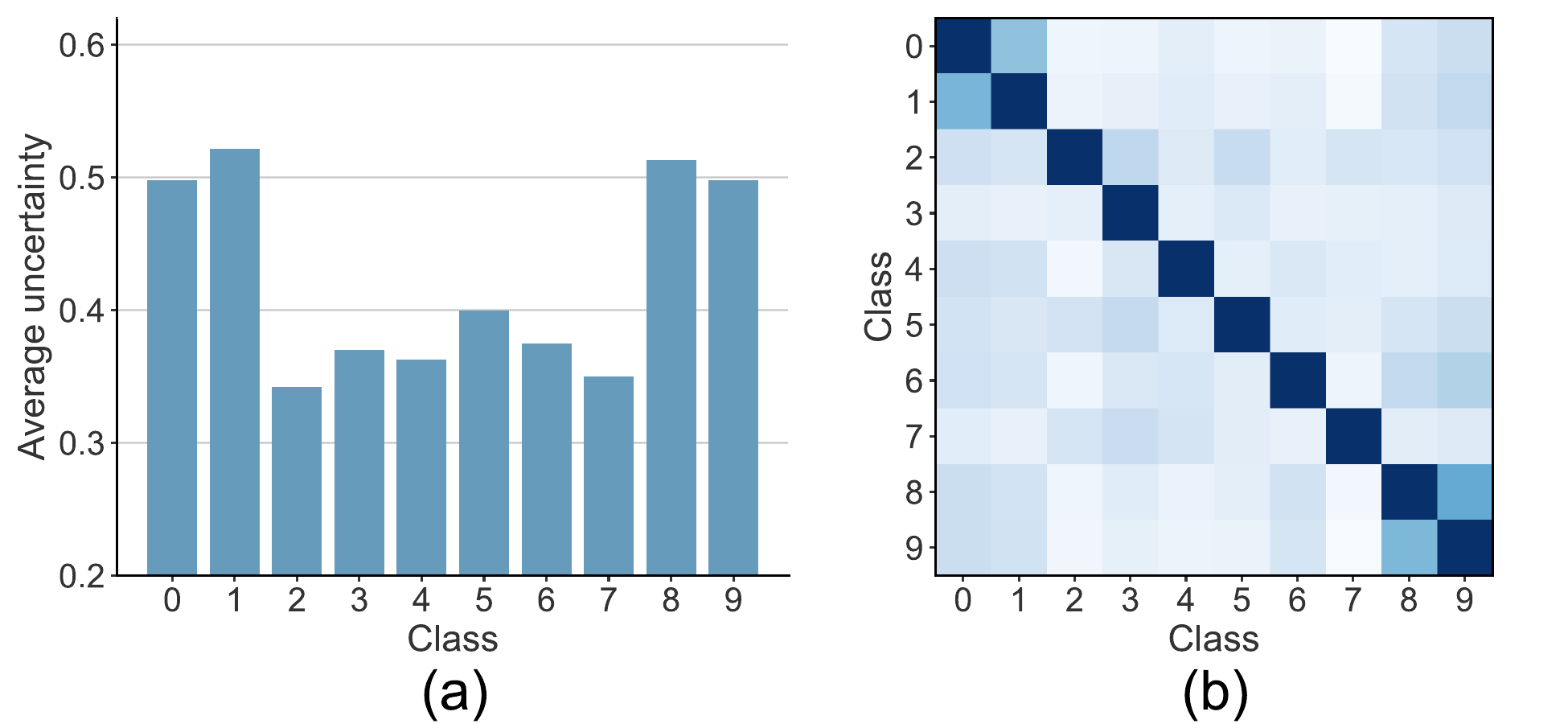} 
	\caption{Visualization of the average uncertainty of each category and correlation matrices in the UCI dataset with selectively flipped labels for samples belonging to classes `$0$' and `$1$', as well as classes `$8$' and `$9$'.}
	\label{fig:Uncertainty_aware}
\end{figure}

\subsubsection{\textbf{Q2: Uncertainty Awareness}}
In real-world datasets, various categories have varying probabilities of being labeled incorrectly. If we can identify the classes that are more likely to be labeled incorrectly during the labeling process, we can apply specialized processing to address these classes, such as involving experts in secondary labeling. As incorrect labeling leads to increased model uncertainty, our model can effectively identify classes that contain noise by assessing their predicted uncertainty.

To observe significant results, we intentionally flipped the labels of samples belonging to classes `0' and `1', as well as classes `8' and `9', within the UCI dataset during training. Subsequently, predictions were made on the test samples, and the average uncertainty for each category was calculated. The results, depicted in Figure \ref{fig:Uncertainty_aware}(a), demonstrate a notable increase in uncertainty for the categories where the labels were corrupted. Figure \ref{fig:Uncertainty_aware}(b) presents a heat map displaying the mean values of all trained correlation matrix parameters. The results clearly illustrate that the model's structure captures the probability of changes in inter-class evidence.

To investigate the impact of noisy supervision on the decision reliability of trusted multi-view models, we evaluated the ability of models trained on noisy label datasets to identify out-of-distribution (OOD) samples. Specifically, the original test samples were treated as in-distribution samples, while test samples with added Gaussian noise of varying standard deviations ($\sigma=0.01, 0.1\ \text{and}\ 1$) were treated as OOD samples. Figure \ref{fig:OOD} illustrates the uncertainty prediction distributions of the TMNR$^\textbf{2}$ model trained on datasets with 30\% and 50\% label noise for both in-distribution and OOD samples.

\begin{figure}[!t]
	\centering
	\subfloat[UCI]{\includegraphics[width=1.7in]{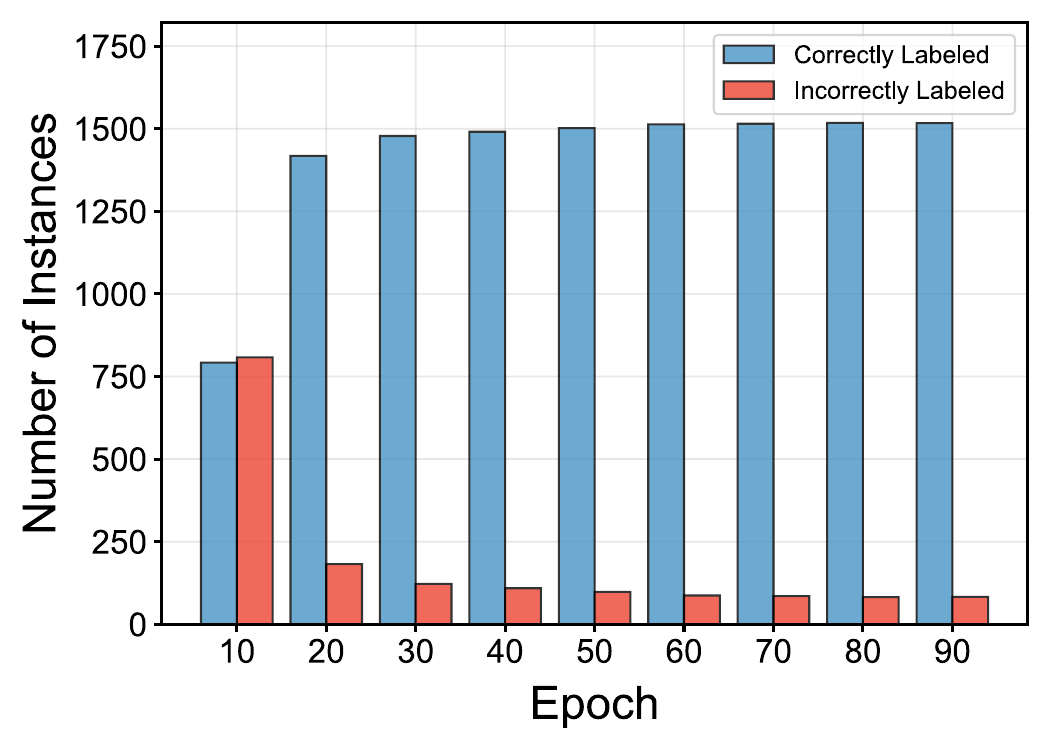}%
		\label{fig_first_case}}
	\hfil
	\subfloat[Caltech]{\includegraphics[width=1.7in]{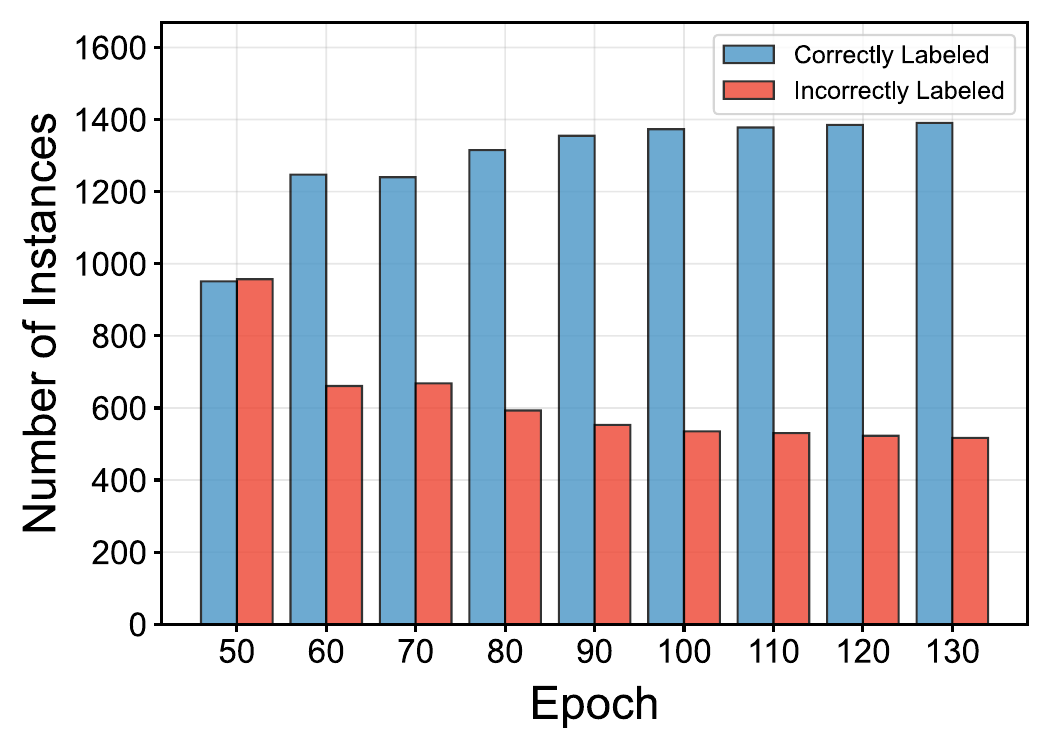}%
		\label{fig_second_case}}
	\caption{Performance of TMNR$^\textbf{2}$ in refining noisy labels on the UCI and Caltech datasets with 50\% label noise.}
	\label{fig:correct}
\end{figure}

It can be observed that: (1) As the label noise increases in the dataset, the model's uncertainty in in-distribution sample decisions also increases, with the distributions narrowing and showing reduced distinguishability. This demonstrates that label noise indeed challenges the reliability of trusted models. (2) As the strength of Gaussian noise in data features increases, the model's decision uncertainty grows to varying extents, which validates the capability of our model to detect data uncertainty under noisy label scenarios.


\subsubsection{\textbf{Q3: Noise Correction Capability}}
To verify the capability of the proposed noise refining module in explicitly identifying and correcting noisy samples, we visualized the changes in the number of correctly and incorrectly labeled training samples over epochs on the UCI and Caltech datasets, as shown in Figure \ref{fig:correct}. Remarkably, on the UCI dataset, which already exhibits high classification accuracy, the noise refining module achieved exceptional correction performance. On the Caltech dataset, approximately one-third of the noisy labels were successfully corrected. Demonstrating the effectiveness of the module.

\begin{table}[tb]
	\caption{Ablation study on three datasets. `TMNR' refers to the base method without improvements, $\mathcal{W}$ represents the improved training pipeline, and $\mathcal{R}$ denotes explicit identification and refinement of noisy samples. The best results are highlighted in \textbf{bold}.}
	\centering
	\begin{tabularx}{0.45\textwidth}{c|YYY|cc}
		\toprule
		\multirow{2}{*}{\textbf{Datasets}} & \multicolumn{3}{c|}{\textbf{Method}} & \multicolumn{2}{c}{\textbf{Instance-Dependent Noise}}                                                                     \\
		                                   & TMNR                                 & $\mathcal{W}$                                         & $\mathcal{R}$ & 30\%                    & 50\%                    \\
		\midrule
		\multirow{4}{*}{\textbf{UCI}} & \ding{52}  & -  & -  & 94.00$\pm$1.46  & 88.90$\pm$0.49  \\
		  & \ding{52}    & \ding{52}    & - & 95.10$\pm$0.87   & 93.45$\pm$1.20          \\
		  & \ding{52}    & -     & \ding{52}     & 94.90$\pm$0.80   & 92.90$\pm$1.51     \\
		& \ding{52}    & \ding{52}   & \ding{52}   & \textbf{95.25$\pm$0.31} & \textbf{94.55$\pm$1.70} \\
		\midrule
		\multirow{4}{*}{\textbf{PIE}} & \ding{52}  & -  & -  & 73.24$\pm$2.08 & 59.85$\pm$2.89   \\
		& \ding{52}    & \ding{52}  & -   & 81.62$\pm$3.45  & 65.15$\pm$2.53        \\
		& \ding{52}    & -          & \ding{52}   & 76.62$\pm$0.81      & 63.38$\pm$5.24      \\
		& \ding{52}    & \ding{52}  & \ding{52}   & \textbf{84.26$\pm$2.08} & \textbf{67.35$\pm$5.43} \\
		\midrule
		\multirow{4}{*}{\textbf{Caltech}} & \ding{52} & - & - & 86.82$\pm$0.83 & 72.89$\pm$1.97    \\
		& \ding{52}  & \ding{52}  & -    & 89.54$\pm$1.00   & 79.79$\pm$1.47    \\
		& \ding{52}  & -    & \ding{52}  & 88.62$\pm$3.84 & 75.31$\pm$3.40  \\
		& \ding{52}  & \ding{52}  & \ding{52}  & \textbf{90.88$\pm$2.99} & \textbf{82.34$\pm$2.16} \\
		\bottomrule
	\end{tabularx}
	\label{tab:Ablation_result}
\end{table}

\subsubsection{\textbf{Q4: Ablation Study and Parameter Sensitivity}} 
To thoroughly investigate the contributions of different components and parameter settings in our framework, we conducted comprehensive ablation studies and parameter sensitivity analyses.

First, we evaluated the effectiveness of the improved training pipeline and the proposed noisy sample refining module.
Starting from the original TMNR method, we applied each improvement individually and evaluated their performance under varying levels of label noise. As shown in Table \ref{tab:Ablation_result}, both improvements consistently enhance performance across different noise levels.

The results demonstrate that: (1) The improved training pipeline ($\mathcal{W}$) significantly boosts performance by reducing the complexity of joint optimization, with substantial gains observed especially on the Caltech dataset (from 72.89\% to 79.79\% under 50\% noise). (2) The explicit noise identification and refinement module ($\mathcal{R}$) provides consistent improvements across all datasets and noise levels, confirming its effectiveness in handling noisy labels. 

Further, we analyzed the sensitivity of the hyperparameters $\beta$ and $\gamma$ in TMNR$^\textbf{2}$ on all datasets containing 40\% label noise. Figure \ref{fig:hyper} illustrates the model's performance when the parameters are varied. It is clear that the optimal parameter combination differs on various datasets, with the model being more sensitive to changes in $\beta$. In contrast, $\gamma$ maintains stable performance across a wide range of values. Therefore, it is recommended that $\beta$ should be carefully tuned. The best parameter combinations found for each dataset are detailed in  our released code.

\begin{table}[tb]
    \caption{Classification accuracy(\%) on UPMC Food101 dataset with different proportions of noise.}
    \centering
    \begin{tabular}{c|cccc}
        \toprule
        \textbf{Methods} & 0\%  & 30\%  & 50\%  & 70\% \\
		\midrule
		IMG  &  65.62$\pm$0.14  & 62.64$\pm$0.33  & 59.65$\pm$0.35 & 47.11$\pm$1.66\\
		TEXT &  83.95$\pm$0.90  & -  & - & - \\
		MCDO &  91.74$\pm$0.20  & 90.33$\pm$0.16  & 86.54$\pm$0.74 & 51.63$\pm$0.72\\
		IEDL &   92.15$\pm$0.34  & 89.99$\pm$0.49    & 88.13$\pm$0.84   & 47.08$\pm$0.20\\
		FC &  91.59$\pm$0.23  & 88.25$\pm$0.62     & 84.71$\pm$1.03 & \underline{52.01$\pm$1.54}\\
		ILFC & 91.73$\pm$0.31 & 90.08$\pm$0.81   & 88.39$\pm$0.90 & 49.98$\pm$1.34\\
		TNLPAD & 92.02$\pm$0.12  & 88.56$\pm$0.53   & 87.37$\pm$0.36 & 49.52$\pm$0.58\\
		ETMC & 91.28$\pm$0.42     & 89.56$\pm$0.87    & 87.23$\pm$1.63 & 47.70$\pm$0.30\\
		PDF  & \textbf{92.70$\pm$0.17}     & 85.07$\pm$0.79   & 81.39$\pm$3.22   & 34.39$\pm$0.16\\
		ECML & 91.98$\pm$0.16    & 90.64$\pm$0.90    & 90.17$\pm$0.15 & 47.39$\pm$0.66\\
		\cmidrule(l{0.5pt}){1-5}
		TMNR  & 92.35$\pm$0.35     & \underline{90.77$\pm$0.60}   & \underline{90.38$\pm$0.35} & 50.15$\pm$1.97\\
		TMNR$^\textbf{2}$  & \underline{92.62$\pm$0.11}   & \textbf{92.63$\pm$0.15} & \textbf{91.95$\pm$0.19}  & \textbf{58.98$\pm$2.58} \\
        \bottomrule
    \end{tabular}
    \label{tab:multimodal_results}
\end{table}

%
\subsubsection{\textbf{Q5: End-to-End Capability Verification}} 
\label{sec:food101}
To verify the end-to-end capability of our proposed methods, we further conducted experiments on the UPMC Food101 dataset, which contains food images with their corresponding textual descriptions. Following previous work \cite{food101set}, we partitioned the dataset into training, validation, and test sets with a ratio of 70\%, 5\%, and 25\%, respectively. Due to its collection process, this dataset inherently contains some data noise, and additional label noise was introduced using the same approach described in Section \ref{createNoise}. We employed ResNet50\cite{ResNet} and BERT\cite{bert} with pre-trained weights as backbone networks to extract features from images and texts respectively. Under this setting, we compared our methods with other classification approaches,  including classification using only unimodality (image and text). All methods using the Adam optimizer and the learning rate of 5e-5. For TMNR$^\textbf{2}$, the number of neighbors $K$ was set to 10, and the consistency threshold $\epsilon$ was consistently set to 0.95 across all noisy datasets during training. We conducted 5 independent runs for each algorithm and reported the mean accuracy along with standard deviations.
 
As shown in Table \ref{tab:multimodal_results}, we compare the performance of our methods with other baselines on this dataset across different noise levels. The results demonstrate that single-modality approaches perform significantly worse than other methods, with the text-only model achieving 83.95\% accuracy on clean data but completely failing to converge under noisy conditions. 
On the clean dataset (0\% noise), PDF achieves the highest accuracy at 92.70\%, with TMNR$^\textbf{2}$ following closely at 92.62\%. However, as noise increases, our methods demonstrate significantly better robustness compared to all baselines. Particularly at high noise levels, TMNR$^\textbf{2}$ outperforms other methods by substantial margins, achieving nearly 7\% improvement over the best baseline at 70\% noise. This superior performance in challenging noisy conditions can be attributed to two key factors: first, our method effectively exploits the complementary information from visual and textual modalities to achieve accurate classification; second, the noise refinement strategy in TMNR$^\textbf{2}$ demonstrates  superior capability in handling complex label corruption patterns in the dataset.

\begin{figure*}[t]
	\centering
	\includegraphics[width=\textwidth]{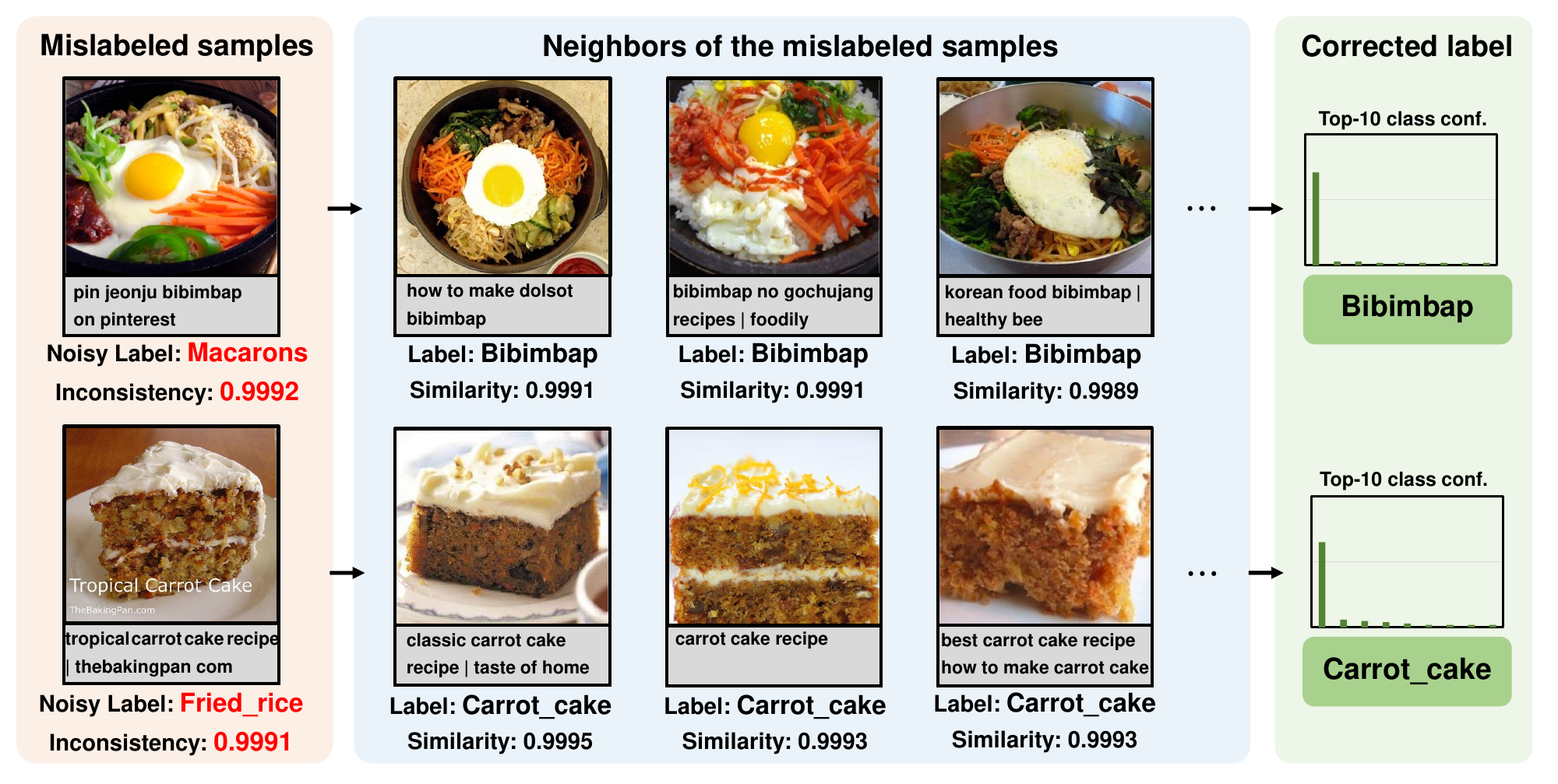}
	\caption{
        Visualization of TMNR$^\textbf{2}$'s noise identification and correction process on Food101 dataset. Left: Two samples identified as likely mislabeled based on label-evidence consistency scores. Middle: Neighbors of the mislabeled samples  with their similarity scores shown, which are used to determine the influence of each neighbor in the correction process. Right: Corrected pseudo-label confidence distributions derived from aggregated neighbor information, demonstrating how TMNR$^\textbf{2}$ leverages neighborhood patterns to refine noisy labels in the multimodal feature space.
	}
	\label{fig:food101visual}
\end{figure*}

Figure \ref{fig:food101visual} illustrates how TMNR$^\textbf{2}$ successfully identifies and corrects mislabeled instances by analyzing consistency patterns across neighbor samples in the multimodal feature space.

\subsection{Discussion}
Based on our comprehensive experiments across various datasets and noise settings, we can summarize the following key insights about the proposed TMNR and TMNR$^\textbf{2}$ methods:

\textbf{Strengths and Their Underlying Mechanisms:}
\begin{itemize}
    \item \textbf{Multi-view complementary information utilization:} 
	Our approach effectively exploits complementary feature representations across views, where different views may capture different aspects of the same instance, allowing the model to maintain accurate predictions even when some views are more susceptible to noise interference. In Food101 experiments, the fusion of visual and textual modalities always performs better.

	\item \textbf{Uncertainty-guided noise correlation modeling:} By incorporating uncertainty estimation into noise correlation matrices, our approach establishes an interpretable connection between predictive uncertainty and label corruption probability, allowing adaptive trust assignment to different samples.

	\item \textbf{Noise identification and refinement:} TMNR$^\textbf{2}$ separates clean and noisy samples for targeted optimization while employing neighborhood-based refinement strategies. This combination allows the model to learn from reliable patterns in clean samples while simultaneously correcting noisy ones through local consistency enforcement, particularly effective at high noise ratios as shown in Figure \ref{fig:correct}.
\end{itemize}

\textbf{Limitations:}
\begin{itemize}
    \item \textbf{Diminishing returns at extreme noise levels:} While our methods demonstrate remarkable resilience to noise, performance still degrades substantially at extremely high noise levels (e.g., 70\% on Food101), indicating room for improvement in identifying reliable patterns under severe corruption.
    
	\item \textbf{Computational resource requirements:} When dealing with large training datasets, instance-specific noise correlation matrices may consume a lot of computational resources, and more efficient methods of matrix parameterization should be explored.
    
	\item \textbf{Sensitivity to hyperparameters:} The method requires careful tuning of multiple hyperparameters. While this provides flexibility, it also means careful parameter tuning is required for optimal performance on new datasets.

\end{itemize}

These insights not only validate the effectiveness of our proposed approaches but also provide direction for future improvements to address the identified limitations.

\section{Conclusion}
\label{conclusion}
In this work, we explore the problem of trusted multi-view classification in noisy label scenarios. By designing a noise correlation matrices to learn the potential patterns of label corruption and supervising with uncertainty information, we propose Trusted Multi-view Noise Refining (TMNR). Furthermore, we enhance TMNR with TMNR², which explicitly identifies noisy samples using neighborhood information and assigns clean and noisy samples to different training objectives in separate modules, effectively reducing the difficulty of joint optimization.
The experimental results demonstrate the effectiveness of the proposed method by comparing it with state-of-the-art trusted multi-view learning and labeled noise learning baseline methods on 7 publicly available datasets.

For future work, we plan to explore more efficient parameterizations of the instance-specific noise correlation matrices to reduce computational resource requirements. Additionally, we believe our approach has potential for improving the reliability of large language and multimodal models by helping identify and correct noisy instruction-response pairs in their training data.

\section*{Acknowledgments}
This research was supported by the National Natural Science Foundation of China (Grant Nos. 62133012, 62303366) and Natural Science Basic Research Program of Shaanxi under Grant No.2023-JC-QN-0648.

\bibliographystyle{IEEEtran}
\bibliography{ref}

\begin{IEEEbiography}[{\includegraphics[width=1in,height=1.25in,clip,keepaspectratio]{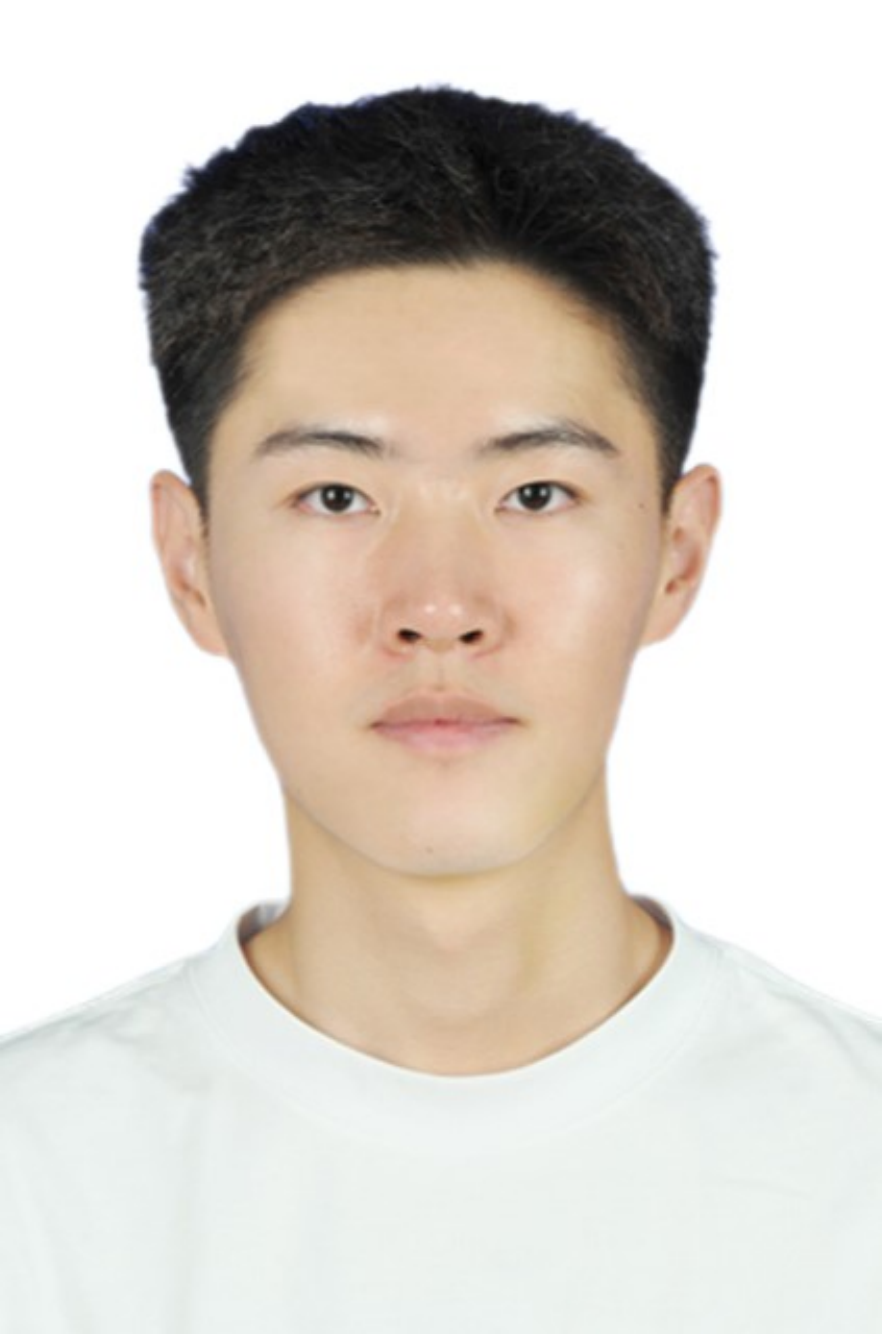}}]{Yilin Zhang}
	received the BS degrees in Computer Science and Technology from Xidian University, Xi'an, China, in 2023.   He is currently working toward the PhD degree with the School of Computer Science and Technology, Xidian University.   His research interests include multi-view learning, label-noise learning and uncertainty calibration.
\end{IEEEbiography}

\begin{IEEEbiography}[{\includegraphics[width=1in,height=1.25in,clip,keepaspectratio]{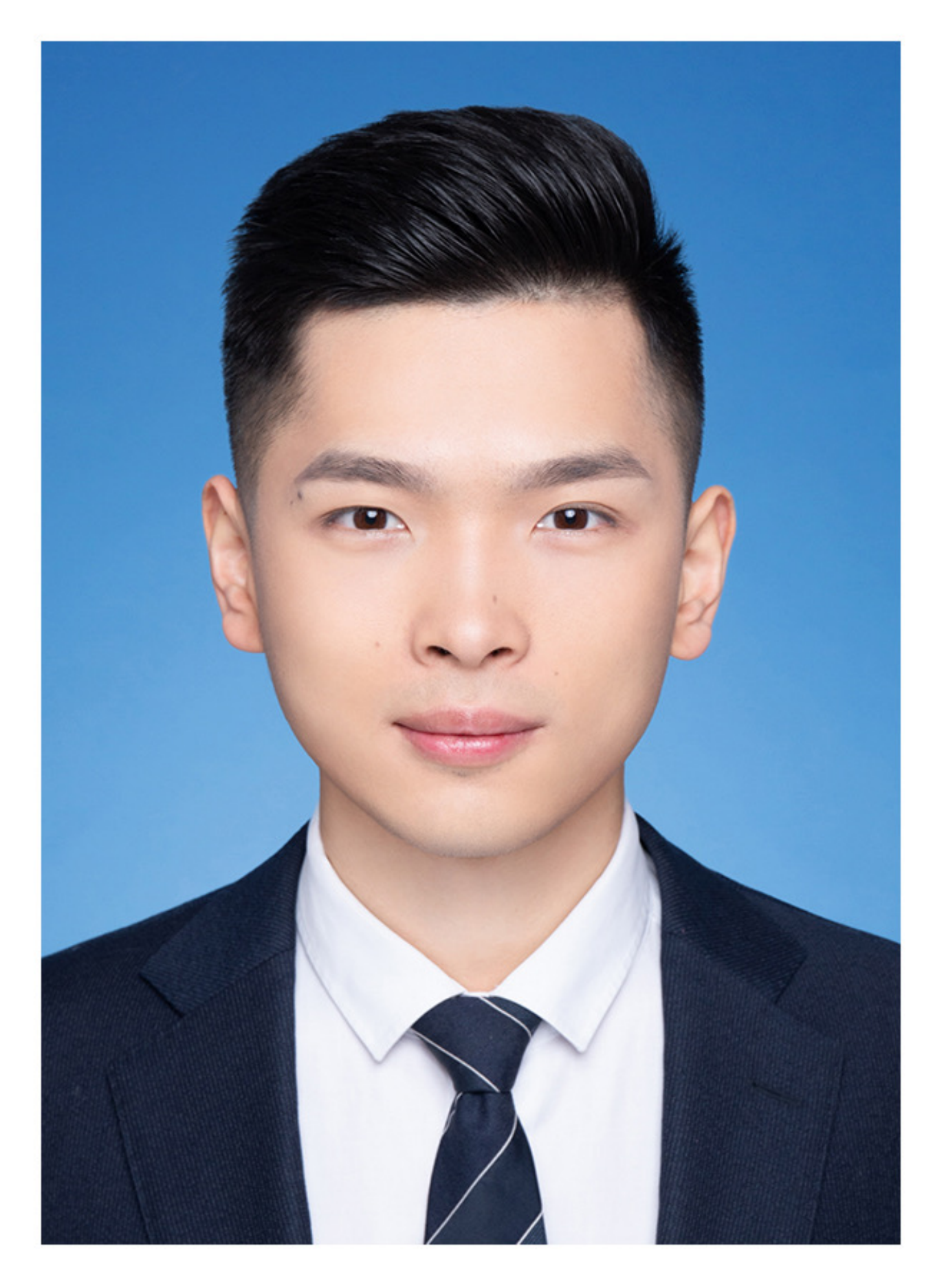}}]{Cai Xu}
    is currently an associate professor at the School of Computer Science and Technology, Xidian University. He received the Outstanding Paper Award at AAAI 24 as the first author. His research interests include trustworthy machine learning and multi-view learning.
\end{IEEEbiography}

\begin{IEEEbiography}[{\includegraphics[width=1in,height=1.25in,clip,keepaspectratio]{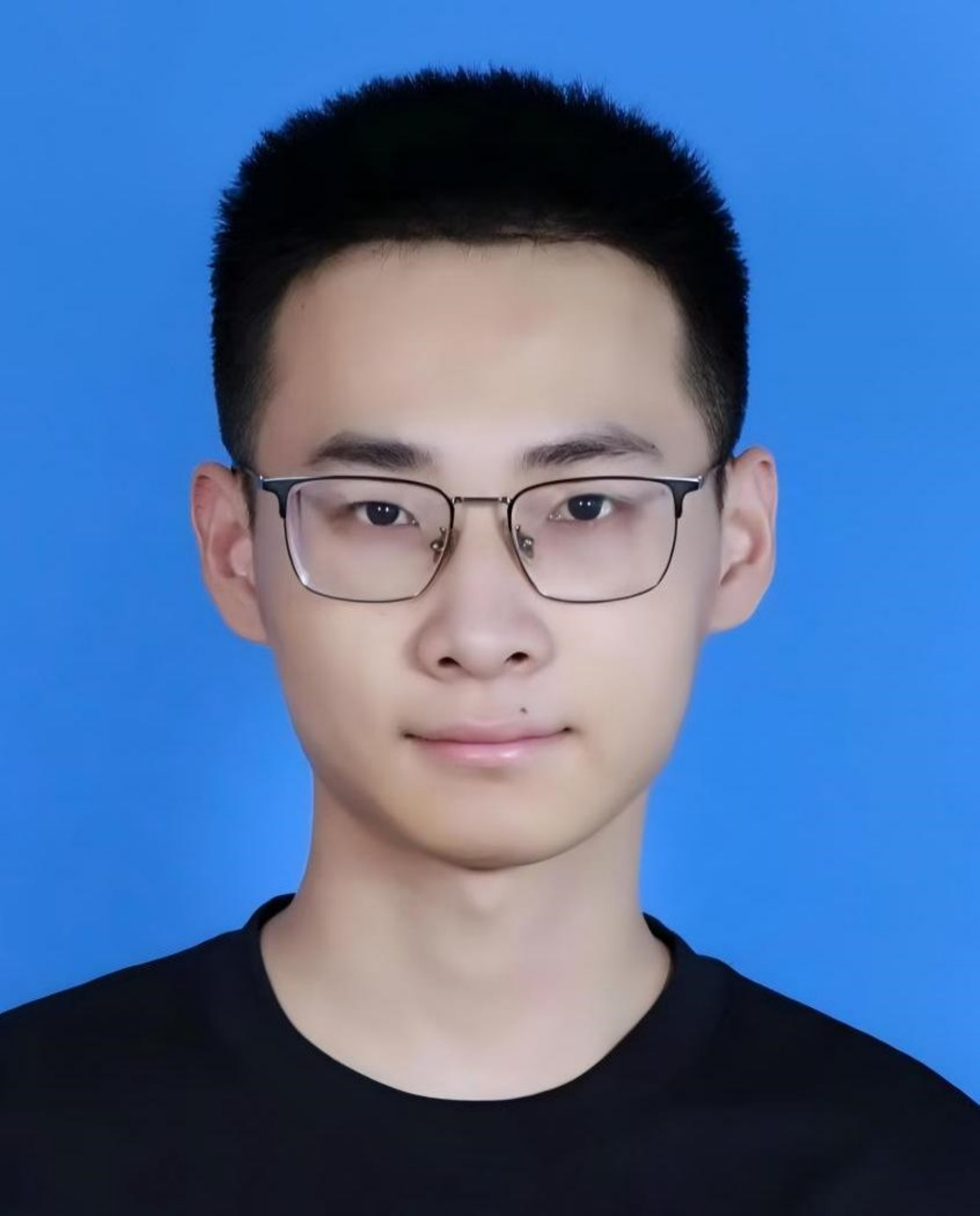}}]{Han Jiang}
	is currently pursuing the BS degree in Computer Science at Shanxi University, China. He has  been admitted to Xidian University, China, where he will pursue the MS degree. His research interests include machine learning and multi-view learning.
\end{IEEEbiography}

\begin{IEEEbiography}[{\includegraphics[width=1in,height=1.25in,clip,keepaspectratio]{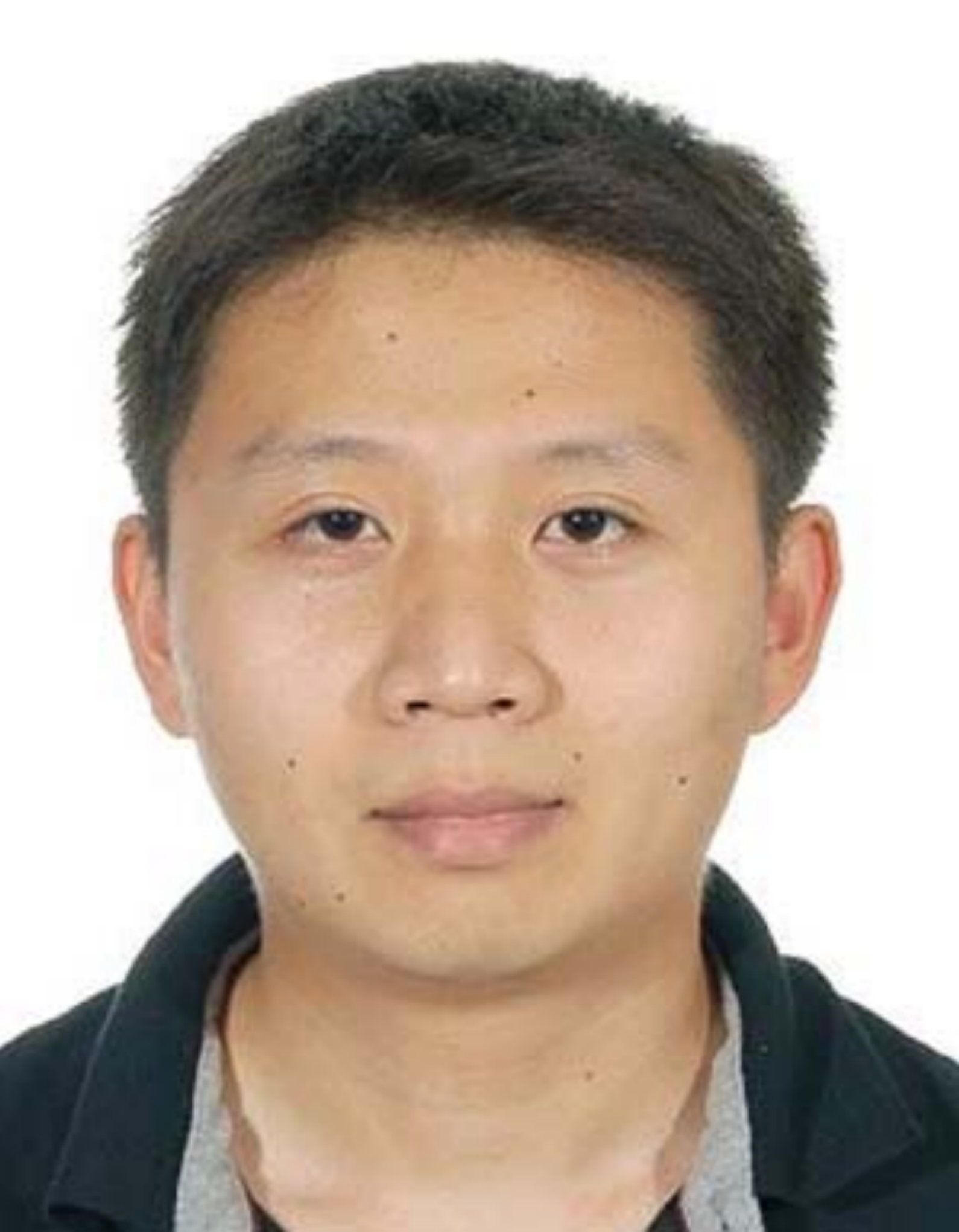}}]{Ziyu Guan}
	received the BS and PhD degrees in computer science from Zhejiang University, Hangzhou, China, in 2004 and 2010, respectively. He had worked as a research scientist with the University of California at Santa Barbara from 2010 to 2012, and as a professor with the School of Information and Technology, Northwest University, China from 2012 to 2018. He is currently a professor with the School of Computer Science and Technology, Xidian University. He is an Associate Editor for well-known journals such as IEEE TKDE, KAIS and JMLC. His research interests include attributed graph mining and search, machine learning, expertise modeling and retrieval, and recommender systems.
\end{IEEEbiography}

\begin{IEEEbiography}[{\includegraphics[width=1in,height=1.25in,clip,keepaspectratio]{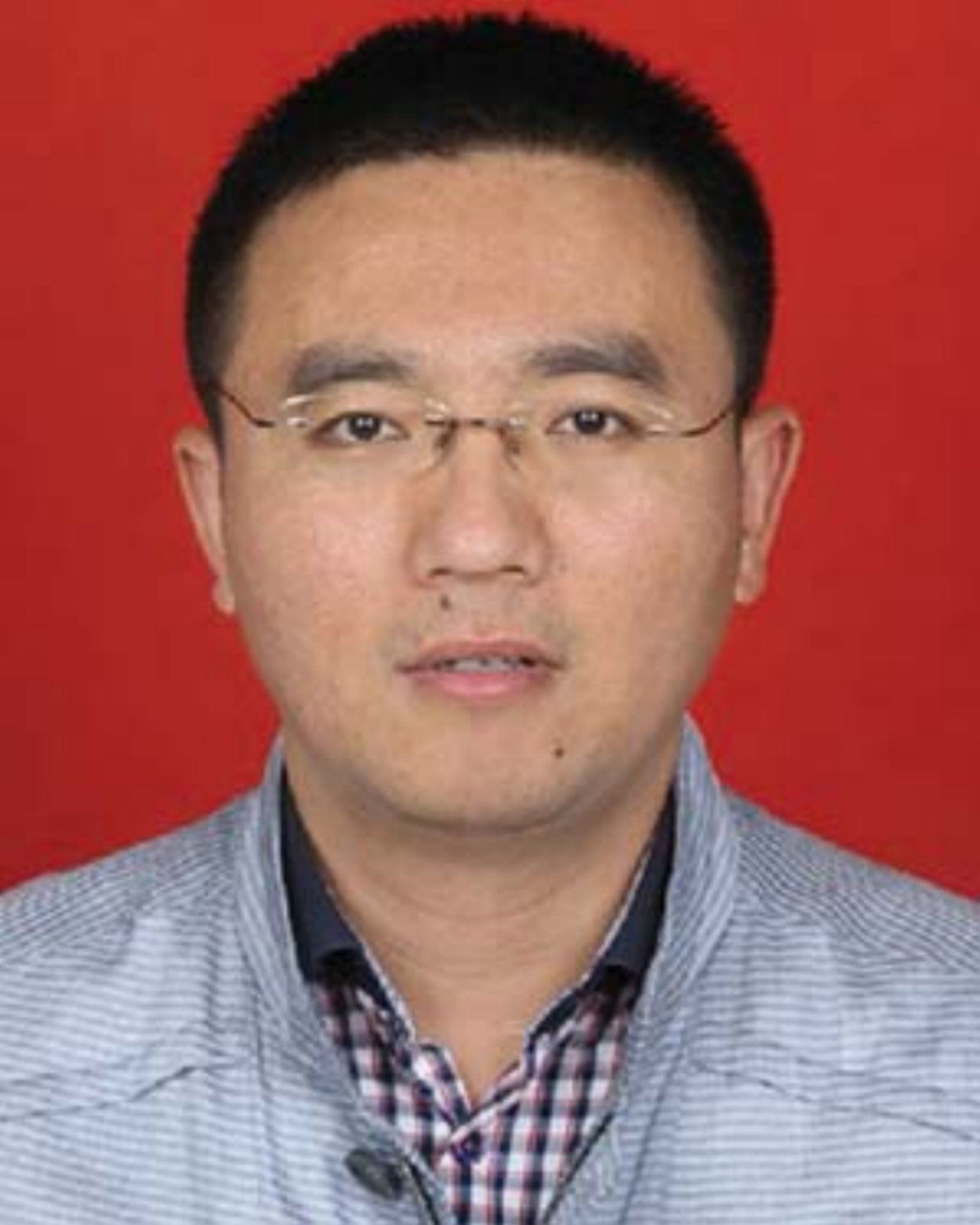}}]{Wei Zhao}
	received the BS, MS and PhD degrees from Xidian University, Xi'an, China, in 2002, 2005, and 2015, respectively. He is currently a professor with the School of Computer Science and Technology, Xidian University. He is an Associate Editor for well-known journals such as IEEE TKDE and TNNLS. His research interests include direction is pattern recognition and intelligent systems, with specific interests in attributed graph mining and search, machine learning, signal processing, and precision guiding technology.
\end{IEEEbiography}

\vspace{-24\baselineskip}
\begin{IEEEbiography}[{\includegraphics[width=1in,height=1.25in,clip,keepaspectratio]{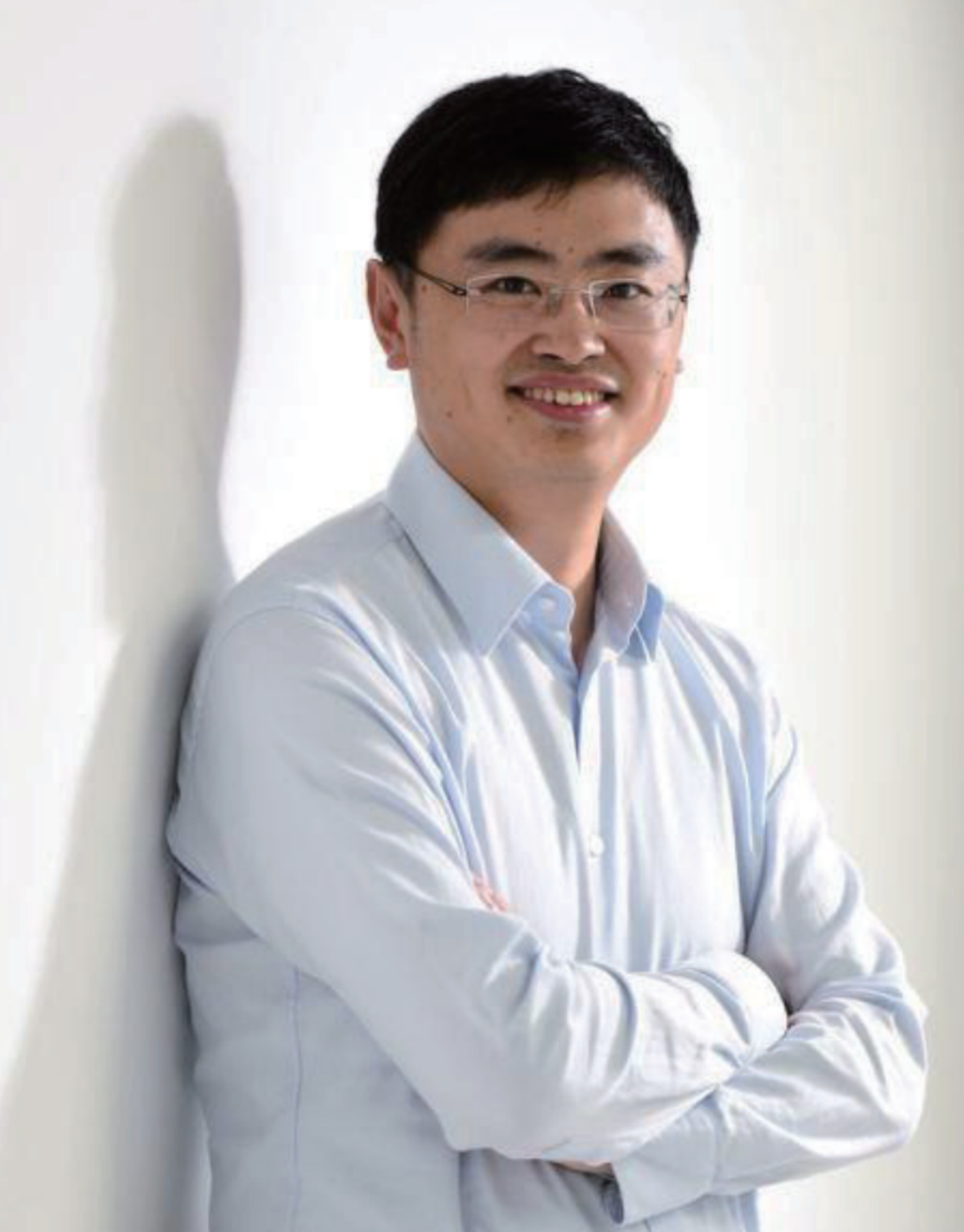}}]{Xiaofei He}
 is currently a Professor at the State Key Lab of CAD\&CG, Zhejiang University, and the CEO of FABU Technology Co., Ltd.  His research interests include machine learning, deep learning, and autonomous driving.  He has authored/co-authored more than 200 technical papers with over 45,000 times citations on Google Scholar.  He was awarded the Best Paper Award at AAAI 2012 and is a Fellow of IAPR.
\end{IEEEbiography}

\vspace{-24\baselineskip}
\begin{IEEEbiography}[{\includegraphics[width=1in,height=1.25in,clip,keepaspectratio]{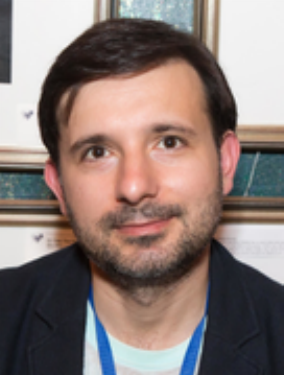}}]{Murat Sensoy}
	is an applied research scientist at Amazon Alexa AI, EC2A 2FA, London, U.K. His research interests include reliable machine learning systems. Şensoy received his Ph.D. degree in computer engineering from Bogazici University.
\end{IEEEbiography}

\end{document}